\pdfoutput=1

\documentclass{article}
\usepackage{natbib}
\setcitestyle{square,comma,numbers}

\usepackage[final]{neurips_2024}

\usepackage[utf8]{inputenc} 
\usepackage[T1]{fontenc}    
\usepackage{url}            
\usepackage{booktabs}       
\usepackage{amsfonts}       
\usepackage{nicefrac}       
\usepackage{microtype}      
\usepackage{xcolor}         

\usepackage{bm}
\usepackage{multirow}
\usepackage{graphicx}
\usepackage{enumitem}
\usepackage{wrapfig}
\usepackage{algorithm}
\usepackage{algorithmic}
\usepackage{amsmath}
\usepackage{caption}
\usepackage{subcaption}
\usepackage{hyperref}       

\newcommand{\mathbbm}[1]{\text{\usefont{U}{bbm}{m}{n}#1}} 

\newcommand{\etal}{\textit{et al.}}
\newcommand{\eg}{\textit{e.g.}}
\newcommand{\ie}{\textit{i.e.}}
\DeclareMathOperator*{\argmax}{arg\,max}

\title{Learning Where to Edit Vision Transformers}

\author{
Yunqiao Yang$^1$\thanks{Part of the work was done when the author interned at Tencent AI Lab.}  \quad  Long-Kai Huang$^2$\thanks{Corresponding authors.} \quad Shengzhuang Chen$^1$  \quad  \\
\textbf{Kede Ma}$^1$ \quad  \textbf{Ying Wei}$^3$\footnotemark[2]  \\
$^1$City University of Hong Kong \quad  $^2$Tencent AI Lab \quad $^3$Zhejiang University \\
\texttt{\{yqyang.cs, szchen9-c\}@my.cityu.edu.hk}  \quad  \texttt{hlongkai@gmail.com}  \\
\texttt{kede.ma@cityu.edu.hk}  \quad  \texttt{ying.wei@zju.edu.cn}
}

\begin{document}

\maketitle

\begin{abstract}
 Model editing aims to data-efficiently correct predictive errors of large pre-trained models while ensuring generalization to neighboring failures and locality to minimize unintended effects on unrelated examples. While significant progress has been made in editing Transformer-based large language models, effective strategies for editing vision Transformers (ViTs) in computer vision remain largely untapped. In this paper, we take initial steps towards correcting predictive errors of ViTs, particularly those arising from subpopulation shifts. Taking a locate-then-edit approach, we first address the ``where-to-edit'' challenge by meta-learning a hypernetwork on CutMix-augmented data generated for editing reliability. This trained hypernetwork produces generalizable binary masks that identify a sparse subset of structured model parameters,  responsive to real-world failure samples. Afterward, we solve the ``how-to-edit'' problem by simply fine-tuning the identified parameters using a variant of gradient descent to achieve successful edits. To validate our method, we construct an editing benchmark that introduces subpopulation shifts towards natural underrepresented images and AI-generated images, thereby revealing the limitations of pre-trained ViTs for object recognition. Our approach not only achieves superior performance on the proposed benchmark but also allows for adjustable trade-offs between generalization and locality. Our code is available at \url{https://github.com/hustyyq/Where-to-Edit}.

\end{abstract}

\section{Introduction}\label{introduction}
In many scientific and engineering disciplines, computational models serve as approximations of complex real-world phenomena. As a consequence, they are inherently prone to predictive errors, aptly encapsulated by George Box's adage: ``\textit{All models are wrong, but some are useful}.'' Model editing~\cite{thrun1995learning, buntine1991theory, de2021editing,mitchell2022fast} has emerged as a promising technique to make (large) pre-trained models \textit{more useful} by enabling targeted updates to model behavior on specific inputs or tasks 
in a data-efficient manner without pre-training again from scratch.
An ideal model editing method should satisfy three major desiderata~\cite{de2021editing, yao2023editing}: 1) \textit{reliability}, ensuring the model behavior is effectively updated for the current sample; 2) \textit{generalization}, so that the changes extend to neighboring samples; and 3) \textit{locality}, meaning the edit should have minimal impact on the model behavior on unrelated samples.

Model editing has allowed many fascinating applications, including error correction, factual knowledge update, bias mitigation, policy compliance, and personalization, though most of them have predominantly been within large language models (LLMs)~\cite{floridi2020gpt, achiam2023gpt, dubey2024llama} in the 
natural language processing (NLP) community~\cite{de2021editing, mitchell2022fast}.
With the enormous and often inaccessible pre-training datasets and the ever-growing model sizes that make retraining computationally demanding,
the need for effectively editing computer vision (CV) models is also becoming urgent. 
Adapting model editing techniques from NLP to CV is non-trivial and presents unique challenges. From the data perspective, NLP deals with one-dimensional, discrete signals that are highly semantic and information-dense, whereas CV requires processing high-dimensional continuous sensor data that is spatially redundant. From the model perspective, lots of model editing methods in NLP are specially designed for LLMs with \textit{unidirectional} (\ie, autoregressive) attention, such as GPT-3~\cite{floridi2020gpt} and GPT-4~\cite{achiam2023gpt}. In contrast, CV models have primarily been based on convolutional networks~\cite{krizhevsky2012imagenet, simonyan2014very, he2016deep}, with more recent implementations using vision Transformers (ViTs)~\cite{dosovitskiy2021image, liu2021swin} that otherwise employ \textit{bidirectional} attention. These differences in data formats and model structures make targeted edits more challenging to implement in CV models, and when such edits are achieved, they often result in suboptimal performance.

In this paper, we take initial steps towards editing pre-trained ViTs for object recognition~\cite{deng2009imagenet}, aiming to correct predictive errors without the need for costly and time-consuming retraining. Specifically, we take a locate-then-edit approach, which breaks down
model editing into two key subproblems: where-to-edit and how-to-edit. Moreover, we prioritize learning where to edit rather than how to edit to facilitate a simpler yet better trade-off between generalization and locality, without needing to store previously trained data. 

For the where-to-edit phase, we first narrow the editing scope using a greedy search-based heuristic. Next, inspired by the proven effectiveness of meta-learning~\cite {finn2017model} in optimizing training strategies for individual samples, we meta-train a hypernetwork to generate a binary task, indicating which parameters are critical for the editing sample. 
To address the issue of limited data, the hypernetwork is trained solely using pseudo-samples, each comprising a natural image paired with its CutMix version~\cite{yun2019cutmix} (see Fig.~\ref{fig:method_overview}). The optimization objective is to align the predicted probability distribution of the CutMix sample to that of the original. By controlling the sizes of patches used in CutMix and randomly varying their locations, we simulate distribution shifts in backgrounds, contextual objects, and object attributes, creating opportunities to learn generalizable binary masks that effectively respond to real-world failures. Additionally, we apply a sparsity constraint to the binary masks, acting as an indirect, data-free regularizer to promote locality. Once the where-to-edit problem is solved,  the how-to-edit phase becomes straightforward: we simply fine-tune the selected parameters using a variant of gradient descent to apply targeted edits.

To validate our method, we construct an editing benchmark that exposes the weaknesses of pre-trained ViTs by introducing two types of subpopulation shifts. 
The first is a natural subpopulation shift~\cite{recht2019imagenet,santurkar2021breeds}, 
with underrepresented natural images of certain categories efficiently identified by the maximum discrepancy (MAD) competition~\cite{wang2020going}. The second is an artificial subpopulation shift, introduced by synthesized images from high-quality text-to-image generative models like Stable Diffusion~\cite{rombach2022high}.

In summary, our key contributions are as follows:
\begin{itemize}[topsep=0pt,itemsep=0pt,leftmargin=20pt]
    \item[$\bullet$] a first-of-its-kind model editing method for pre-trained ViTs that leverages meta-learning to prioritize the where-to-edit phase;
    \item[$\bullet$] an editing benchmark that provides valuable resources for future model editing research in CV; 
    \item[$\bullet$] an extensive experimental demonstration that our method achieves the best Pareto front between generalization and locality on the proposed benchmark, while offering flexible trade-offs in the how-to-edit phase.
\end{itemize}

\section{Related Work}
In this section, we provide a brief overview of current model editing methods in NLP and CV.

\subsection{Model Editing in NLP}
\noindent\textbf{Memory-based Methods} rely on external mechanisms, such as wrappers~\cite{mitchell2022memory} and caches~\cite{hartvigsen2023aging}, to store factual updates without modifying the internal model parameters. A common theme in these studies is the use of a gating mechanism to determine whether a test sample falls within the editing scope; if so, the base model behavior is overridden. For instance, SERAC~\cite{mitchell2022memory} and GRACE~\cite{hartvigsen2023aging} employ a scope classifier as a form of hard gating, while Murty~\etal~\cite{murty2022fixing} utilized a soft gating function, allowing for smoother integration. More recent approaches like IKE~\cite{zheng2023can} and MeLLo~\cite{zhong2023mquake} alter the input prompts of an LLM for knowledge update, where the gating mechanism is implicitly embedded within the LLM itself. Generally, memory-based methods offer advantages such as non-destructive updates, modularity, and suitability for continual and few-shot learning settings. However, they face scalability issues when handling a large number of edits. Additionally, the editing success heavily depends on the accuracy of the gating mechanism.

\noindent\textbf{Parameter-based Methods} modify the internal model parameters, which offers a more fine-grained approach to editing. These methods can roughly be categorized into two subgroups: locate-then-edit approaches and hypernetwork-based approaches. Locate-then-edit methods focus on identifying a subset of key parameters for editing. For instance, ROME~\cite{meng2022locating}, MEMIT~\cite{meng2023mass}, and MEMIT$_{\textsc{CSK}}$~\cite{gupta2023editing} leverage causal mediation analysis (\ie, representation denoising) to locate hidden states (\ie, intermediate representations, not model parameters) responsible for knowledge storage. The theory of associative memory~\cite{kohonen1972correlation} is then applied to transfer the state localization results to model parameters. Recent studies~\cite{hase2023does} suggest that knowledge localization may not reliably inform successful edits. Furthermore, the very notion that knowledge can be localized may be inherently flawed, as factual information in LLMs may be encoded in a distributed manner~\cite{mitchell2022fast}. \textit{Single-step} integrated gradient across multiple editing samples~\cite{dai2022knowledge, wu2023depn} is another commonly used statistic for localization. Here, we adopt a more principled meta-learning strategy to locate key parameters, using \textit{multi-step} gradient information that more accurately captures the changes in model behavior.

Hypernetwork-based methods, such as KnowledgeEditor~\cite{de2021editing}, MEND~\cite{mitchell2022fast}, and MALMEN~\cite{tan2023massive}, train an external network to directly generate parameter updates for the editing sample, which is represented by either feedforward feature representation~\cite{de2021editing} or backward gradient decomposition~\cite{mitchell2022fast}. Localization techniques can be applied beforehand to restrict the functional space of the hypernetwork. Existing hypernetwork-based methods emphasize the how-to-edit aspect but treat the where-to-edit superficially, and often result in suboptimal performance, especially when adapting to CV applications. In contrast, our method prioritizes learning where to edit, achieving a better balance between generalization and locality.

\subsection{Model Editing in CV}
Limited research on model editing has been conducted in CV. Bau~\etal~\cite{bau2020rewriting} took a locate-then-edit approach to rewrite generative adversarial networks. Santurkar~\etal~\cite{santurkar2021editing} adapted this method for editing image classifiers based on convolutional networks by mapping the representation of the new visual concept to that of a previously learned concept. However, this approach requires prior knowledge of the new visual concept, its location within the image, and the specific target concept for correction. In practical applications, such detailed information may not always be available. In contrast, our method relaxes all these assumptions and is one of the first applied to ViTs.

\section{Learning Where to Edit ViTs}\label{methods}
\begin{figure*}[t]
\centering
\includegraphics[width=0.9\textwidth]{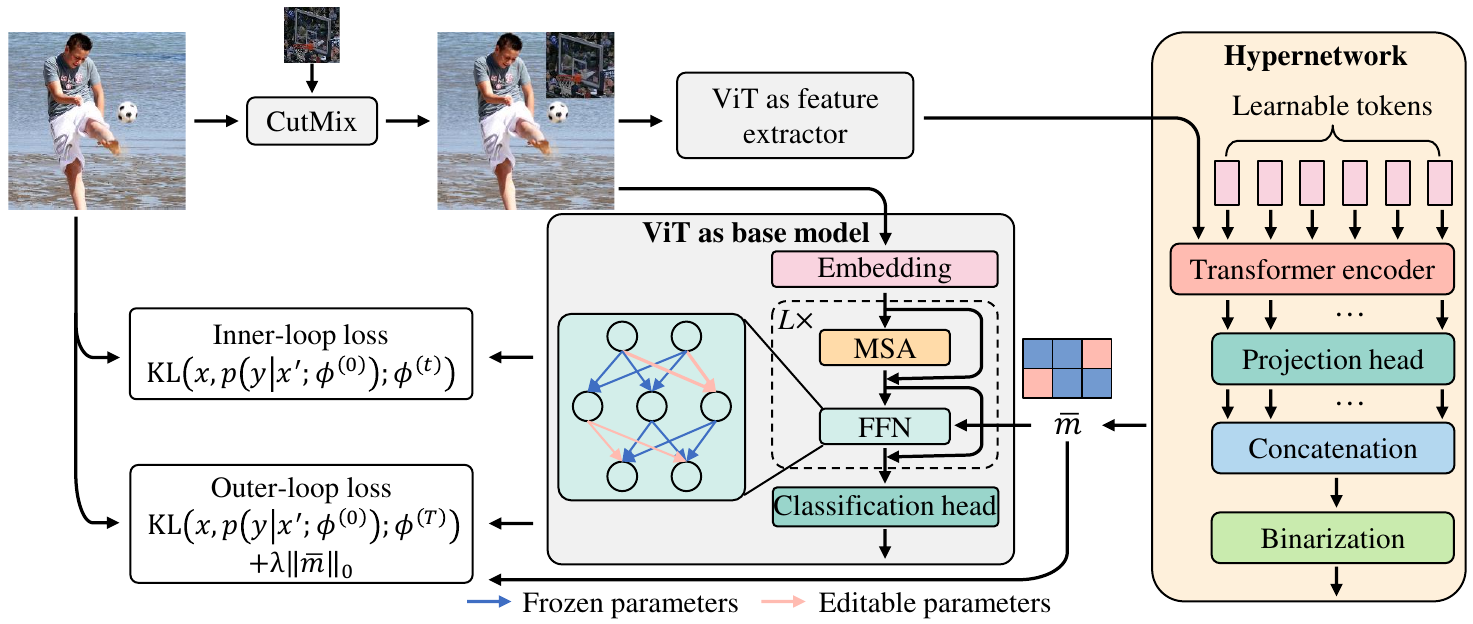}
\caption{System diagram of the proposed model editing method.}
\label{fig:method_overview}
\end{figure*}

In this section, we first present the preliminaries, followed by a detailed description of the proposed method for learning where to edit ViTs. The system diagram of our method is shown in Fig.~\ref{fig:method_overview}.

\subsection{Preliminaries}
\paragraph{Problem Formulation} Given a base 
computational model $f(\cdot;\theta):\mathcal{X} \mapsto \mathcal{Y}$, parameterized by $ \theta$, 
model editing aims to modify the model behavior for specific inputs $ x\in \mathcal{X}$ (or regions of the input space, $\mathcal{S}\subset \mathcal{X}$) while keeping its overall performance intact. Denote the post-edited model as $f\left(\cdot;{\theta}^{(e)}\right)$, where ${\theta}^{(e)}$ represents the updated parameter vector\footnote{We slightly abuse the notation ${ \theta}^{(e)}$ to encompass any possible modifications, including those by memory-based methods.} after editing. 
Typically, $f\left(\cdot;\theta^{(e)}\right)$ is evaluated based on three main criteria: reliability, generalization, and locality. 
\begin{itemize}
    \item \textbf{Reliability}: For any editing sample  $( x,y)$, the edited model $f\left(x;\theta^{(e)}\right) = y$.
    \item \textbf{Generalization}: For any neighboring\footnote{Conceptually, in high-level vision, two images are considered neighbors, if they are semantically similar, such as belonging to the same category or subpopulation.} sample $(x',y')\in \mathcal{N}(x,y)$, $f\left(x';\theta^{(e)}\right) = y'$, even if $(x',y')$ is not directly used in the editing process.
    \item \textbf{Locality}: For any sample $ (x',y') \notin \mathcal{N}(x,y)$, the model behavior should remain unchanged, \ie, $f
    \left( x';\theta^{(e)}\right) = f( x';\theta)$.
\end{itemize}
 An ideal model editing method shall ensure reliable edits while balancing generalization and locality effectively. As initial model editing attempts in CV, we limit our scope to single-example editing.

\paragraph{Vision Transformers}  A ViT~\cite{dosovitskiy2021image} feature extractor, denoted by $ e(\cdot;\phi)$ with parameter vector $ \phi$, consists of a linear embedding layer followed by $L$ attention blocks. Each block is composed of 
a multiheaded self-attention (MSA) layer and a feedforward neural network (FFN). The FFN, which underpins most model editing methods, including ours, comprises
two  fully-connected (FC) layers: 
${ \mathrm{FFN}( z) = \mathrm{GELU}( z W +  b){W}' +  {b}'}$. Here, ${W \in \mathbb{R}^{N \times N_m}} $ and ${{W'} \in \mathbb{R}^{N_m \times N}}$ are weight matrices, where $N_m$ denotes the intermediate dimension.  ${ b \in \mathbb{R}^{N_m \times 1}}$ and ${{b'} \in \mathbb{R}^{N\times 1}}$ are  bias terms. The activation function $\mathrm{GELU}(\cdot)$ is the Gaussian error linear unit~\cite{hendrycks2016gaussian}.  

An input image $ x$ is first partitioned into $M$ non-overlapping,  fixed-size patches, each linearly embedded in an $N$-dimensional feature space together with a class token \texttt{[cls]}, 
yielding a concatenation of patch embeddings of size $(M+1)\times N$.
These embeddings are processed through the $L$ attention blocks for feature extraction.
A linear classification head, $h(\cdot)$, maps the extracted features to a probability distribution over classes in $\mathcal{Y}$, represented as 
${ p(y=c| {x};\phi) = h_c( e(x;\phi)})$, where $c\in\mathcal{Y}$. For notation simplicity, we omit the parameters in the classification head $h(\cdot)$, as they constitute only a small fraction of the total parameters and are generally frozen during model editing.

\subsection{Model Editing at Training Time: Where-to-edit}
The simplest way of editing a ViT is through vanilla fine-tuning, which involves updating all model parameters. However, modern ViTs have millions to billions of parameters, and fine-tuning on a single sample $(x,y)$ can lead to overfitting, while incurring substantial computation costs.
To overcome these, prior research~\cite{dai2022knowledge, he2023sensitivity} first identifies a subset of key parameters, followed by editing:
\begin{equation}
    {\phi}^{\star} = { \phi} + \bar{m} \odot \Delta{\phi}, 
\end{equation}
where $\bar{m}$ is a binary mask of the same dimension as ${\phi}$, $\Delta\phi$ represents the parameter update, and
$\odot$ is the Hadamard product.

Prevailing localization strategies in NLP rely on casual mediation analysis~\cite{meng2022locating}, integrated gradients~\cite{dai2022knowledge}, or pure heuristic methods~\cite{hu2022lora}, which may not be ideal for ViTs due to differences in data modalities and model architectures. In this work, we follow the locate-the-edit approach, and decompose model editing into two subproblems: where-to-edit (\ie, computing $\bar{m}$) and how-to-edit (\ie, computing $\Delta \phi$), with a focus on where-to-edit. Drawing inspiration from the demonstrated success of meta-learning~\cite{lake2015human, finn2017model} in tailoring training strategies for individual samples, we meta-train a hypernetwork to generate the binary mask $\bar{m}$ for each editing sample.


Meta-learning~\cite{lake2015human, finn2017model}, also known as learning-to-learn, involves training models on a collection of training episodes~\cite{chen2020closer} to enable effective generalization and adaptation to novel, unseen episodes. In our context, a training episode corresponds to a single editing example. We employ optimization-based meta-learning approaches~\cite{finn2017model, rajeswaran2019meta}, framing where-to edit as a bi-level optimization problem. In the inner loop, key parameters, indicated by $\bar{m}$, are updated for the editing sample by optimizing a reliability loss via gradient-descent over $T$ iterations. In the outer loop, the hypernetwork $g(\cdot;\varphi)$, parameterized by $\varphi$, is refined to generate $\bar{m}$. Mathematically, we have
\begin{equation}
\begin{aligned}
    \label{eq:outer-loops} 
    \min_{\varphi} \quad & \ell\left({x},  {y};\phi^{(T)}\right)+ \lambda\Vert  \bar{m}\Vert_0 \\
    \mathrm{s.t.} \quad 
    & \bar{m} = g(x;\varphi)\\
    & \Delta{ \phi}^{(t)}  = \Delta{ \phi}^{(t-1)} - \alpha \nabla_{ \phi} \ell\left( {x}, {y};  \phi^{(t-1)}\right), \, t \in \{1, 2, \ldots,T\} \\
     & { \phi}^{(t)} = { \phi}^{(0)} + \bar{m} \odot \Delta{\phi}^{(t)},\, t\in\{1, 2, \ldots,T\},\\
\end{aligned}
\end{equation}
where $(x,y)$ is the editing sample. $ \phi^{(T)}$ is the updated parameter after $T$ iterations of inner-loop optimization, and $\phi^{(0)}$ denotes the pre-trained parameters of the base model as initialization. The term $\Delta \phi^{(t)}$ is the parameter update after the $t$-th iteration, with $\Delta \phi^{(0)} = 0$. The loss function $\ell\left(x,y;\phi^{(t)}\right)$ measures the reliability of the edit. To encourage sparsity in the binary mask $\bar{m}$, we add an $\ell_0$-norm  term in the outer-loop objective, which acts as an indirect, data-free regularizer to encourage locality. The scalar $\lambda$ controls the trade-off between the two terms. In our implementation, the hypernetwork takes the last-stage features corresponding to the \texttt{[cls]} token from the ViT feature extractor $e\left(\cdot;\phi^{(0)}\right)$ as input, \ie, $\bar{m} = g\left(e\left(x;\phi^{(0)}\right);\varphi\right)$.

\subsection{Optimization Challenges}
Despite mathematical elegance, solving the bi-level optimization problem in~\eqref{eq:outer-loops} presents three challenges. First, meta-training the hypernetwork necessitates a sizable of high-quality editing samples, which are expensive and time-consuming to collect in practice.
To address this, we generate pseudo-samples using a data augmentation technique known as CutMix~\cite{yun2019cutmix}. Second, identifying key parameters within the entirety of the ViT presents a vast search space. This combinatorial complexity not only introduces unacceptable computational costs but also makes the localization of key parameters a challenging endeavor~\cite{liu2019metapruning,shang2022neural}. To alleviate this, we shrink the editing scope based on a greedy search-based heuristic.
Third, generating a binary mask typically involves a binarization operation in $g(\cdot;\varphi)$, which produces zero gradients almost everywhere and is thus ineffective in optimizing. To resolve this,  we use a gradient-friendly approximation to binarization.

\noindent\textbf{Pseudo-sample Generation}
We employ CutMix~\cite{yun2019cutmix} to generate pseudo-samples for editing. 
Specifically, given a natural image ${x'}$, we apply CutMix~\cite{yun2019cutmix} to randomly overlay a small patch from another irrelevant image onto ${x'}$, producing a pseudo-sample $x$. This patch-based perturbation tends to alter the predicted probability distribution, resulting in $p\left(y=c|x;\phi^{(0)}\right) \ne p\left(y=c|{x'};\phi^{(0)}\right)$,  for $c\in \mathcal{Y}$. This motivates us to instantiate the reliability loss  $\ell\left(x, y; \phi^{(t)}\right)$ in Problem~\eqref{eq:outer-loops} as the Kullback-Leibler (KL) divergence~\cite{hinton2015distilling} between $p\left(y|{x'};\phi^{(0)}\right)$ and $p\left(y|x;\phi^{(t)}\right)$:
\begin{equation}
     \ell \left(x, \left\{p\left(y| {x'};\phi^{(0)}\right)\right\}; {\phi}^{(t)}\right) = \sum_{c\in\mathcal{Y}} p\left(y=c| {x'};\phi^{(0)}\right) \log\left(\frac{p\left(y=c| {x'};\phi^{(0)}\right)}{p\left(y=c| {x};\phi^{(t)}\right)}\right),
\end{equation}
where $\left\{p\left(y|{x'};\phi^{(0)}\right)\right\}$ is treated as the soft ground-truth label.

\begin{figure} 
    \centering
        \begin{subfigure}[]{0.33\textwidth}
        \includegraphics[width = 0.92\columnwidth]{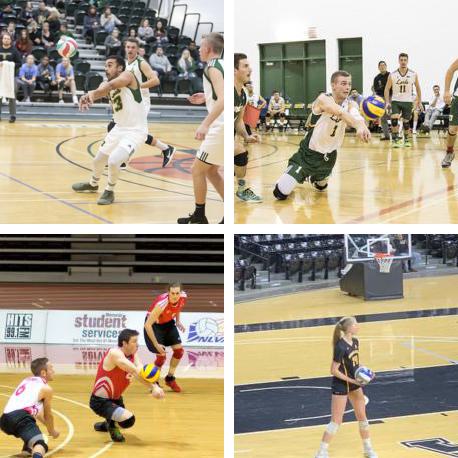}
        \caption{Editing samples.}
        \label{fig:Volleyball_samples}
    \end{subfigure}
    \hfill
    \centering
    \begin{subfigure}[]{0.66\textwidth}
        \includegraphics[width = \columnwidth]{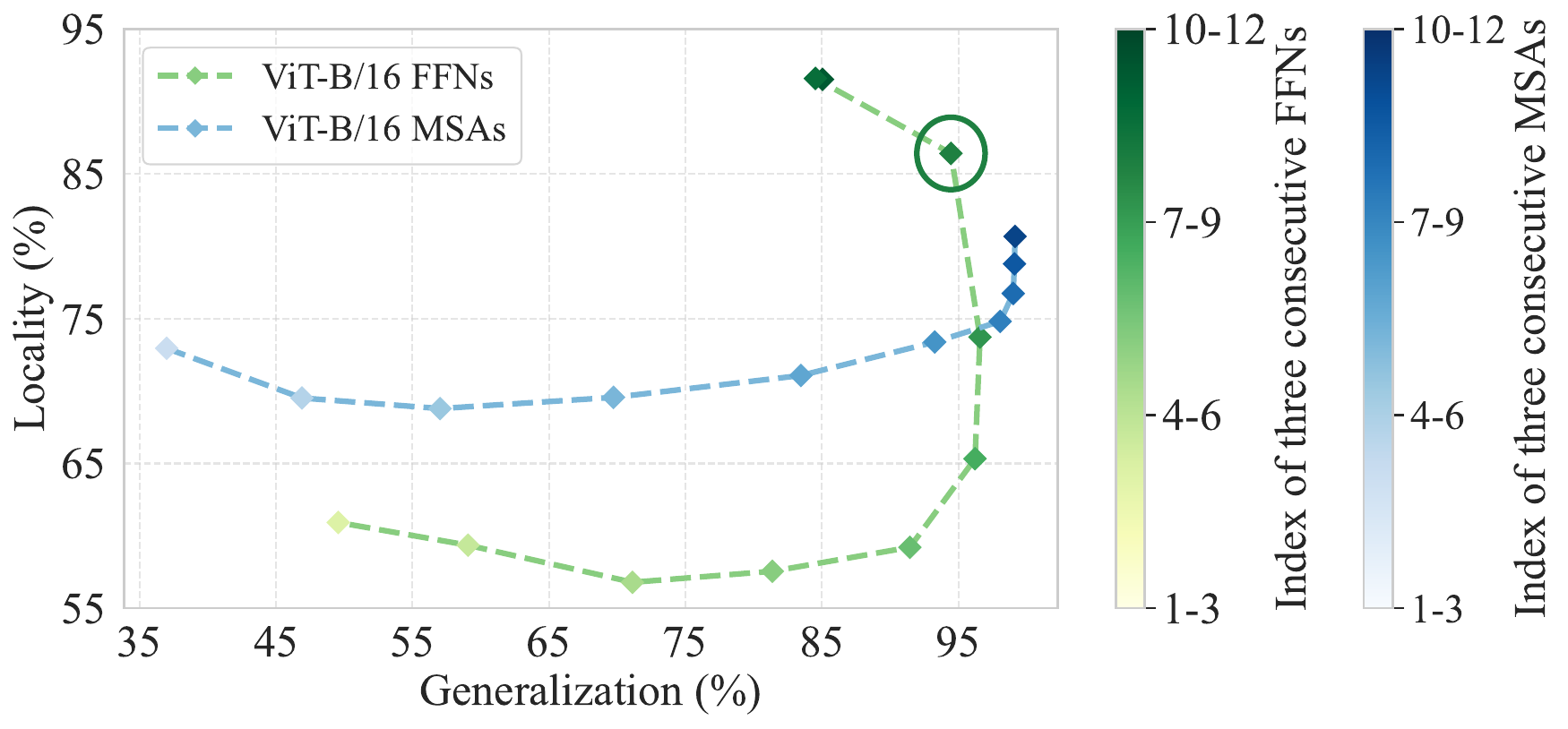}
        \caption{Editing results via vanilla fine-tuning.}
        \label{fig:shrinkage}
    \end{subfigure}    
    \caption{The left subfigure shows representative editing examples, highlighting the predictive errors of the base ViT when predicting \texttt{volleyball} as \texttt{basketball}. The right subfigure depicts the generalization and locality trade-offs when editing different groups of FFNs or MSAs in the base ViT. It is evident that editing the $8$-th to $10$-th FFNs achieves the optimal Pareto front.}
        \label{fig:edit_diff_layers}
\end{figure}

\noindent\textbf{Editing Scope Shrinkage} Previous studies~\cite{meng2022locating, mitchell2022fast} have suggested that modifying FFNs within a Transformer is more effective for achieving successful edits~\cite{geva2022transformer, geva2021transformer}.  For example, MEND\cite{mitchell2022fast} focuses on editing the last three FFNs, while ROME~\cite{meng2022locating} targets the middle FFNs.
Here, we conduct a similar empirical investigation to 
identify a subset of consecutive FFNs in a ViT, by greedy search for the optimal generalization and locality trade-off. 
Specifically, we fine-tune ten groups of FFNs (or MSAs) in three consecutive layers~\cite{mitchell2022fast} of a pre-trained ViT/B-16, denoted as \{$1$-$3$, $2$-$4$, $\ldots$, $10$-$12$\}. The editing set comprises $100$ predictive failures of the ViT, where \texttt{volleyball} is mistaken for \texttt{basketball} (see Fig.~\ref{fig:Volleyball_samples}), identified by the MAD competition~\cite{wang2020going} (see more details in Sec.~\ref{sec:construction_natural}). The average results across the editing set are shown in Fig.~\ref{fig:shrinkage}, where we see that editing MSAs is not conducive to preserving locality. In contrast, editing the $8$-th to $10$-th FNNs tends to achieve the best trade-off, which are selected as the default layers for subsequent experiments.

To further limit the output space of the hypernetwork, we employ structured tuning~\cite{dai2022knowledge} by selecting specific rows/columns of the weight matrices in the FFNs for updating. As suggested in~\cite{dai2022knowledge}, we select the weights along the intermediate dimension $N_m$, which further reduces the output dimension of the hypernetwork to $ N_m \times 6$ (\ie, three FFNs with two FCs each).

\noindent\textbf{Binarization Approximation} As a special case of quantization in signal processing, binarization can be approximated to enable gradient-based training through three main approaches: straight-through estimation~\cite{bengio2013estimating}, uniform noise addition~\cite{balle2017end}, and soft-to-hard annealing~\cite{jang1999binarization}. Here, we use a fixed parametric sigmoid function with favorable gradient behavior as the approximation:
\begin{equation}\label{eq:qa}
    \hat{m} = \mathrm{Sigmoid}(k\times m) , 
\end{equation}
where $m$ is a continuous map computed by the hypernetwork right before binarization, and
$k$ is a hyperparameter that controls the degree to which the sigmoid curve approximates the desired binarization operation. Empirically, we set $k=10$. We have also experimented with a soft-to-hard annealing for $k$, and observed comparable results. After adopting Eq.~\eqref{eq:qa}, we substitute $\bar{m}$ with $\hat{m}$ and replace the $\ell_0$-norm with the $\ell_1$-norm in Problem~\eqref{eq:outer-loops} to facilitate gradient-based optimization.

\subsection{Model Editing at Test Time: How-to-edit} 
At test time, we solve the how-to-edit problem in a manner similar to the inner-loop optimization. The two minor differences lie in the loss function and the binarization operation. 

At test time, we are provided with the editing sample $x$ and its ground-truth label $y$. Therefore, the KL divergence during training reduces the cross-entropy loss during testing: 
\begin{equation}
\ell\left(x,y;\phi^{(t)}\right)= -\sum_{c\in \mathcal{Y}} \mathbb{I}[y=c]\log\left(p\left(y=c|x;\phi^{(t)}\right)\right).
\end{equation}
Also, we can directly employ the threshold-based binarization without approximation to obtain
\begin{equation}\label{eq:hq}
\bar{m}_i = q(m_i)= \begin{cases} 
      1 & m_i \ge \rho \\
      0 &  m_i <\rho, 
   \end{cases}
\end{equation}
where $i$ is the positional index, and $\rho$ is a hyperparameter that can be adjusted for different model editing applications. When $\rho$ is set to zero, all parameters in the selected FFNs are updated with improved reliability. As $\rho$ increases, fewer parameters are updated, which favors locality.

\subsection{Hypernetwork Architecture}

Similar to the ViT feature extractor $e\left(\cdot; \phi^{(0)}\right)$, the hypernetwork $g(\cdot;\varphi)$  comprises five attention blocks, an FC layer as the projection head, and a binarization operation.  As shown in  Fig.~\ref{fig:method_overview},  we introduce six learnable tokens, each corresponding to an FC layer within the three selected FFNs of the base ViT. These tokens are concatenated with the image features derived from $e\left(\cdot;\phi^{(0)}\right)$ and serve as input to the hypernetwork to compute the binary mask $\bar{m}$. 

\section{Editing Benchmark with Subpopulation Shifts}\label{dataset_construction}
\begin{figure}
    \centering
        \includegraphics[width = 0.95\columnwidth]{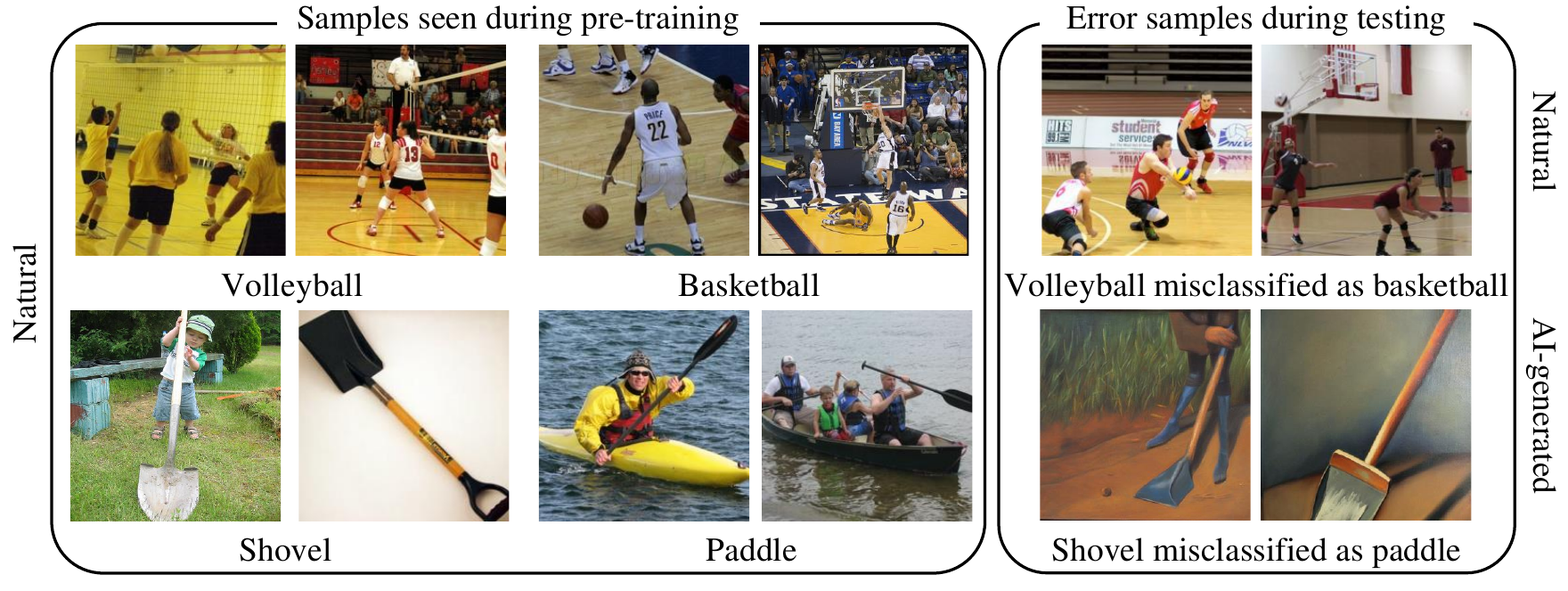}
        \caption{Visual examples seen by the base ViT/B-16 during pre-training, contrasted with visual examples in the proposed editing benchmark as predictive errors of the base ViT/B-16.}
        \label{fig:edting_example}
\end{figure}
In this section, we establish an editing benchmark that exposes failures of the base ViT in object recognition by introducing subpopulation shifts to underrepresented natural and AI-generated images.
\subsection{Natural Image Subset}\label{sec:construction_natural}
To build the natural image subset, we first compile a large dataset of unlabeled images, denoted as $\mathcal{U}$, from Flickr, by leveraging keywords relevant to the object categories in ImageNet-1k~\cite{deng2009imagenet}. Next, we employ the MAD competition~\cite{wang2020going} to facilitate failure identification of the base ViT to be edited. Under the principle of model falsification as model comparison, MAD chooses to identify images that best distinguish two classifiers, $f(\cdot)$ and $f'(\cdot)$, by maximizing their prediction discrepancies. This can be mathematically formulated as 
\begin{align}
    {x}^{(i)} = \argmax_{x'\in \mathcal{U}\setminus\mathcal{D}_n} d\left(f(x'), f'(x')\right),
\end{align}
where $\mathcal{D}_n = \{{x}^{(j)}\}_{j = 1}^{i-1}$ is the set of $i-1$ images that have been identified. $d(\cdot, \cdot)$ is the multi-hop distance defined over the WordNet~\cite{fellbaum1998wordnet} to measure prediction discrepancy at a semantic level. Intuitively, if one classifier is weaker, the identified image set $\mathcal{D}_n$ is more likely to include its predictive failures, thereby substantially reducing the human effort for failure identification. Moreover, the ``ground-truth'' labels for these failures can be first suggested by the stronger model and then verified by two of the authors. To leverage this intuition, we pair our base model (\ie, a ViT/B-16 pre-trained on ImageNet-1k) with a stronger one (\ie, the same ViT/B-16 pre-trained using CLIP~\cite{radford2021learning} and fine-tuned on ImageNet), which generally exhibits better generalization to unseen data. In total, we collect $2,354$ MAD-searched natural images, which are partitioned into $16$ groups, \ie, $\mathcal{D}_n = \{\mathcal{S}^{(i)}\}_{i=1}^{16}$, based on the predictions by the two models. Each group is named according to the format ``prediction of the stronger model''-``prediction of the base model,''  with the statistics and visual examples given in the Appendix.

\subsection{AI-generated Image Subset}
Classifiers pre-trained on natural images often struggle to generalize to AI-generated images~\cite{vendrow2023dataset, wiles2022discovering}. To exploit this, we construct an AI-generated image subset containing two groups of images, denoted as $\mathcal{D}_a=\{\mathcal{S}^{(i)}\}_{i=17}^{18}$. The $17$-th group includes $860$ images with an art style shift (\ie, oil painting) generated by Textural Inversion~\cite{vendrow2023dataset}, while the $18$-th group comprises   $1,092$ images with a lighting condition shift (\ie, stage light)  produced by PUG~\cite{bordes2024pug}. Both Textural Inversion and PUG are text-to-image generators, wherein the ``ground-truth'' label is embedded in the input text prompt and subsequently verified by two of the authors.
Additional details of the AI-generated image subset can be found in the Appendix.

\section{Experiments}\label{Experiments}
In this section, we first describe the experimental
setups 
and then present comparison results on the proposed editing benchmark.

\begin{figure}[tbp]
        \centering
        \includegraphics[width = \columnwidth]{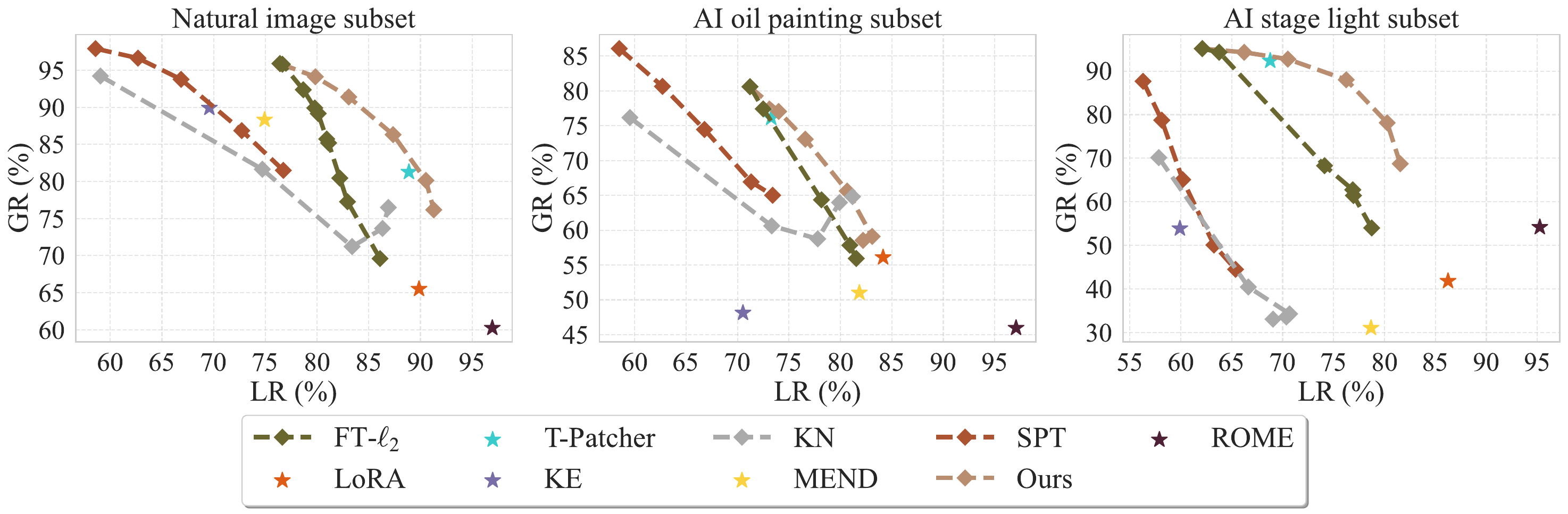}
        \caption{Editing results for ViT/B-16 on the proposed benchmark.}
        \label{fig:exp_curves}
\end{figure}

\subsection{Experiment Setups}\label{sec:exp_setup}
\paragraph{Evaluation Metrics}
Following~\cite{huang2023transformer}, we evaluate all model editing methods on the single-example editing task and compare their performance using three evaluation metrics. The first is the \textit{success rate} (SR), which indicates the reliability (\ie, accuracy) of the edited model $f\left(\cdot;\theta^{(e)}\right)$:
\begin{equation}
\mathrm{SR}(f,\mathcal{D}_r) = \frac{1}{\vert\mathcal{D}_r\vert}\sum_{(x,y) \in \mathcal{D}_r}\mathbbm{I} \left[y = f\left(x;\theta^{(e)}(x,y)\right)\right], \\
\end{equation}
where $\mathcal{D}_r = \mathcal{D}_n \bigcup \mathcal{D}_a$ consists of all MAD-searched and AI-generated images, and we make it explicit the dependence of the updated parameters $\theta^{(e)}$ on the editing sample $(x,y)$.
The second metric is the \textit{generalization rate} (GR), which assesses the accuracy of the edited model on neighboring samples that fall within the editing scope: 
\begin{equation}
\mathrm{GR}(f,\mathcal{S}) = \frac{1}{\vert\mathcal{S}\vert(\vert\mathcal{S}\vert-1)}\sum_{(x',y')\in\mathcal{S}}\sum_{(x,y) \in \mathcal{S}\setminus(x',y')}\mathbbm{I} \left[y = f\left(x;\theta^{(e)}(x',y')\right)\right], \\
\end{equation}
where $\mathcal{S}$ denotes one of the $18$ groups in the proposed editing benchmark. We further average the GR values across all groups as an overall indicator of generalization.
The third metric is  the \textit{locality rate} (LR), which examines whether the edited model maintains its predictions on unrelated samples outside the editing scope:
\begin{equation}
\mathrm{LR}(f,\mathcal{D}_r, \mathcal{D}_l) = \frac{1}{\vert\mathcal{D}_r\vert\vert\mathcal{D}_l\vert}\sum_{(x',y')\in\mathcal{D}_r}\sum_{(x,y) \in \mathcal{D}_l}\mathbbm{I} \left[y = f\left(x;\theta^{(e)}(x',y')\right)\right], \\
\end{equation}
where $\mathcal{D}_l$ includes out-of-scope images. Using the validation set from ImageNet-1k as $\mathcal{D}_l$ does not adequately examine locality, as the majority are easy samples that lie far from the decision boundary~\cite{gao2023adaptive}. To more closely examine the adverse effects of model editing, we have carefully curated $2,071$ images near the decision boundary of the base model from the validation sets of ImageNet-1k~\cite{russakovsky2015imagenet}, ImageNet-R~\cite{hendrycks2021many}, and ImageNet-Sketch~\cite{wang2019learning}, whose predictions are more susceptible to change. 
Our selection criteria rely on the 
 predicted probabilities of the pre-trained ViT/B-16 model as follows: 1) the predicted probability for the true label is the highest, and 2) the difference between the top two predicted probabilities is less than $0.05$, suggesting a highly ambiguous class. We also employ the GR-LR curve to delineate the generalization and locality trade-off.

\paragraph{Base Models} For all model editing methods, we experiment with two ViT backbones, ViT-B/16 and ViT/S-16, both pre-trained on ImageNet-21k and ImageNet-1k~\cite{steiner2022how, russakovsky2015imagenet}.

\paragraph{Competing Methods} 
We compare our method with several recent model editing approaches as follows. 1) {Fine-tuning} (FT) updates the $8$-th to $10$-th FFNs, which have been identified as the most effective layers using greedy search (see Fig.~\ref{fig:edit_diff_layers}). 2) FT-$\ell_2$~\cite{meng2023mass} incorporates $\ell_2$-norm regularization during fine-tuning. 3) {T-Patcher}~\cite{huang2023transformer} adds and tunes a single neuron in the last FFN. 4) {KN}~\cite{dai2022knowledge} and 5) {SPT}~\cite{he2023sensitivity} select key parameters based on integrated gradient information. 6) {ROME}~\cite{meng2022locating} is implemented to adjust the second FC layer of the last FFN by solving a constrained least squares problem.  7) {LoRA}~\cite{hu2022lora} introduces trainable low-rank matrices to update the queries and values of all MSAs. 8) {KE}~\cite{de2021editing} and  9) {MEND}~\cite{mitchell2022fast} employ hypernetworks to generate parameter updates for the last three FFNs. In line with previous work~\cite{mitchell2022fast, meng2023mass}, early stopping is applied when the training loss drops below $0.01$ or the maximum of $100$ editing steps is reached. Detailed implementations of the competing methods and additional training configurations are provided in the Appendix.

\subsection{Main Results}


Fig.~\ref{fig:exp_curves} shows the GR-LR curves for different editing methods applied to ViT-B/16, averaged across $18$ groups in the proposed benchmark. We highlight several interesting observations. First, correcting a single predictive error is generally feasible, as evidenced by a nearly $100\%$ SR for most methods.
Second, achieving high levels of generalization and locality simultaneously proves to be a significant challenge. T-Patcher and ROME utilize previously seen data to maintain locality. Nevertheless, T-Patcher, which relies on an editing scope classifier, exhibits noticeable generalization variability across different editing samples. ROME, being specifically designed for language-based GPT~\cite{floridi2020gpt}, shows limited promise in generalizing to ViTs.  LoRA manages to maintain locality because of its low-rank updates but struggles to generalize. Both KE and MEND  exhibit low locality on the MAD-searched natural images and poor generalization to the AI-generated images. Third, our method achieves the new state-of-the-art without relying on previously trained data to explicitly enforce locality. Similar conclusions can be drawn for ViT-S/16, shown in the Appendix.

We then evaluate our method across different parameter sparsity levels in the three FFNs from $\{0.25, 0.50, 0.75, 0.90, 0.95 \}$, corresponding to $\{12.4\%, 8.25\%, 4.13\%, 1.65\%, 0.83\% \}$ parameters of the entire model, by adjusting $\rho$ in Eq.~\eqref{eq:hq}. The competing methods---FT-$\ell_2$, KN, and SPT---are adjusted to comparable levels of parameter sparsity by tuning their respective hyperparameters. Note that our method reduces to FT when $\rho = 0$. The resulting GR-LR curves are shown in Fig.~\ref{fig:exp_curves}. As expected, increasing the parameter sparsity in KN, SPT, and our method improves locality at the expense of generalization. Notably, our method achieves the best Pareto front among all methods, which we believe arises from our proposed strategy of learning where to edit towards editing success.

\begin{figure}[t]
    \centering
    \begin{subfigure}[]{0.49\columnwidth}
        \includegraphics[width = \columnwidth]{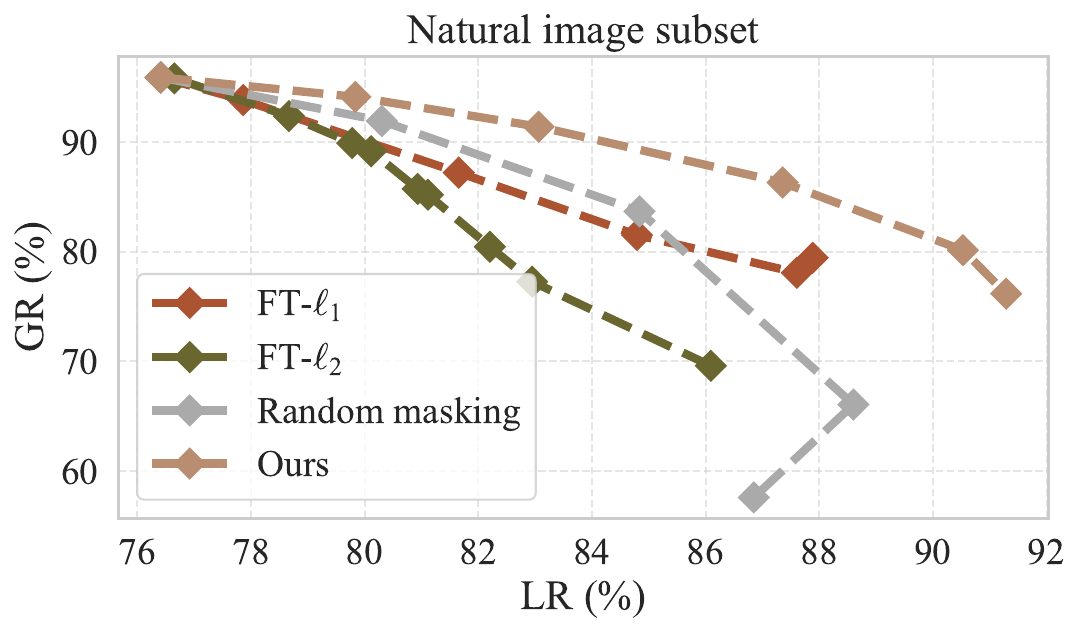}
        \caption{Localization effectiveness.}
        \label{fig:933_923_B_C2R}
    \end{subfigure}
    \hfill
    \centering
    \begin{subfigure}[]{0.49\columnwidth}
        \includegraphics[width = \columnwidth]{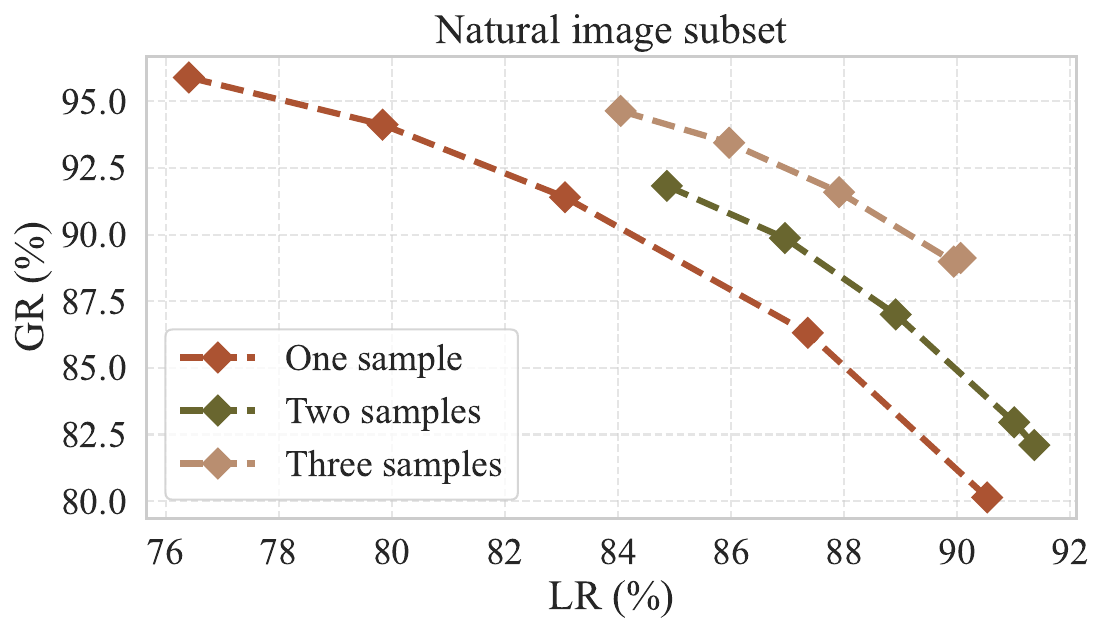}
        \caption{More editing samples.}
        \label{fig:exp_curves_samples}
    \end{subfigure}
    \caption{Ablation results of the hypernetwork for ViT/B-16.}
    \vspace{-6pt}
\end{figure}

\subsection{Ablation Studies}
\paragraph{Localization Effectiveness} 
To substantiate that the effectiveness of our method is indeed due to the successful localization of a specific subset of key parameters, rather than merely due to sparsity, we compare the binary masks produced by our hypernetwork to random masks at the same sparsity levels, together with FT-$\ell_1$ and FT-$\ell_2$. 
As depicted in  Fig.~\ref{fig:933_923_B_C2R}, FT-$\ell_1$ generally surpasses FT-$\ell_2$ at various regularization levels as $\ell_1$-norm is more effective in zeroing out less important parameters. Applying random masks shows effects akin to FT-$\ell_1$. When the ratio of editing parameters falls below $1.65\%$, the performance of random masking becomes significantly inferior to our method.

\begin{figure}[t]
    \centering
    \begin{subfigure}[]{0.48\textwidth}
        \includegraphics[width = \columnwidth]{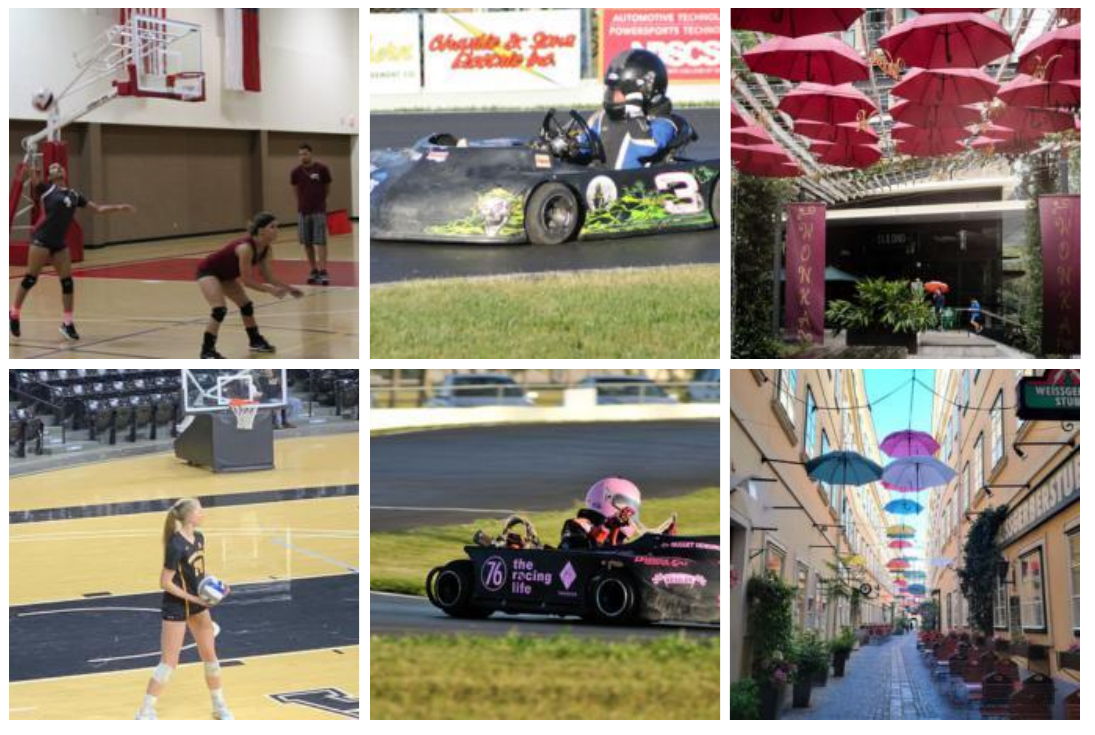}
        \caption{Six representative editing examples from three different groups.}
        \label{fig:example_figs}
    \end{subfigure}
    \hfill
    \centering
    \begin{subfigure}[]{0.48\textwidth}
        \includegraphics[width = \columnwidth]{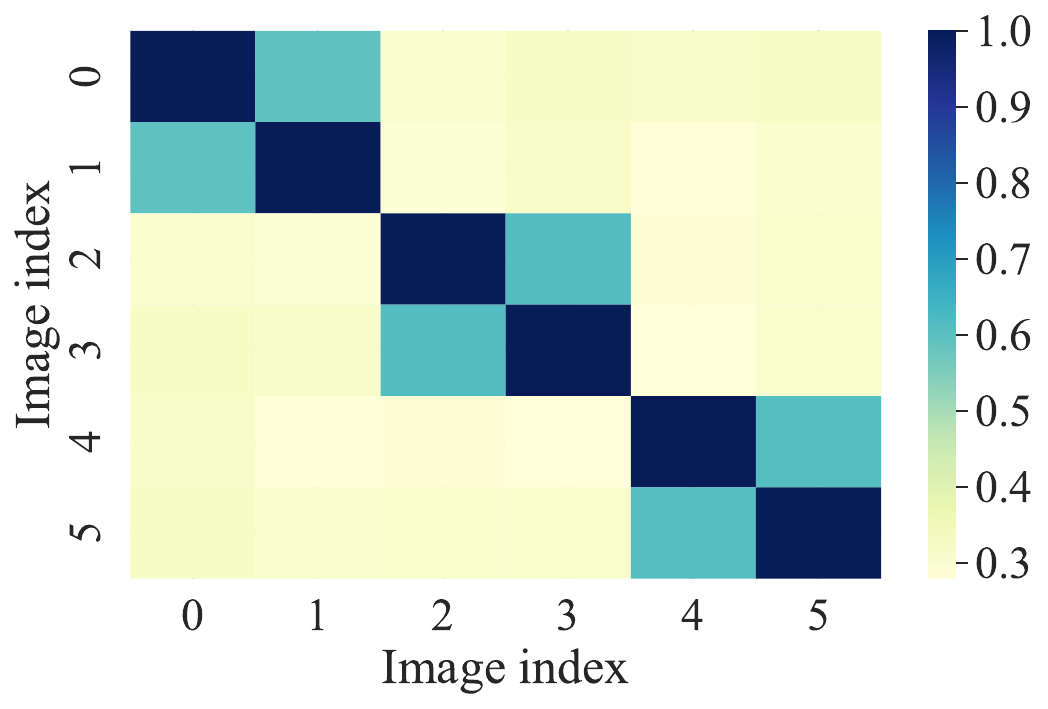}
        \caption{Binary mask IoU results between pairs of samples in (a), indexed in column-major order.}
        \label{fig:sample_masks_overlaps_B}
    \end{subfigure}
    \caption{Mask specificity results.}
    \vspace{-6pt}
\end{figure}

\paragraph{Mask Specificity} To confirm the specificity of the parameters identified by the hypernetwork for different editing samples, we compute the intersection over union (IoU) of the corresponding binary masks at the $0.95$ sparsity level for samples within and outside the same groups in the natural image subset.  Fig.~\ref{fig:sample_masks_overlaps_B} illustrates that the identified parameters demonstrate substantial overlaps for images within the same group and much lower overlaps between images from different groups. These findings support that the hypernetwork successfully pinpoints key parameters necessary to correct specific errors while effectively excluding parameters associated with other unrelated samples. This learned mask specificity allows our method to balance effectively between generalization and locality.

\paragraph{More Editing Samples} 
We further evaluate our method when multiple editing samples in the same group (\ie, with similar failure causes) are available. As a straightforward extension,  we compute the average of the continuous masks generated from each sample, followed by binarization using Eq.~\eqref{eq:hq}.  Fig.~\ref{fig:exp_curves_samples} presents the results of using one, two, and three samples for model editing. Remarkably, the editing performance improves with more editing samples, which can be attributed to more precise parameter localization as a result of the ensemble of masks.

\paragraph{More Ablation Studies} More ablation studies (\eg, the alternative pseudo-sample generation strategy, the sparsity regularization in the outer loop, the gradient step and learning rate in the inner loop, and the number of attention blocks in the hypernetwork) are in the Appendix.

\section{Conclusion and Discussion}
We have introduced a model editing method to correct predictive errors in ViTs. Our method prioritizes where-to-edit over how-to-edit by 
meta-training a hypernetwork to identify a subset of structured parameters for editing. By applying $\ell_1$-norm regularization, our method promotes sparsity in the generated mask, thereby indirectly ensuring locality without needing to retrain on previously used data. Comprehensive tests on the proposed editing benchmark confirm that our method effectively corrects predictive errors in ViTs. Moreover, the introduced edits are not only reliable but also generalize well to neighboring samples, while maintaining a high rate of locality.

Our work is among the early endeavors in CV model editing, and it raises several intriguing questions for future research. First, our approach utilizes the CutMix technique~\cite{yun2019cutmix} to generate cost-effective pseudo-samples for training, but its effectiveness has only been confirmed empirically. The reasons why the hypernetwork trained on such synthetic data achieves reasonable generalization and the identification of optimal synthetic data generation techniques remain wide open.
Second, it would be beneficial to adapt our method to other vision architectures, such as convolutional networks or Swin Transformers~\cite{liu2021swin}, and extend its application to other vision areas like dense prediction, generative modeling, and multimodal LLMs. Third, exploring how to apply our method in a batch-editing setting represents a promising avenue. In such scenarios, the use of a decoupling trick (see more details in the Appendix) may prove essential for effectively reducing computational and memory demands.

{
\bibliographystyle{plain}
\bibliography{neurips_2024}

\begin{thebibliography}{10}

\bibitem{achiam2023gpt}
Josh Achiam, Steven Adler, Sandhini Agarwal, Lama Ahmad, Ilge Akkaya, Florencia~L Aleman, Diogo Almeida, Janko Altenschmidt, Sam Altman, Shyamal Anadkat, et~al.
\newblock {GPT}-4 technical report.
\newblock {\em arXiv preprint arXiv:2303.08774}, 2023.

\bibitem{balle2017end}
Johannes Ball{\'e}, Valero Laparra, and Eero~P Simoncelli.
\newblock End-to-end optimized image compression.
\newblock In {\em International Conference on Learning Representations}, 2017.

\bibitem{bau2020rewriting}
David Bau, Steven Liu, Tongzhou Wang, Jun-Yan Zhu, and Antonio Torralba.
\newblock Rewriting a deep generative model.
\newblock In {\em European Conference on Computer Vision}, pages 351--369, 2020.

\bibitem{bengio2013estimating}
Yoshua Bengio, Nicholas L{\'e}onard, and Aaron Courville.
\newblock Estimating or propagating gradients through stochastic neurons for conditional computation.
\newblock {\em arXiv preprint arXiv:1308.3432}, 2013.

\bibitem{bordes2024pug}
Florian Bordes, Shashank Shekhar, Mark Ibrahim, Diane Bouchacourt, Pascal Vincent, and Ari Morcos.
\newblock {PUG}: Photorealistic and semantically controllable synthetic data for representation learning.
\newblock In {\em Advances in Neural Information Processing Systems}, pages 8072--8081, 2023.

\bibitem{buntine1991theory}
Wray Buntine.
\newblock Theory refinement on {Bayesian} networks.
\newblock In {\em Conference on Uncertainty in Artificial Intelligence}, pages 52--60, 1991.

\bibitem{chen2020closer}
Jiaxin Chen, Xiao-Ming Wu, Yanke Li, Qimai Li, Li-Ming Zhan, and Fu-lai Chung.
\newblock A closer look at the training strategy for modern meta-learning.
\newblock In {\em Advances in Neural Information Processing Systems}, pages 396--406, 2020.

\bibitem{dai2022knowledge}
Damai Dai, Li~Dong, Yaru Hao, Zhifang Sui, Baobao Chang, and Furu Wei.
\newblock Knowledge neurons in pretrained {Transformers}.
\newblock In {\em Annual Meeting of the Association for Computational Linguistics}, pages 8493--8502, 2022.

\bibitem{de2021editing}
Nicola De~Cao, Wilker Aziz, and Ivan Titov.
\newblock Editing factual knowledge in language models.
\newblock In {\em Empirical Methods in Natural Language Processing}, pages 6491--6506, 2021.

\bibitem{deng2009imagenet}
Jia Deng, Wei Dong, Richard Socher, Li-Jia Li, Kai Li, and Li~Fei-Fei.
\newblock {ImageNet}: A large-scale hierarchical image database.
\newblock In {\em IEEE Conference on Computer Vision and Pattern Recognition}, pages 248--255, 2009.

\bibitem{dosovitskiy2021image}
Alexey Dosovitskiy, Lucas Beyer, Alexander Kolesnikov, Dirk Weissenborn, Xiaohua Zhai, Thomas Unterthiner, Mostafa Dehghani, Matthias Minderer, Georg Heigold, Sylvain Gelly, Jakob Uszkoreit, and Neil Houlsby.
\newblock An image is worth 16x16 words: {Transformers} for image recognition at scale.
\newblock In {\em International Conference on Learning Representations}, 2021.

\bibitem{dubey2024llama}
Abhimanyu Dubey, Abhinav Jauhri, Abhinav Pandey, Abhishek Kadian, Ahmad Al-Dahle, Aiesha Letman, Akhil Mathur, Alan Schelten, Amy Yang, Angela Fan, et~al.
\newblock The {Llama} 3 herd of models.
\newblock {\em arXiv preprint arXiv:2407.21783}, 2024.

\bibitem{fellbaum1998wordnet}
Christiane Fellbaum.
\newblock {\em {WordNet: An Electronic Lexical Database}}.
\newblock MIT Press, 1998.

\bibitem{finn2017model}
Chelsea Finn, Pieter Abbeel, and Sergey Levine.
\newblock Model-agnostic meta-learning for fast adaptation of deep networks.
\newblock In {\em International Conference on Machine Learning}, pages 1126--1135, 2017.

\bibitem{floridi2020gpt}
Luciano Floridi and Massimo Chiriatti.
\newblock {GPT}-3: Its nature, scope, limits, and consequences.
\newblock {\em Minds and Machines}, 30:681--694, 2020.

\bibitem{gao2023adaptive}
Irena Gao, Gabriel Ilharco, Scott Lundberg, and Marco~T Ribeiro.
\newblock Adaptive testing of computer vision models.
\newblock In {\em IEEE International Conference on Computer Vision}, pages 4003--4014, 2023.

\bibitem{geva2022transformer}
Mor Geva, Avi Caciularu, Kevin Wang, and Yoav Goldberg.
\newblock Transformer feed-forward layers build predictions by promoting concepts in the vocabulary space.
\newblock In {\em Empirical Methods in Natural Language Processing}, pages 30--45, 2022.

\bibitem{geva2021transformer}
Mor Geva, Roei Schuster, Jonathan Berant, and Omer Levy.
\newblock Transformer feed-forward layers are key-value memories.
\newblock In {\em Empirical Methods in Natural Language Processing}, pages 5484--5495, 2021.

\bibitem{guo2022adversarially}
Chong Guo, Michael Lee, Guillaume Leclerc, Joel Dapello, Yug Rao, Aleksander Madry, and James Dicarlo.
\newblock Adversarially trained neural representations are already as robust as biological neural representations.
\newblock In {\em International Conference on Machine Learning}, pages 8072--8081, 2022.

\bibitem{gupta2023editing}
Anshita Gupta, Debanjan Mondal, Akshay Sheshadri, Wenlong Zhao, Xiang Li, Sarah Wiegreffe, and Niket Tandon.
\newblock Editing common sense in transformers.
\newblock In {\em Empirical Methods in Natural Language Processing}, pages 8214--8232, 2023.

\bibitem{hartvigsen2023aging}
Thomas Hartvigsen, Swami Sankaranarayanan, Hamid Palangi, Yoon Kim, and Marzyeh Ghassemi.
\newblock Aging with grace: Lifelong model editing with discrete key-value adaptors.
\newblock In {\em Advances in Neural Information Processing Systems}, pages 47934--47959, 2023.

\bibitem{hase2023does}
Peter Hase, Mohit Bansal, Been Kim, and Asma Ghandeharioun.
\newblock Does localization inform editing? {Surprising} differences in causality-based localization vs. knowledge editing in language models.
\newblock In {\em Advances in Neural Information Processing Systems}, pages 17643--17668, 2023.

\bibitem{he2023sensitivity}
Haoyu He, Jianfei Cai, Jing Zhang, Dacheng Tao, and Bohan Zhuang.
\newblock Sensitivity-aware visual parameter-efficient fine-tuning.
\newblock In {\em IEEE International Conference on Computer Vision}, pages 11825--11835, 2023.

\bibitem{he2016deep}
Kaiming He, Xiangyu Zhang, Shaoqing Ren, and Jian Sun.
\newblock Deep residual learning for image recognition.
\newblock In {\em IEEE Conference on Computer Vision and Pattern Recognition}, pages 770--778, 2016.

\bibitem{hendrycks2021many}
Dan Hendrycks, Steven Basart, Norman Mu, Saurav Kadavath, Frank Wang, Evan Dorundo, Rahul Desai, Tyler Zhu, Samyak Parajuli, Mike Guo, Dawn Song, Jacob Steinhardt, and Justin Gilmer.
\newblock The many faces of robustness: A critical analysis of out-of-distribution generalization.
\newblock In {\em IEEE International Conference on Computer Vision}, pages 8340--8349, 2021.

\bibitem{hendrycks2016gaussian}
Dan Hendrycks and Kevin Gimpel.
\newblock Gaussian error linear units ({GELU}s).
\newblock {\em arXiv preprint arXiv:1606.08415}, 2016.

\bibitem{hinton2015distilling}
Geoffrey Hinton, Oriol Vinyals, and Jeffrey Dean.
\newblock Distilling the knowledge in a neural network.
\newblock {\em ArXiv preprint arXiv:1503.02531}, 2015.

\bibitem{hu2022lora}
Edward~J Hu, Yelong Shen, Phillip Wallis, Zeyuan Allen-Zhu, Yuanzhi Li, Shean Wang, Lu~Wang, and Weizhu Chen.
\newblock {LoRA}: Low-rank adaptation of large language models.
\newblock In {\em International Conference on Learning Representations}, 2022.

\bibitem{huang2023transformer}
Zeyu Huang, Yikang Shen, Xiaofeng Zhang, Jie Zhou, Wenge Rong, and Zhang Xiong.
\newblock Transformer-patcher: One mistake worth one neuron.
\newblock In {\em International Conference on Learning Representations}, 2023.

\bibitem{jang1999binarization}
Jeong-Hun Jang and Ki-Sang Hong.
\newblock Binarization of noisy gray-scale character images by thin line modeling.
\newblock {\em Pattern Recognition}, 32(5):743--752, 1999.

\bibitem{kingma2015adam}
Diederik~P Kingma and Jimmy Ba.
\newblock Adam: A method for stochastic optimization.
\newblock In {\em International Conference on Learning Representations}, 2015.

\bibitem{kohonen1972correlation}
Teuvo Kohonen.
\newblock Correlation matrix memories.
\newblock {\em IEEE Transactions on Computers}, 100(4):353--359, 1972.

\bibitem{krizhevsky2012imagenet}
Alex Krizhevsky, Ilya Sutskever, and Geoffrey~E Hinton.
\newblock {ImageNet} classification with deep convolutional neural networks.
\newblock In {\em Advances in Neural Information Processing Systems}, pages 1097--1105, 2012.

\bibitem{lake2015human}
Brenden~M Lake, Ruslan Salakhutdinov, and Joshua~B Tenenbaum.
\newblock Human-level concept learning through probabilistic program induction.
\newblock {\em Science}, 350(6266):1332--1338, 2015.

\bibitem{liu2021swin}
Ze~Liu, Yutong Lin, Yue Cao, Han Hu, Yixuan Wei, Zheng Zhang, Stephen Lin, and Baining Guo.
\newblock Swin {Transformer}: Hierarchical vision {Transformer} using shifted windows.
\newblock In {\em IEEE International Conference on Computer Vision}, pages 10012--10022, 2021.

\bibitem{liu2019metapruning}
Zechun Liu, Haoyuan Mu, Xiangyu Zhang, Zichao Guo, Xin Yang, Kwang-Ting Cheng, and Jian Sun.
\newblock {MetaPruning}: Meta learning for automatic neural network channel pruning.
\newblock In {\em IEEE International Conference on Computer Vision}, pages 3296--3305, 2019.

\bibitem{madry2018towards}
Aleksander Madry, Aleksandar Makelov, Ludwig Schmidt, Dimitris Tsipras, and Adrian Vladu.
\newblock Towards deep learning models resistant to adversarial attacks.
\newblock In {\em International Conference on Learning Representations}, 2018.

\bibitem{meng2022locating}
Kevin Meng, David Bau, Alex Andonian, and Yonatan Belinkov.
\newblock Locating and editing factual associations in {GPT}.
\newblock In {\em Advances in Neural Information Processing Systems}, pages 17359--17372, 2022.

\bibitem{meng2023mass}
Kevin Meng, Arnab~S Sharma, Alex Andonian, Yonatan Belinkov, and David Bau.
\newblock Mass-editing memory in a {Transformer}.
\newblock In {\em International Conference on Learning Representations}, 2023.

\bibitem{mitchell2022fast}
Eric Mitchell, Charles Lin, Antoine Bosselut, Chelsea Finn, and Christopher~D Manning.
\newblock Fast model editing at scale.
\newblock In {\em International Conference on Learning Representations}, 2022.

\bibitem{mitchell2022memory}
Eric Mitchell, Charles Lin, Antoine Bosselut, Christopher~D Manning, and Chelsea Finn.
\newblock Memory-based model editing at scale.
\newblock In {\em International Conference on Machine Learning}, pages 15817--15831, 2022.

\bibitem{murty2022fixing}
Shikhar Murty, Christopher~D Manning, Scott Lundberg, and Marco~T Ribeiro.
\newblock Fixing model bugs with natural language patches.
\newblock In {\em Empirical Methods in Natural Language Processing}, pages 11600--11613, 2022.

\bibitem{radford2021learning}
Alec Radford, Jong~Wook Kim, Chris Hallacy, Aditya Ramesh, Gabriel Goh, Sandhini Agarwal, Girish Sastry, Amanda Askell, Pamela Mishkin, Jack Clark, et~al.
\newblock Learning transferable visual models from natural language supervision.
\newblock In {\em International Conference on Machine Learning}, pages 8748--8763, 2021.

\bibitem{rajeswaran2019meta}
Aravind Rajeswaran, Chelsea Finn, Sham~M Kakade, and Sergey Levine.
\newblock Meta-learning with implicit gradients.
\newblock In {\em Advances in Neural Information Processing Systems}, pages 113--124, 2019.

\bibitem{recht2019imagenet}
Benjamin Recht, Rebecca Roelofs, Ludwig Schmidt, and Vaishaal Shankar.
\newblock Do {ImageNet} classifiers generalize to {ImageNet}?
\newblock In {\em International Conference on Machine Learning}, pages 5389--5400, 2019.

\bibitem{rombach2022high}
Robin Rombach, Andreas Blattmann, Dominik Lorenz, Patrick Esser, and Bj{\"o}rn Ommer.
\newblock High-resolution image synthesis with latent diffusion models.
\newblock In {\em IEEE Conference on Computer Vision and Pattern Recognition}, pages 10684--10695, 2022.

\bibitem{russakovsky2015imagenet}
Olga Russakovsky, Jia Deng, Hao Su, Jonathan Krause, Sanjeev Satheesh, Sean Ma, Zhiheng Huang, Andrej Karpathy, Aditya Khosla, Michael Bernstein, Alexander~C Berg, and Li~Fei-Fei.
\newblock {ImageNet} large scale visual recognition challenge.
\newblock {\em International Journal of Computer Vision}, 115:211--252, 2015.

\bibitem{salman2020adversarially}
Hadi Salman, Andrew Ilyas, Logan Engstrom, Ashish Kapoor, and Aleksander Madry.
\newblock Do adversarially robust {ImageNet} models transfer better?
\newblock In {\em Advances in Neural Information Processing Systems}, pages 3533--3545, 2020.

\bibitem{santurkar2021editing}
Shibani Santurkar, Dimitris Tsipras, Mahalaxmi Elango, David Bau, Antonio Torralba, and Aleksander Madry.
\newblock Editing a classifier by rewriting its prediction rules.
\newblock In {\em Advances in Neural Information Processing Systems}, pages 23359--23373, 2021.

\bibitem{santurkar2021breeds}
Shibani Santurkar, Dimitris Tsipras, and Aleksander Madry.
\newblock {BREEDS}: Benchmarks for subpopulation shift.
\newblock In {\em International Conference on Learning Representations}, 2021.

\bibitem{shang2022neural}
Haopu Shang, Jia-Liang Wu, Wenjing Hong, and Chao Qian.
\newblock Neural network pruning by cooperative coevolution.
\newblock In {\em International Joint Conference on Artificial Intelligence}, pages 4814--4820, 2022.

\bibitem{simonyan2014very}
Karen Simonyan and Andrew Zisserman.
\newblock Very deep convolutional networks for large-scale image recognition.
\newblock In {\em International Conference on Learning Representations}, 2015.

\bibitem{steiner2022how}
Andreas Steiner, Alexander Kolesnikov, Xiaohua Zhai, Ross Wightman, Jakob Uszkoreit, and Lucas Beyer.
\newblock How to train your {ViT}? {Data}, {augmentation}, and {regularization} in vision {Transformers}.
\newblock {\em Transactions on Machine Learning Research}, 2022.

\bibitem{tan2023massive}
Chenmien Tan, Ge~Zhang, and Jie Fu.
\newblock Massive editing for large language models via meta learning.
\newblock In {\em International Conference on Learning Representations}, 2024.

\bibitem{thrun1995learning}
Sebastian Thrun and Tom~M Mitchell.
\newblock Learning one more thing.
\newblock In {\em International Joint Conference on Artificial Intelligence}, pages 1217--1223, 1995.

\bibitem{vendrow2023dataset}
Joshua Vendrow, Saachi Jain, Logan Engstrom, and Aleksander Madry.
\newblock Dataset interfaces: Diagnosing model failures using controllable counterfactual generation.
\newblock {\em ArXiv preprint arXiv:2302.07865}, 2023.

\bibitem{wang2019learning}
Haohan Wang, Songwei Ge, Zachary Lipton, and Eric~P Xing.
\newblock Learning robust global representations by penalizing local predictive power.
\newblock In {\em Advances in Neural Information Processing Systems}, pages 10506--10518, 2019.

\bibitem{wang2020going}
Haotao Wang, Tianlong Chen, Zhangyang Wang, and Kede Ma.
\newblock I am going mad: Maximum discrepancy competition for comparing classifiers adaptively.
\newblock In {\em International Conference on Learning Representations}, 2020.

\bibitem{wiles2022discovering}
Olivia Wiles, Isabela Albuquerque, and Sven Gowal.
\newblock Discovering bugs in vision models using off-the-shelf image generation and captioning.
\newblock In {\em Advances in Neural Information Processing Systems Workshop on Machine Learning Safety}, 2022.

\bibitem{wu2023depn}
Xinwei Wu, Junzhuo Li, Minghui Xu, Weilong Dong, Shuangzhi Wu, Chao Bian, and Deyi Xiong.
\newblock {DEPN}: Detecting and editing privacy neurons in pretrained language models.
\newblock In {\em Empirical Methods in Natural Language Processing}, pages 2875--2886, 2023.

\bibitem{yao2023editing}
Yunzhi Yao, Peng Wang, Bozhong Tian, Siyuan Cheng, Zhoubo Li, Shumin Deng, Huajun Chen, and Ningyu Zhang.
\newblock Editing large language models: Problems, methods, and opportunities.
\newblock In {\em Empirical Methods in Natural Language Processing}, pages 10222--10240, 2023.

\bibitem{yun2019cutmix}
Sangdoo Yun, Dongyoon Han, S~Joon Oh, Sanghyuk Chun, Junsuk Choe, and Youngjoon Yoo.
\newblock {CutMix}: Regularization strategy to train strong classifiers with localizable features.
\newblock In {\em International Conference on Computer Vision}, pages 6023--6032, 2019.

\bibitem{zheng2023can}
Ce~Zheng, Lei Li, Qingxiu Dong, Yuxuan Fan, Zhiyong Wu, Jingjing Xu, and Baobao Chang.
\newblock Can we edit factual knowledge by in-context learning?
\newblock In {\em Empirical Methods in Natural Language Processing}, pages 4862--4876, 2023.

\bibitem{zhong2023mquake}
Zexuan Zhong, Zhengxuan Wu, Christopher~D Manning, Christopher Potts, and Danqi Chen.
\newblock {MQuAKE}: Assessing knowledge editing in language models via multi-hop questions.
\newblock In {\em Empirical Methods in Natural Language Processing}, pages 15686--15702, 2023.

\end{thebibliography}
}

\newpage

\appendix

\section{More Details about the Editing Benchmark}
\subsection{Natural Image Subset} \label{app:natural}
\begin{table}[htbp]
        \renewcommand\thetable{A}
	\caption{Statistics of the natural image subset. The first column lists identifiers for each object category in ImageNet-1k. The ``Class Name'' in the second column is in the format as ``prediction by the stronger model''-``prediction by the base model.'' }
	\label{app_table:natrual_dataset_statistics}
	\centering
        {\begin{tabular}{ccc} 
		\toprule
		  Group Identifier &  Class Name  & Sample Number \\
		\hline
		{890-430} & \texttt{volleyball}-\texttt{basketball} & 123 \\
		{933-923} & \texttt{cheeseburger}-\texttt{plate} & 133 \\
		{470-644} & \texttt{candle}-\texttt{matchstick} & 113 \\
		{900-437} & \texttt{water tower}-\texttt{beacon} & 159 \\
  		{609-586} & \texttt{jeep}-\texttt{half track} & 410 \\
		{543-422} & \texttt{dumbbell}-\texttt{barbell} & 240 \\
		{879-762} & \texttt{umbrella}-\texttt{restaurant} & 49 \\
		{417-865} & \texttt{balloon}-\texttt{toyshop} &  75 \\
		{573-751} & \texttt{go-kart}-\texttt{racer} &  172 \\
  		{880-671} & \texttt{unicycle}-\texttt{mountain bike} & 149 \\		
            {954-582} & \texttt{banana}-\texttt{grocery store} & 75 \\
		{752-890} & \texttt{racket}-\texttt{volleyball} & 137 \\
		{640-539} & \texttt{manhole cover}-\texttt{doormat} & 80 \\
		{407-654} & \texttt{ambulance}-\texttt{minibus} & 155 \\
  		{562-975} & \texttt{fountain}-\texttt{lakeside} & 155 \\
            {888-718} & \texttt{viaduct}-\texttt{pier} & 129 \\
		\bottomrule
	\end{tabular}}
\end{table}

\begin{figure}[]
        \renewcommand\thefigure{A}
	\centering
   	\begin{minipage}[]{0.49\columnwidth}
            \centering
    	{\includegraphics[width = \columnwidth]{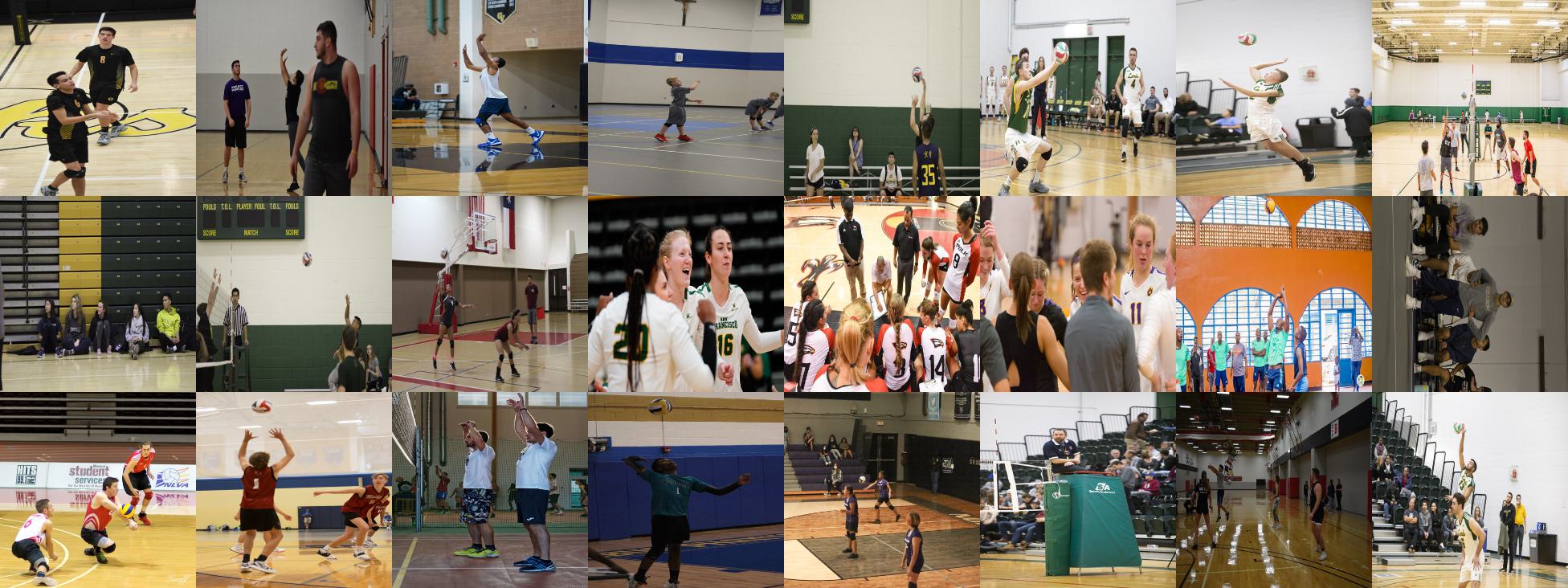}}
            \centerline{Group 890-430: \texttt{volleyball}-\texttt{basketball}}
    	\label{fig:890_430}
	\end{minipage}
        \hfill
        \begin{minipage}[]{0.49\columnwidth}
            \centering
    	{\includegraphics[width = \columnwidth]{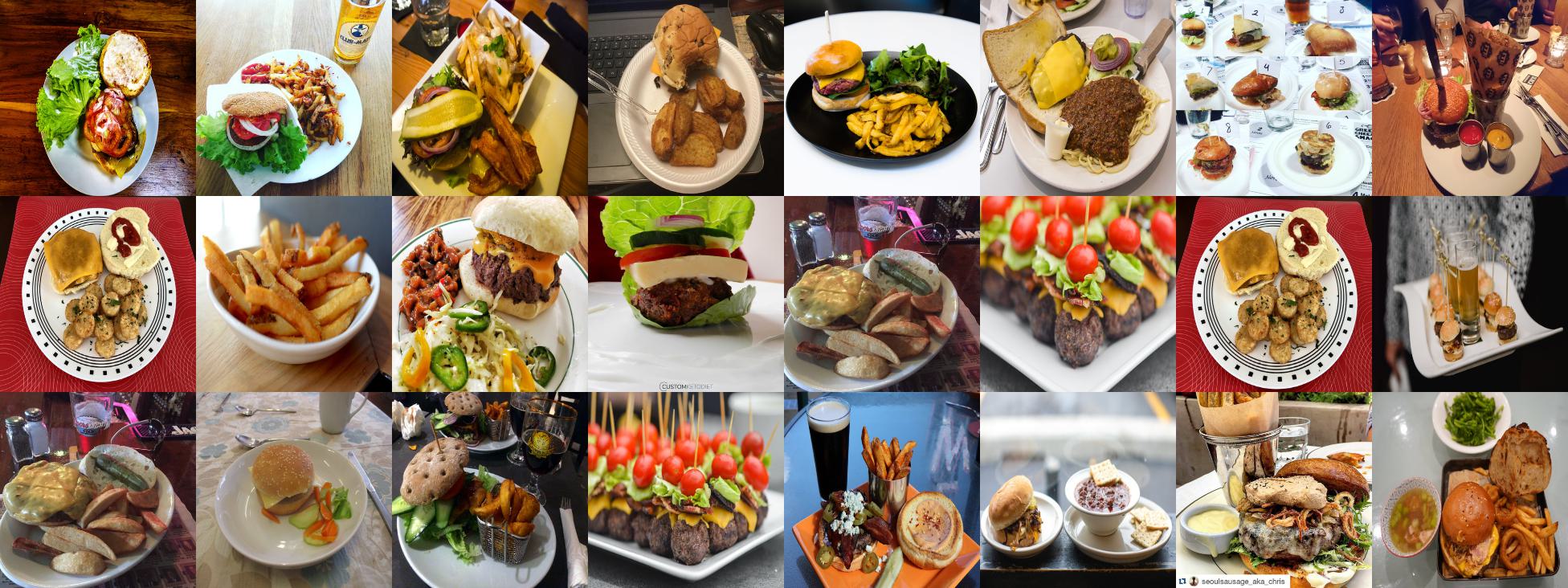}}
             \centerline{Group 933-923: \texttt{cheeseburger}-\texttt{plate}}
    	\label{fig:933_923}
	\end{minipage}
        \hfill
         \begin{minipage}[]{0.49\columnwidth}
            \centering
    	{\includegraphics[width = \columnwidth]{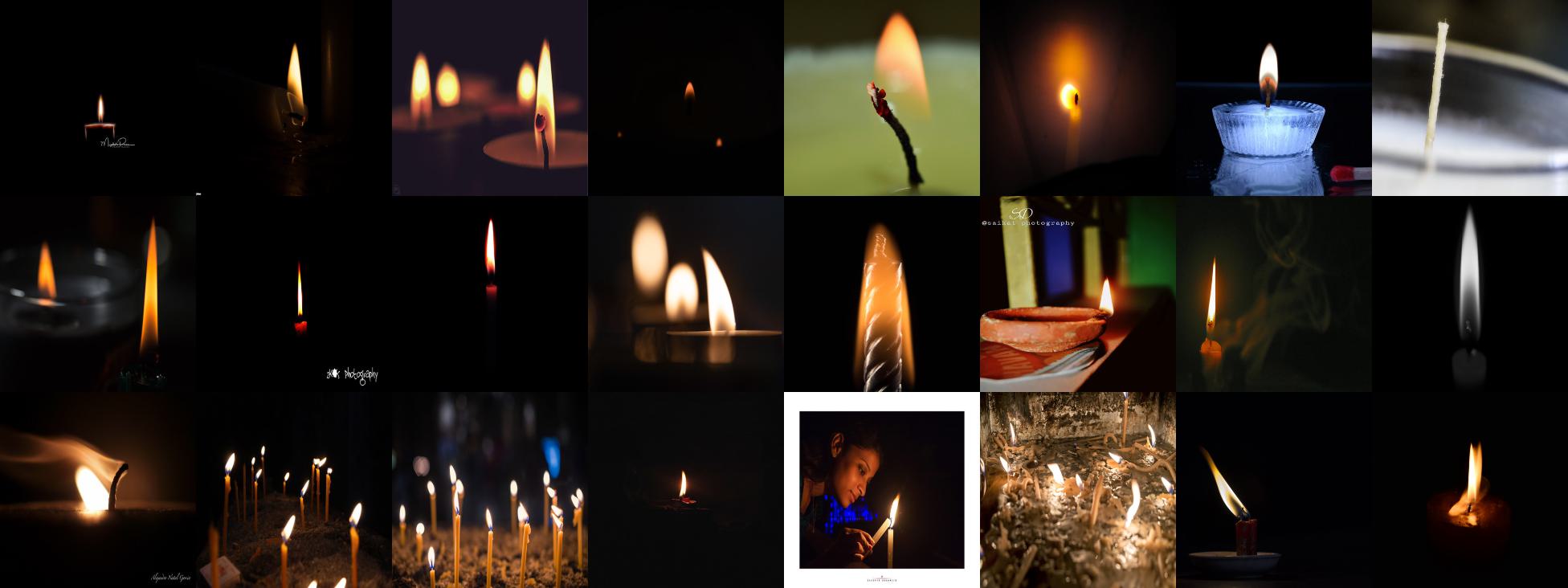}}
            \centerline{Group 470-644:\texttt{candle}-\texttt{matchstick}}
    	\label{fig:470_644}
	\end{minipage}
        \hfill
         \begin{minipage}[]{0.49\columnwidth}
            \centering
    	{\includegraphics[width = \columnwidth]{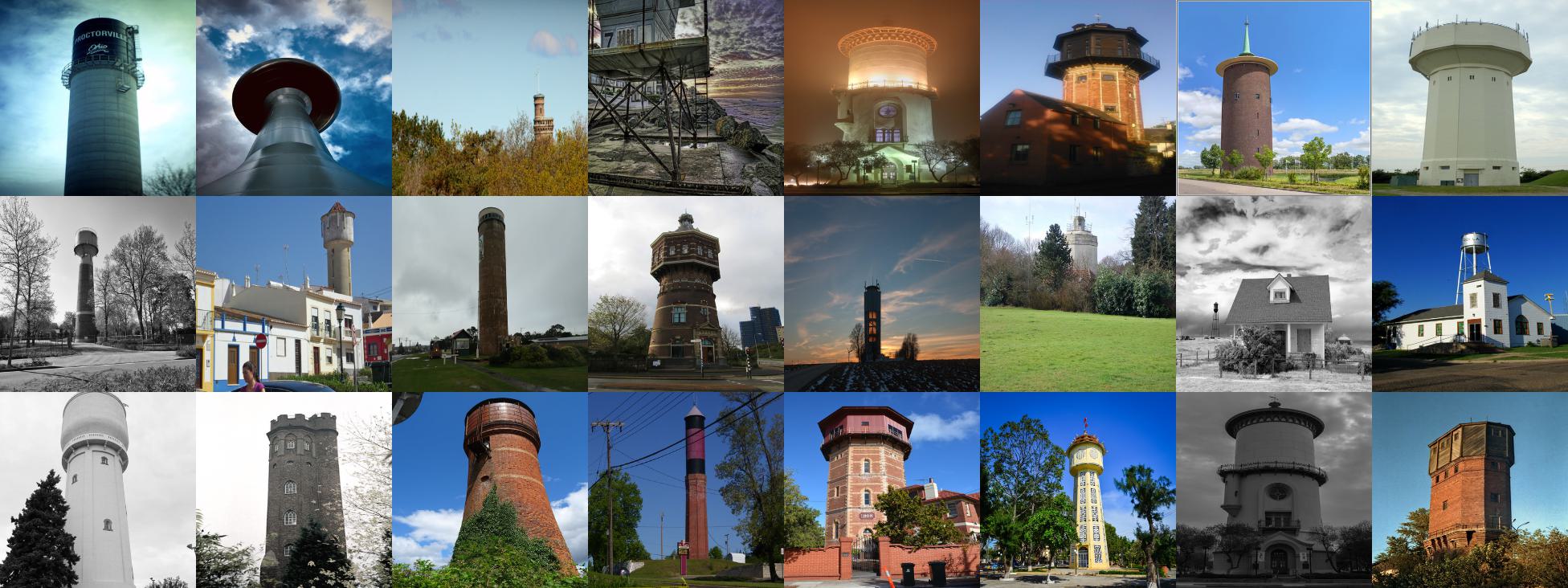}}
            \centerline{Group 900-437: \texttt{water tower}-\texttt{beacon}}
    	\label{fig:900_437}
	\end{minipage}
        \hfill
         \begin{minipage}[]{0.49\columnwidth}
            \centering
    	{\includegraphics[width = \columnwidth]{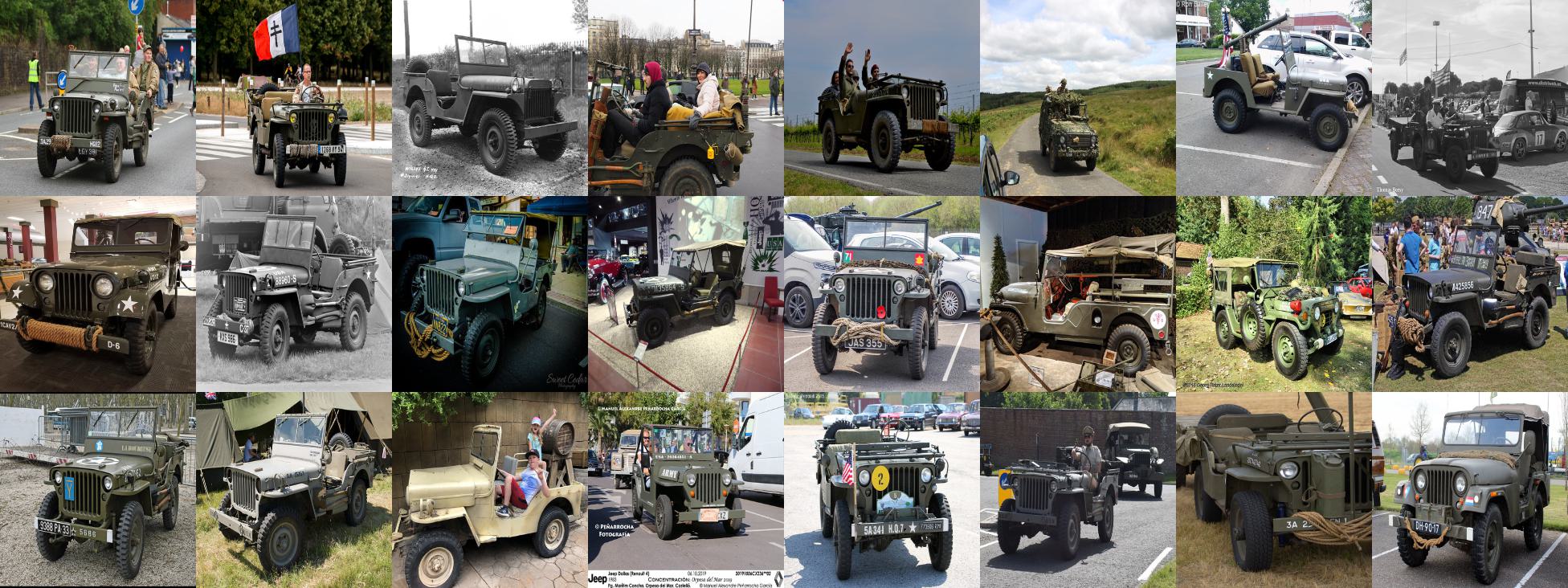}}
            \centerline{Group 609-586: \texttt{jeep}-\texttt{half track}}
    	\label{fig:609_586}
	\end{minipage}
        \hfill
         \begin{minipage}[]{0.49\columnwidth}
            \centering
    	{\includegraphics[width = \columnwidth]{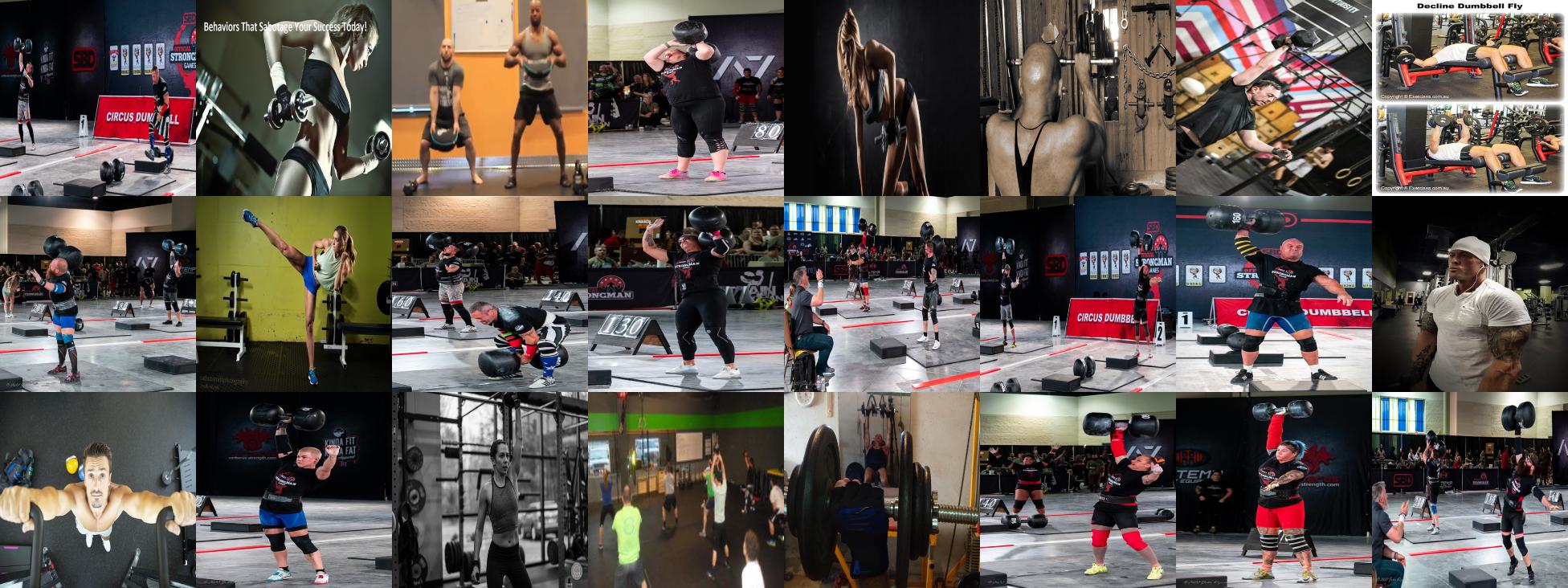}}
            \centerline{Group 543-422: \texttt{dumbbell}-\texttt{barbell}}
    	\label{fig:543_422}
	\end{minipage}
        \hfill
         \begin{minipage}[]{0.49\columnwidth}
            \centering
    	{\includegraphics[width = \columnwidth]{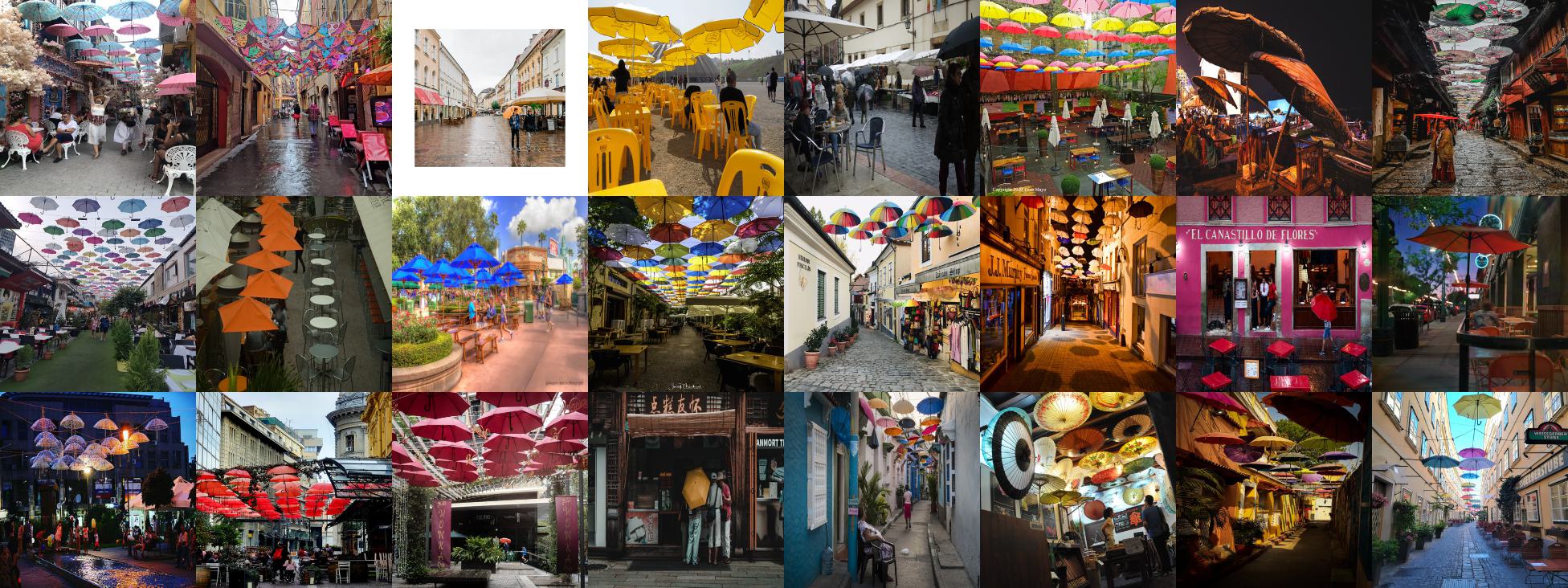}}
            \centerline{Group 879-762: \texttt{umbrella}-\texttt{restaurant}}
    	\label{fig:879_762}
	\end{minipage}
        \hfill
        \centering
   	\begin{minipage}[]{0.49\columnwidth}
            \centering
    	{\includegraphics[width = \columnwidth]{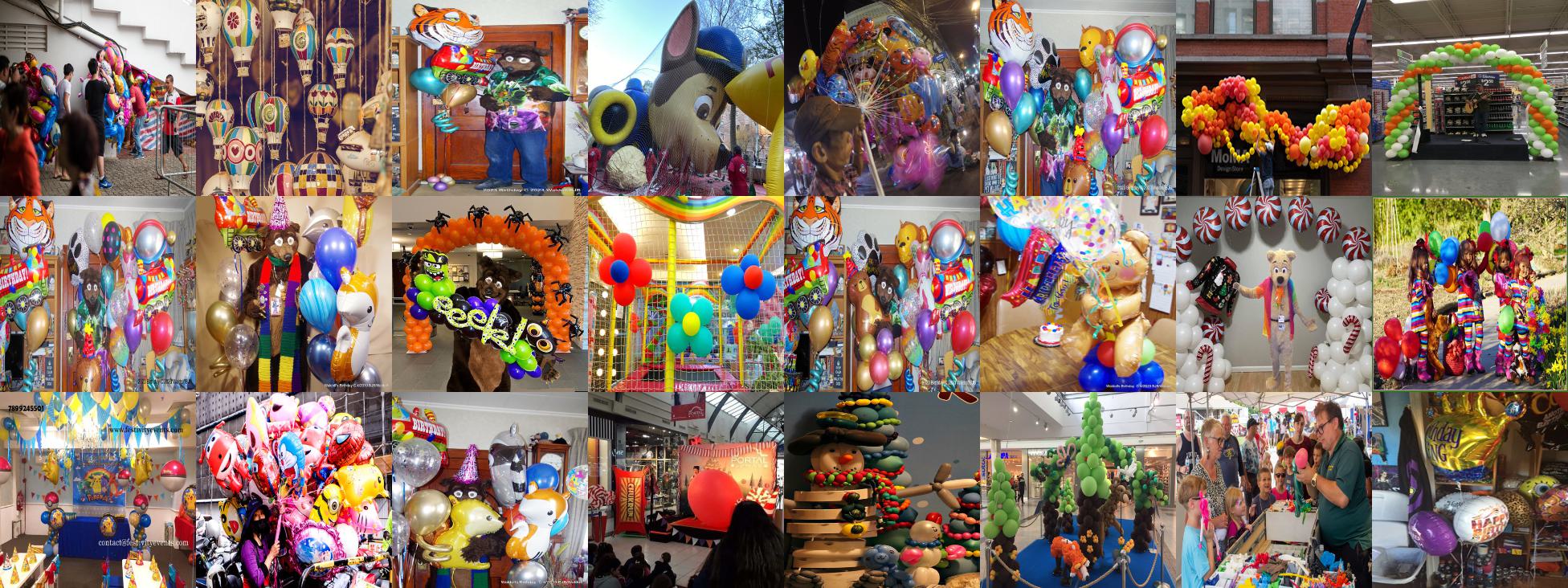}}
            \centerline{Group 417-865: \texttt{balloon}-\texttt{toyshop}}
    	\label{fig:417_865}
	\end{minipage}
        \caption{Visual examples in each group of the natural image subset. Part 1/2.}
        \label{fig:natural_dataset_1}
 \end{figure}

\begin{figure}[]
        \renewcommand\thefigure{B}
        \begin{minipage}[]{0.49\columnwidth}
            \centering
    	{\includegraphics[width = \columnwidth]{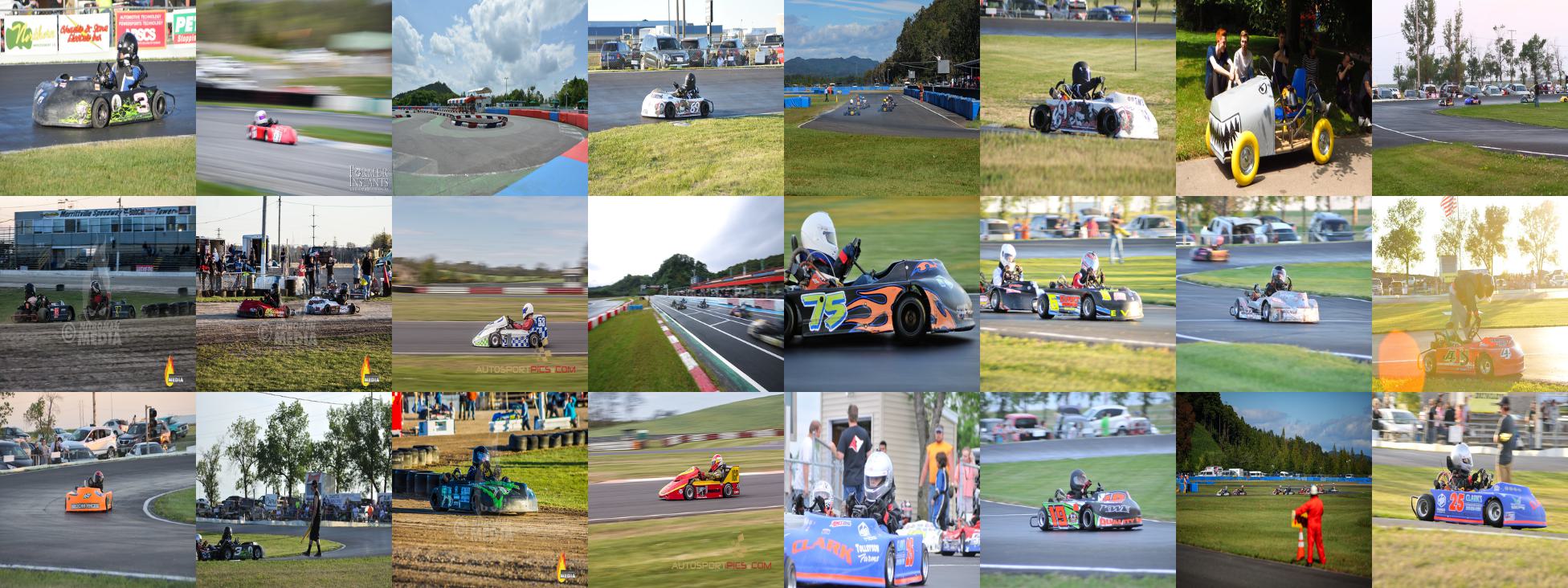}}
            \centerline{Group 573-751: \texttt{go-kart}-\texttt{racer}}
    	\label{fig:573_751}
	\end{minipage}
        \hfill
         \begin{minipage}[]{0.49\columnwidth}
            \centering
    	{\includegraphics[width = \columnwidth]{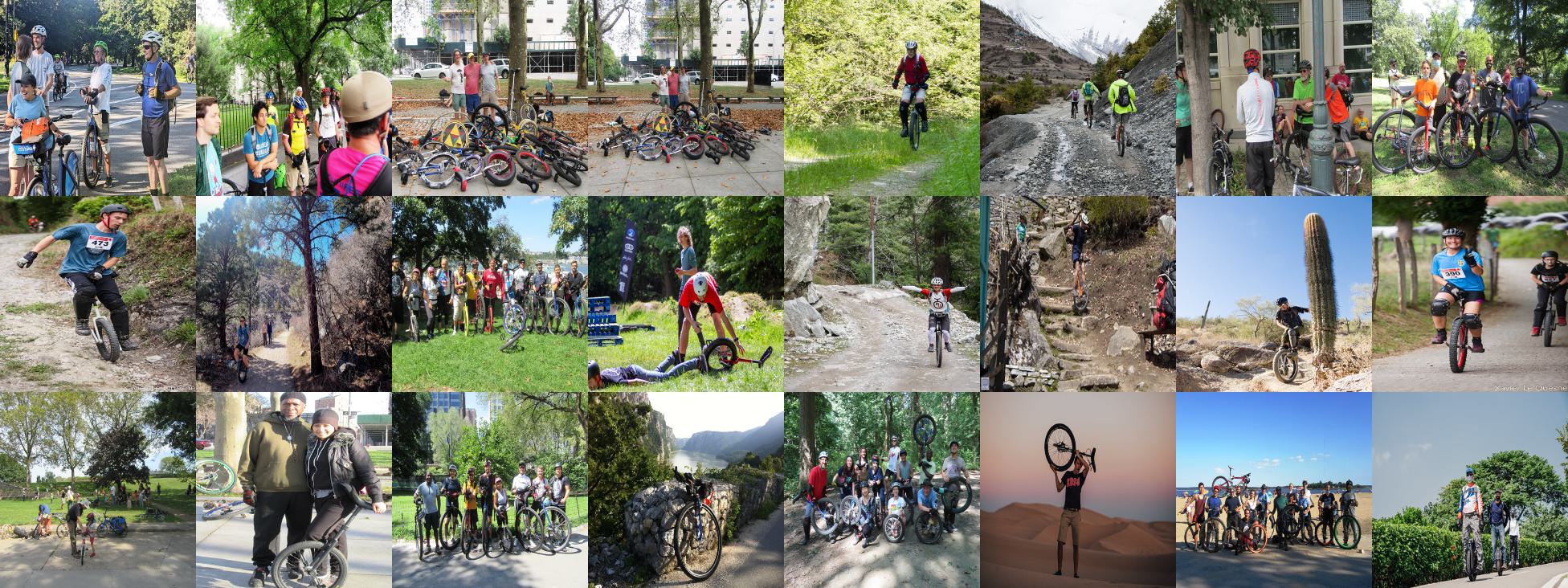}}
            \centerline{Group 880-671: \texttt{unicycle}-\texttt{mountain bike}}
    	\label{fig:880_671}
	\end{minipage}
        \hfill
         \begin{minipage}[]{0.49\columnwidth}
            \centering
    	{\includegraphics[width = \columnwidth]{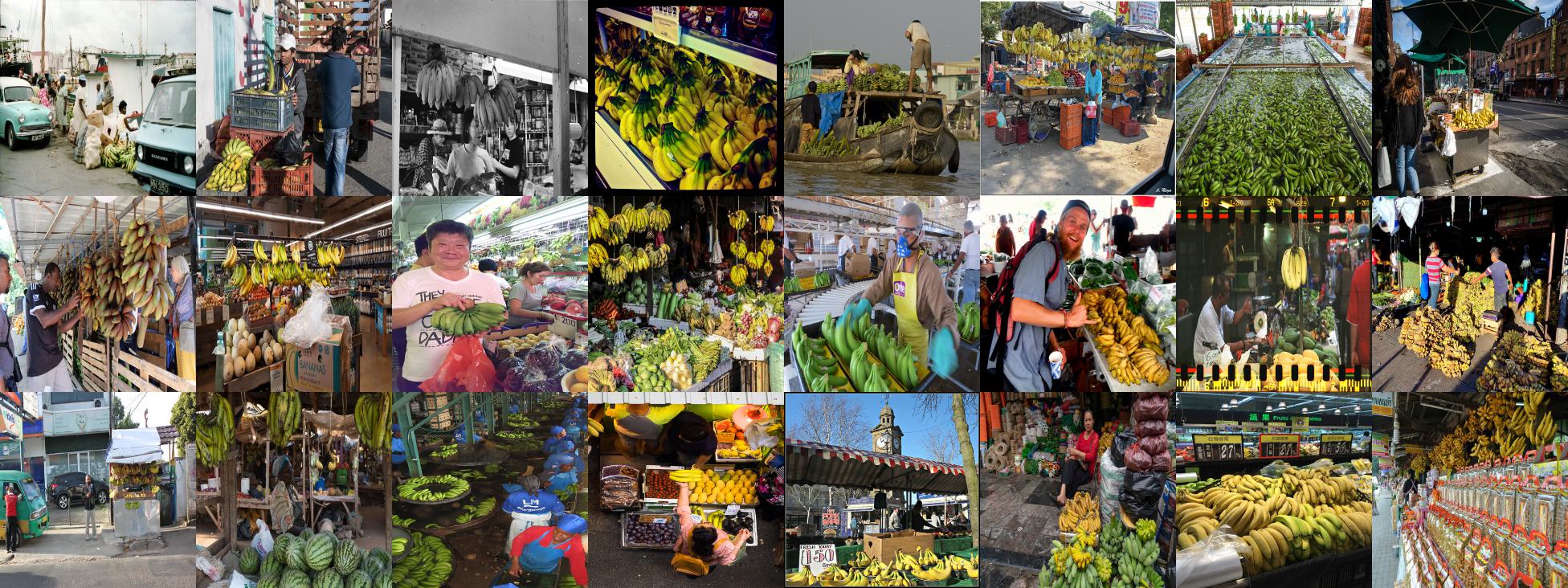}}
            \centerline{Group 954-582: \texttt{banana}-\texttt{grocery store}}
    	\label{fig:954_582}
	\end{minipage}
        \hfill
         \begin{minipage}[]{0.49\columnwidth}
            \centering
    	{\includegraphics[width = \columnwidth]{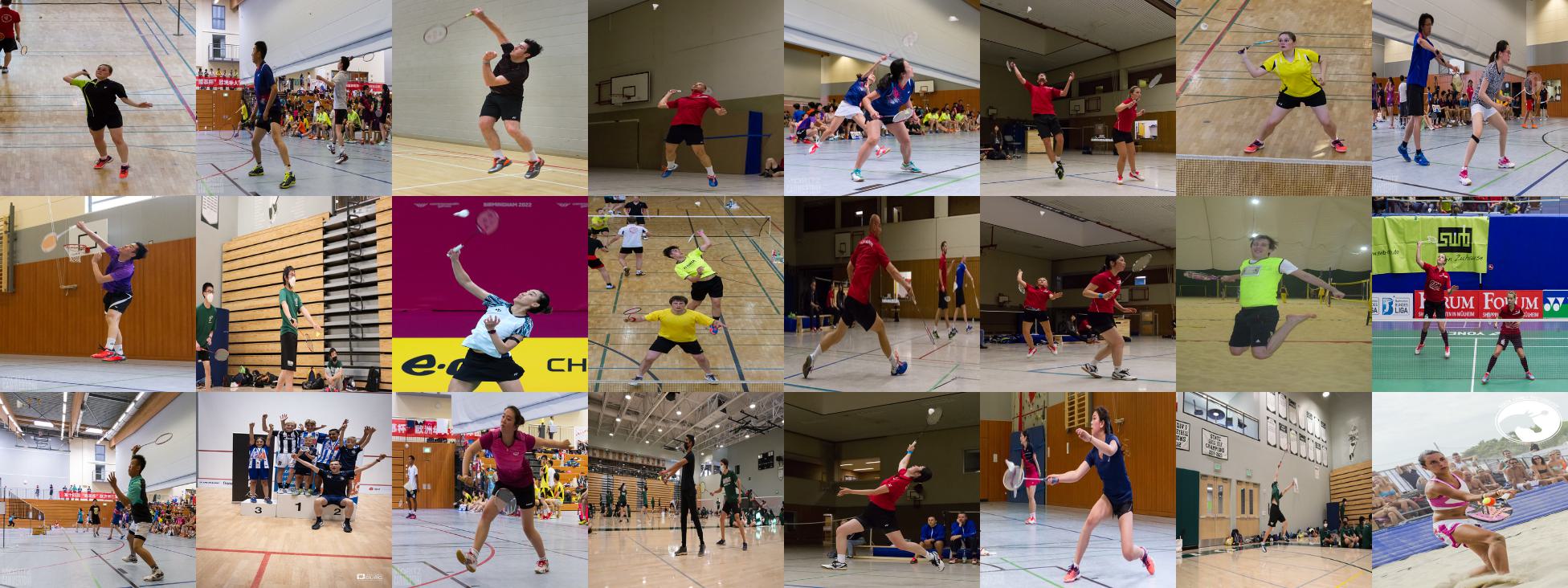}}
            \centerline{Group 752-890: \texttt{racket}-\texttt{volleyball}}
    	\label{fig:752_890}
	\end{minipage}
        \hfill
         \begin{minipage}[]{0.49\columnwidth}
            \centering
    	{\includegraphics[width = \columnwidth]{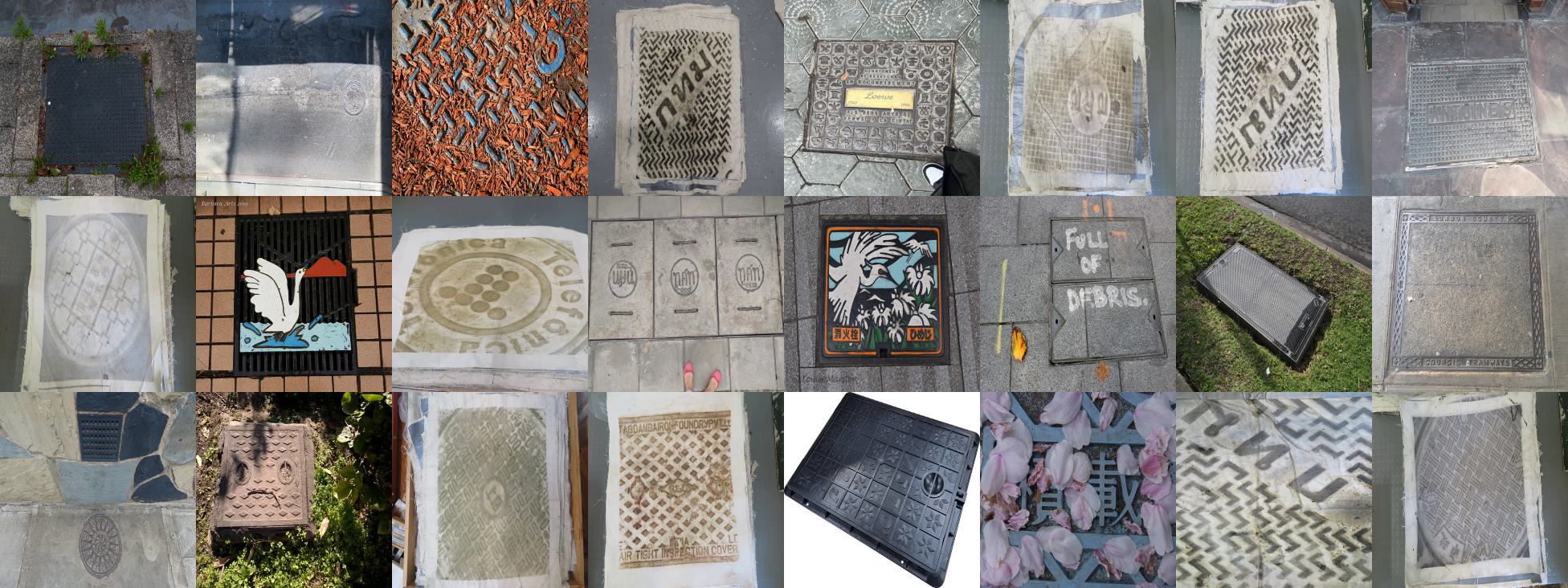}}
            \centerline{Group 640-539:  \texttt{manhole cover}-\texttt{doormat}}
    	\label{fig:640_539}
	\end{minipage}
        \hfill
         \begin{minipage}[]{0.49\columnwidth}
            \centering
    	{\includegraphics[width = \columnwidth]{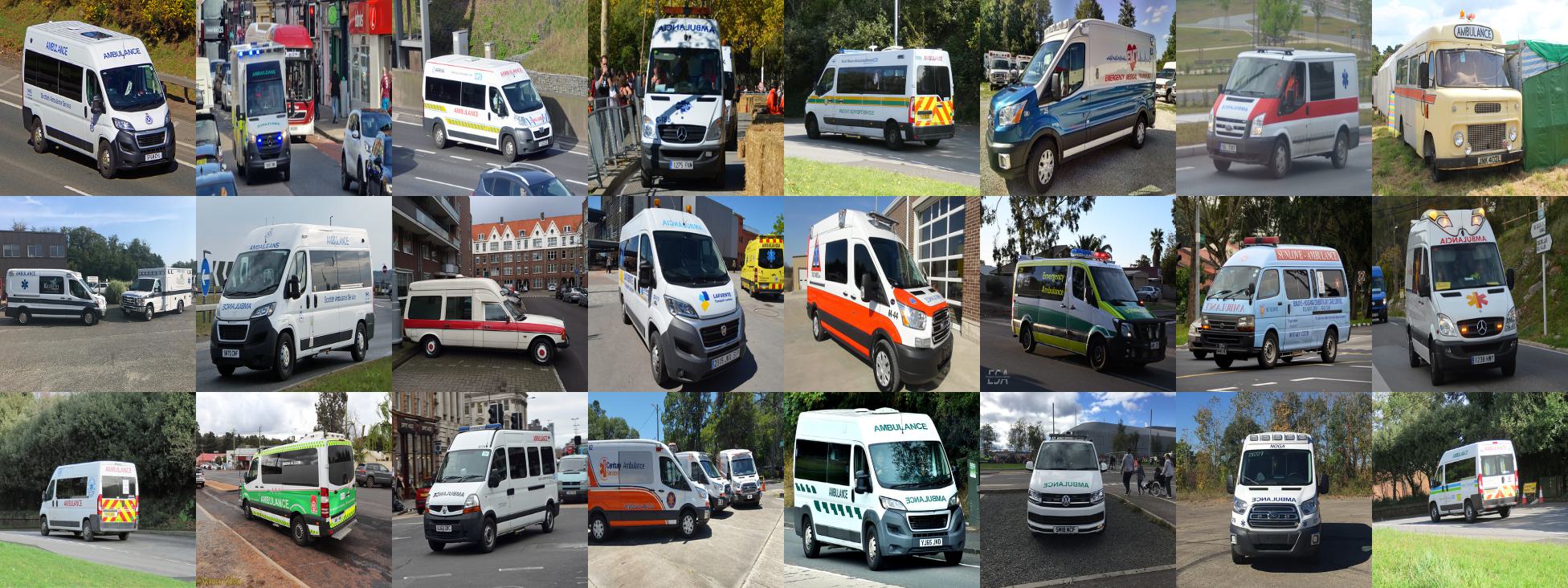}}
            \centerline{Group 407-654: \texttt{ambulance}-\texttt{minibus}}
    	\label{fig:407_654}
	\end{minipage}
        \hfill
             \begin{minipage}[]{0.49\columnwidth}
            \centering
    	{\includegraphics[width = \columnwidth]{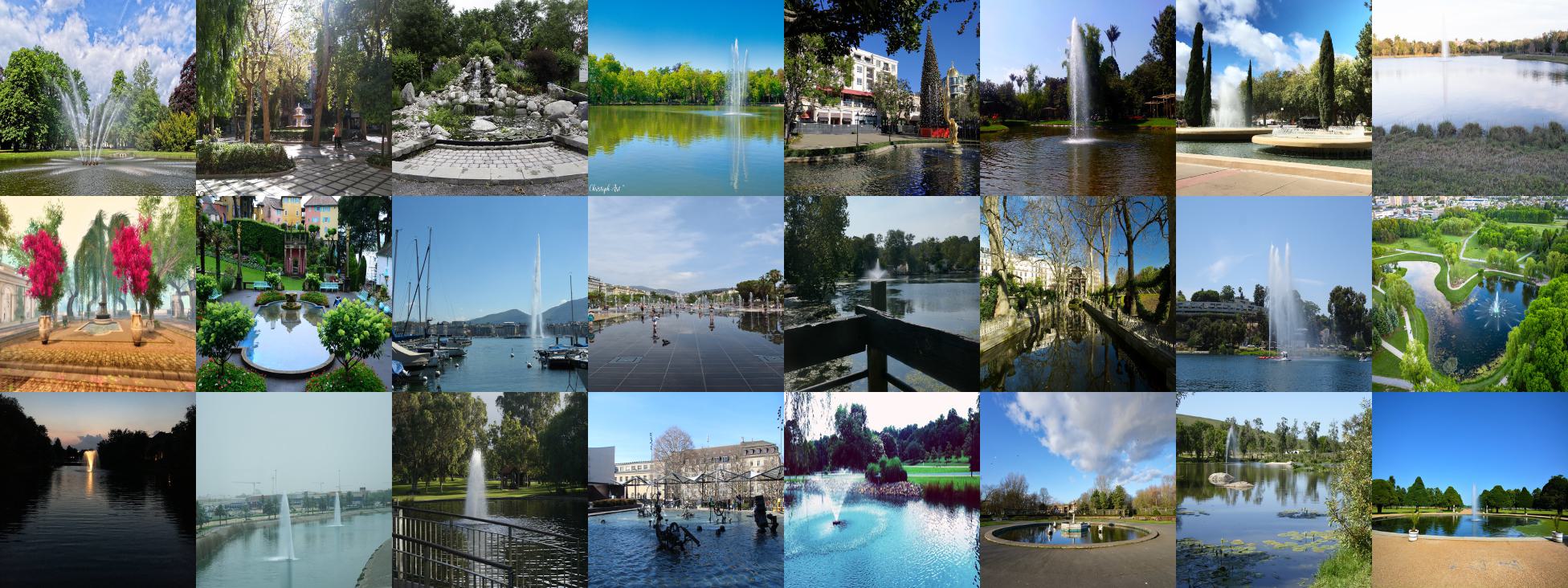}}
            \centerline{Group 562-975: \texttt{fountain}-\texttt{lakeside}}
    	\label{fig:562_975}
	\end{minipage}
        \hfill
         \begin{minipage}[]{0.49\columnwidth}
            \centering
    	{\includegraphics[width = \columnwidth]{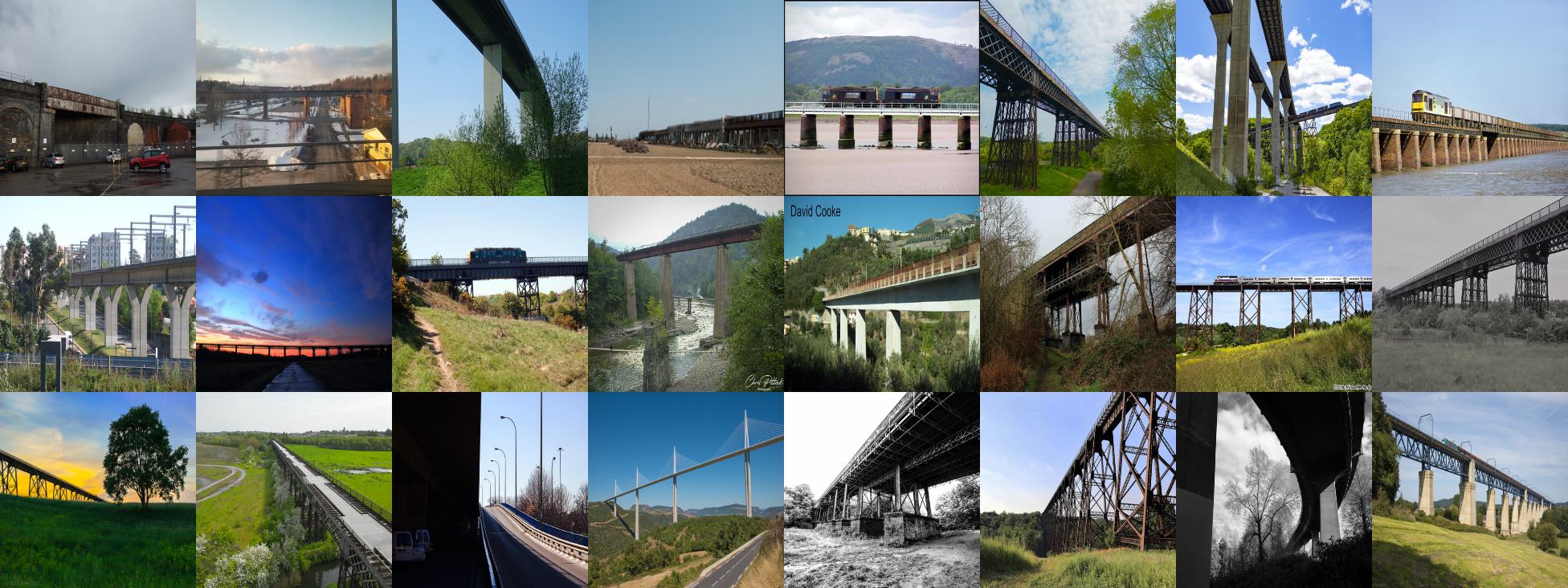}}
            \centerline{Group 888-718: \texttt{fountain}-\texttt{lakeside}}
    	\label{fig:888_718}
	\end{minipage}
 \caption{Visual examples in each group of the natural image subset. Part 2/2.}
 \label{fig:natural_dataset_2}

 \end{figure}

We divide the MAD-searched natural image subset into $16$ groups, whose statistics are listed in Table~\ref{app_table:natrual_dataset_statistics}. Visual examples in each group are shown in Figs.~\ref{fig:natural_dataset_1} and~\ref{fig:natural_dataset_2}. These images are sourced from Flickr, prior to the advent of Stable Diffusion, and are licensed under creative commons.  

\subsection{AI-generated Image Subset} \label{app:ai}
We adopt Textural Inversion~\cite{vendrow2023dataset} and PUG~\cite{bordes2024pug} to construct the AI-generated image subset, encompassing the oil painting and stage light shifts, respectively. The statistics are given in Table~\ref{app_table:ai_dataset_statistics}. 

Specific classes in the oil painting subset include \texttt{stingray}, \texttt{bullfrog}, \texttt{box turtle}, \texttt{garter snake}, \texttt{harvestman}, \texttt{crayfish}, \texttt{hermit crab}, \texttt{mongoose}, \texttt{rhinoceros beetle}, \texttt{weevil}, \texttt{wood rabbit}, \texttt{capuchin}, \texttt{african elephant}, \texttt{breastplate}, \texttt{drumstick}, \texttt{envelope}, \texttt{hand blower}, \texttt{shovel}, \texttt{spatula}, \texttt{syringe}, \texttt{wine bottle}, and \texttt{corn}.

Specific classes in the stage light subset include \texttt{barrel}, \texttt{cofee mug}, \texttt{washer}, \texttt{jack o lantern}, \texttt{vase}, \texttt{throne}, \texttt{soccer ball}, \texttt{basketball}, \texttt{car wheel}, \texttt{vacuum}, \texttt{birdhouse}, \texttt{laptop}, \texttt{piano}, \texttt{pool table}, \texttt{carousel}, \texttt{jellyfish}, \texttt{convertible}, \texttt{motor scooter}, \texttt{mask}, \texttt{sewing machine}, \texttt{hay}, \texttt{gasmask}, \texttt{bell pepper}, \texttt{drum}, \texttt{table lamb}, \texttt{backpack}, \texttt{chicken hen}, \texttt{tennis ball}, \texttt{safe}, \texttt{pay phone}, \texttt{cabbage}, and \texttt{pineapple}.

\begin{figure}[t]
\renewcommand\thefigure{C}
        \centering
         \begin{minipage}[]{0.32\columnwidth}
            \centering
    	{\includegraphics[width = \columnwidth]{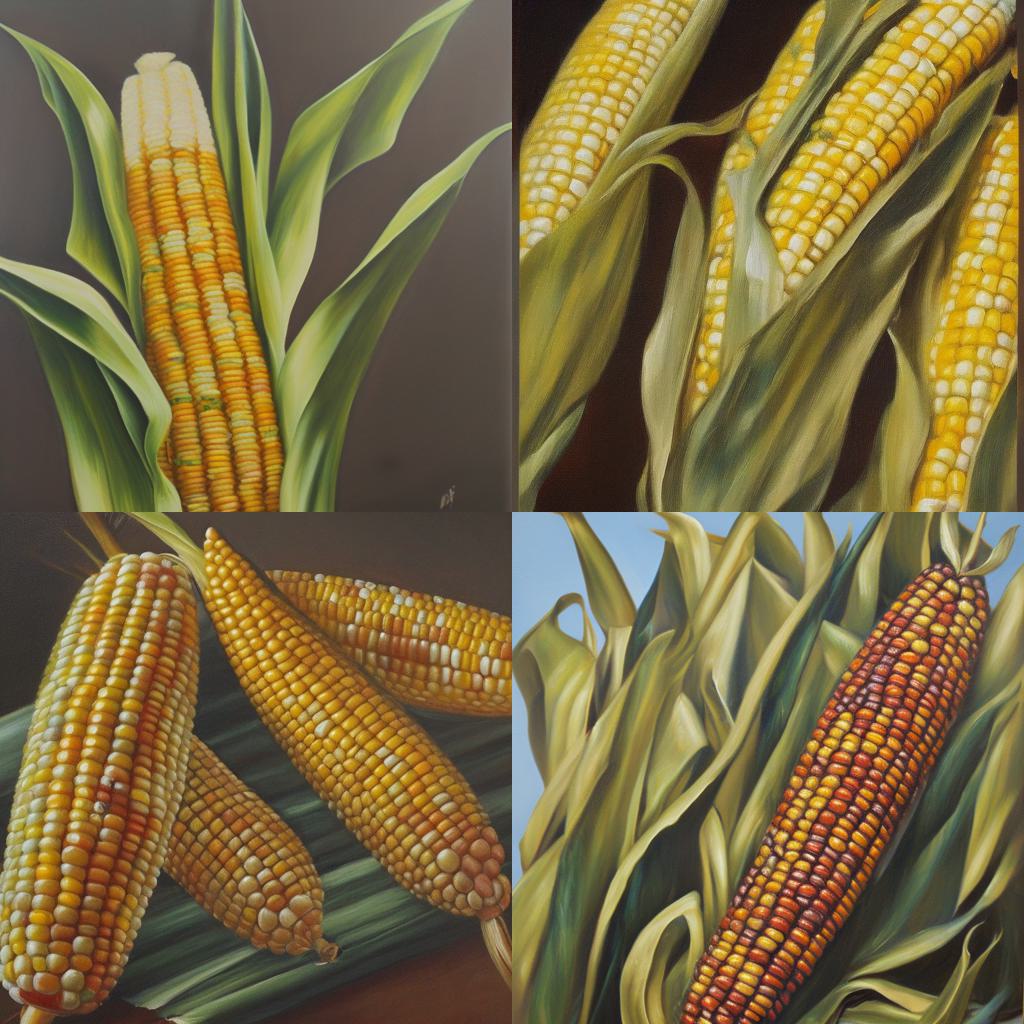}}
            \centerline{\texttt{corn}}
    	\label{fig:corn}
	\end{minipage}
        \hfill
             \begin{minipage}[]{0.32\columnwidth}
            \centering
    	{\includegraphics[width = \columnwidth]{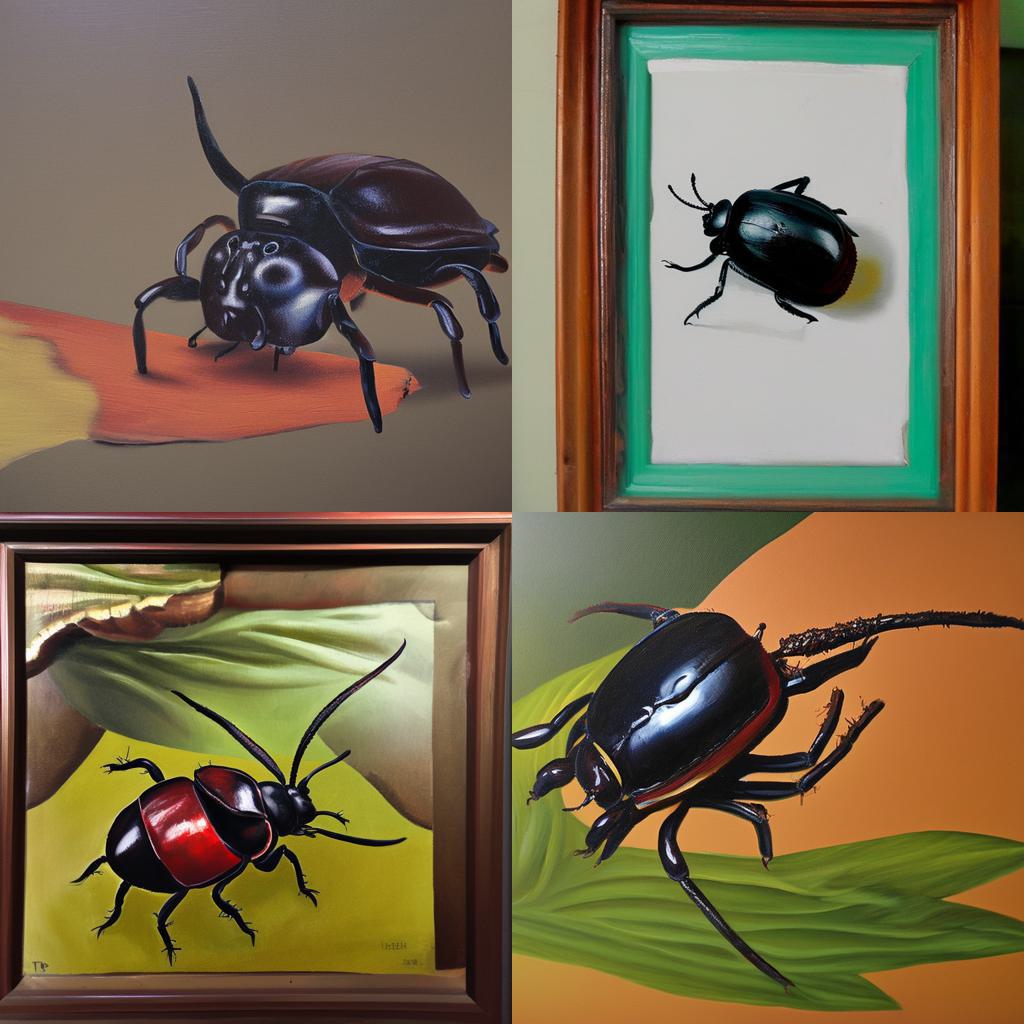}}
            \centerline{\texttt{rhinoceros beetle}}
            \label{fig:rhinoceros}
	\end{minipage}
        \hfill
         \begin{minipage}[]{0.32\columnwidth}
            \centering
    	{\includegraphics[width = \columnwidth]{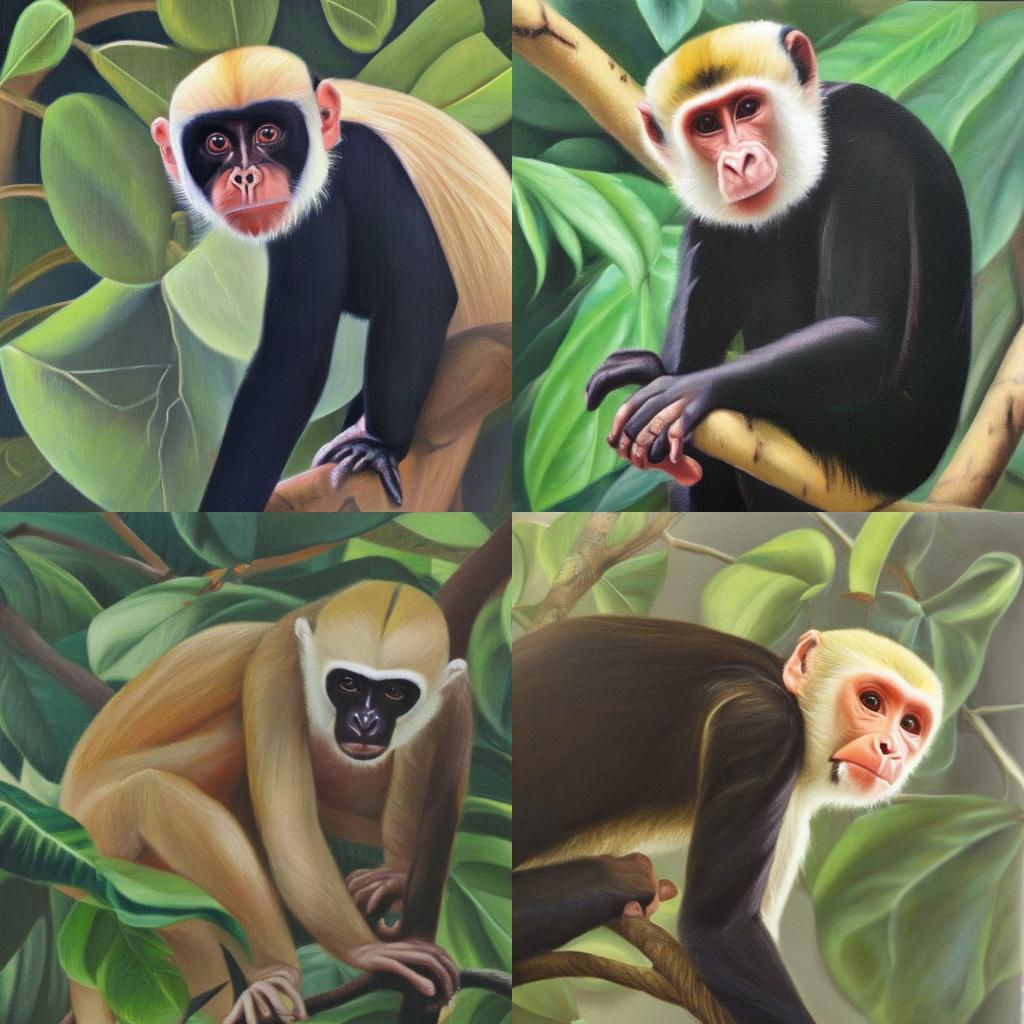}}
            \centerline{\texttt{capuchin}}
    	\label{fig:capuchin}
	\end{minipage}
         \begin{minipage}[]{0.32\columnwidth}
            \centering
    	{\includegraphics[width = \columnwidth]{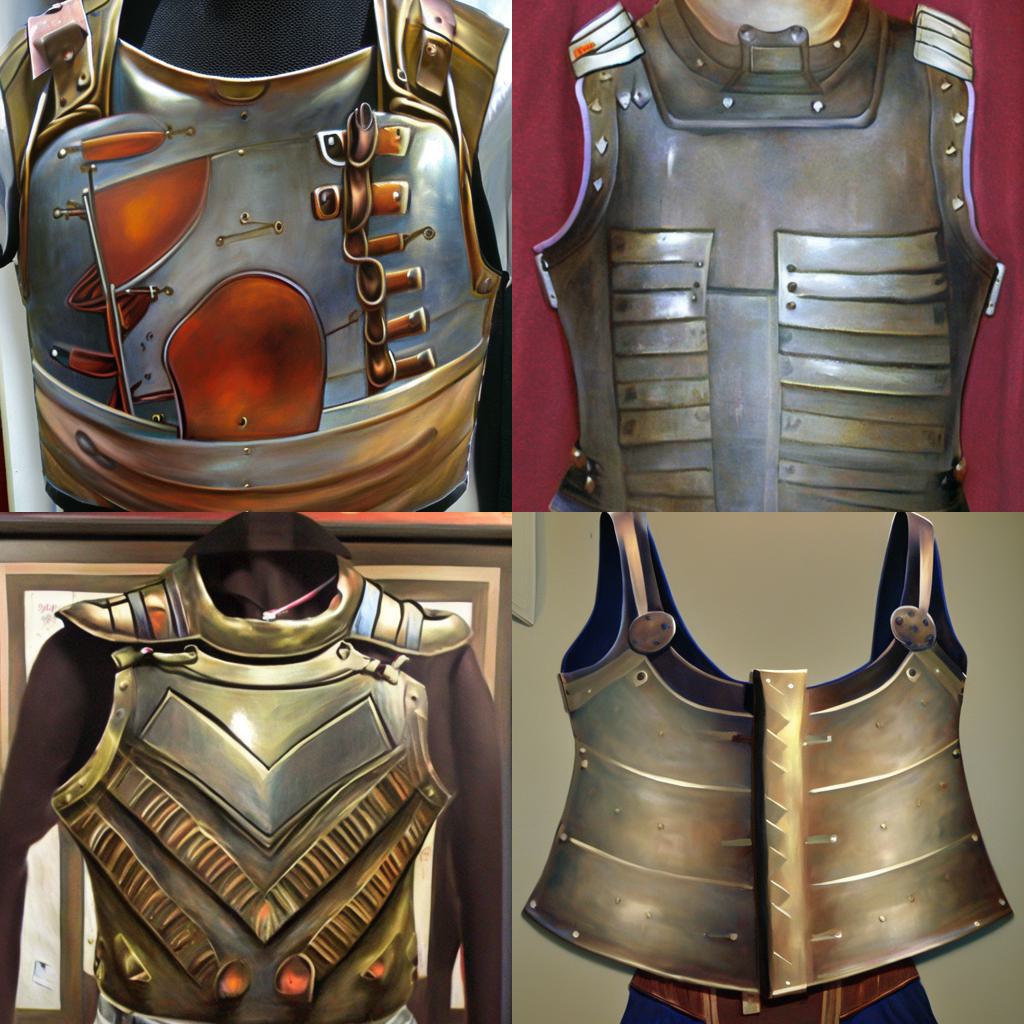}}
            \centerline{\texttt{breastplate}}
    	\label{fig:breastplate}
	\end{minipage}
        \hfill
             \begin{minipage}[]{0.32\columnwidth}
            \centering
    	{\includegraphics[width = \columnwidth]{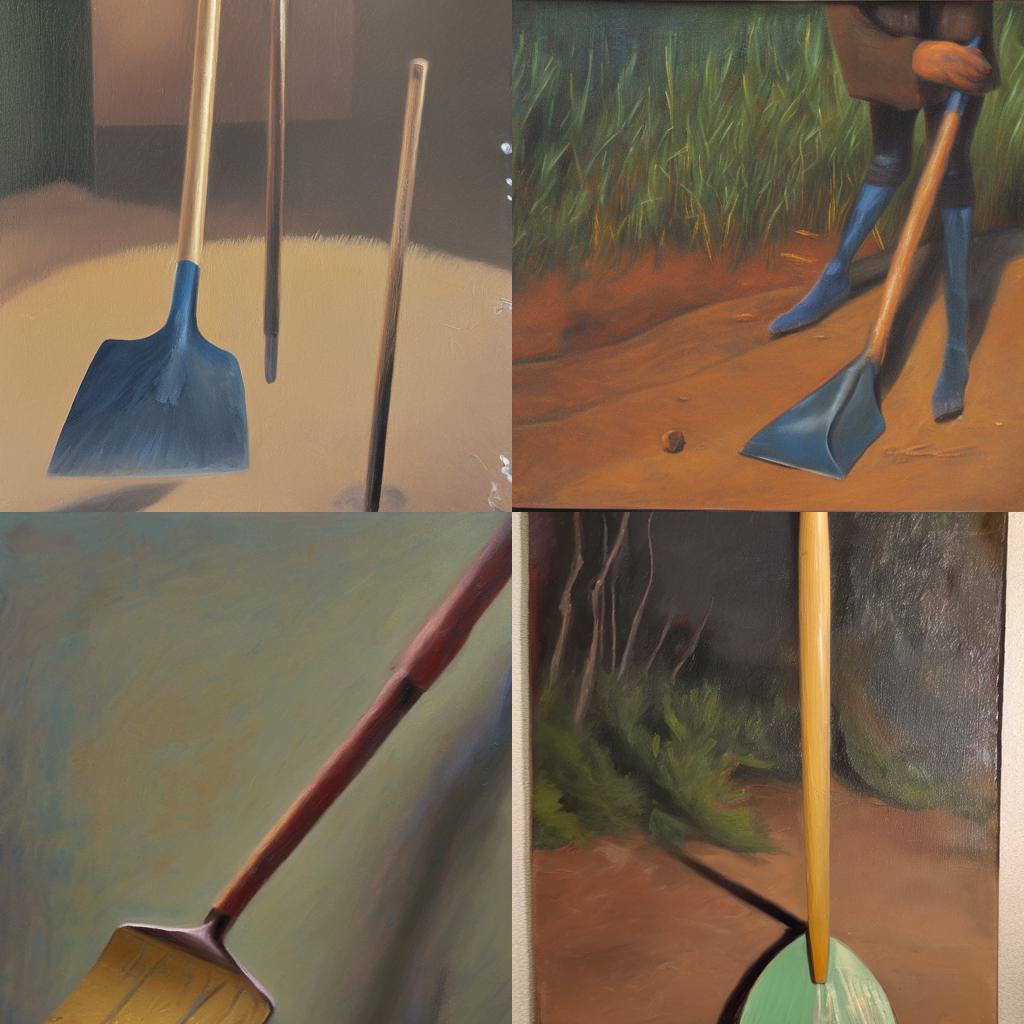}}
            \centerline{\texttt{shovel}}
            \label{fig:shovel}
	\end{minipage}
        \hfill
         \begin{minipage}[]{0.32\columnwidth}
            \centering
    	{\includegraphics[width = \columnwidth]{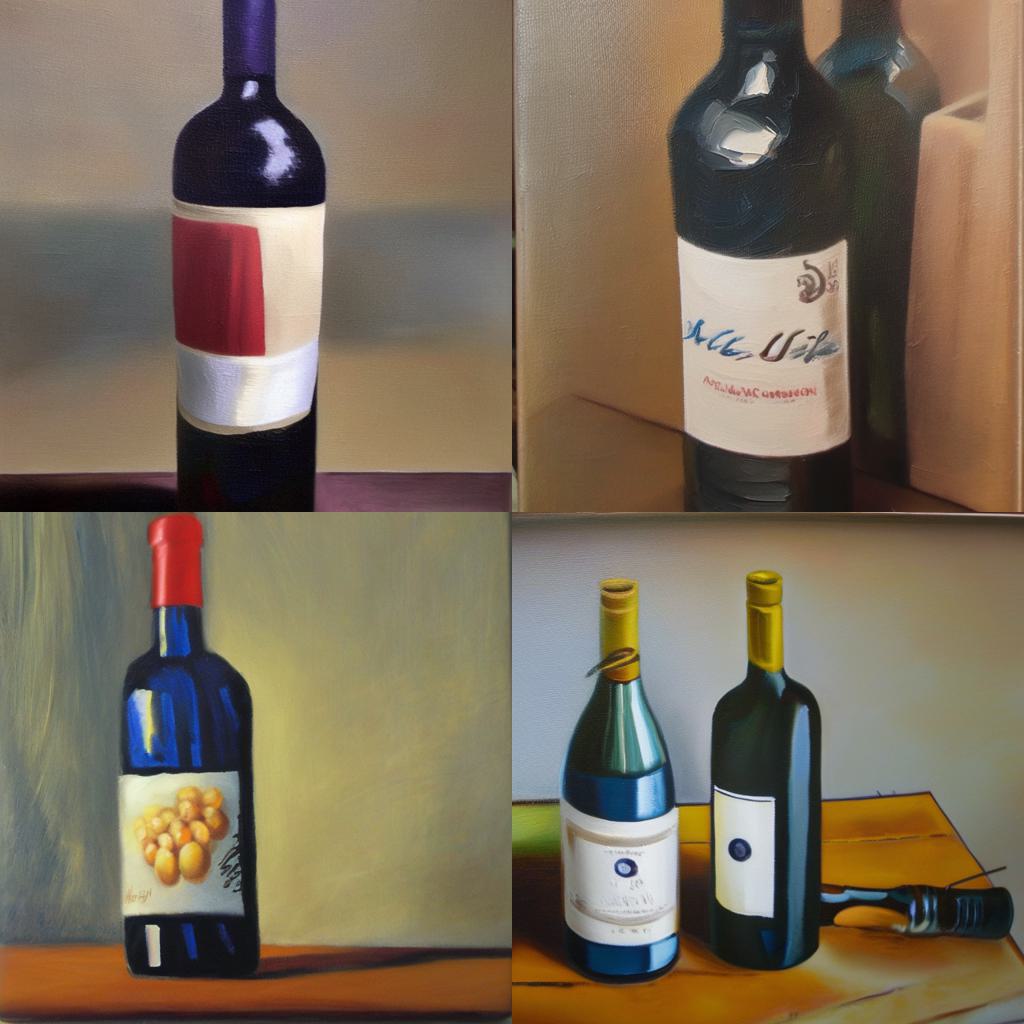}}
            \centerline{\texttt{wine bottle}}
    	\label{fig:wine_bottle}
	\end{minipage}
        \caption{Visual examples of the AI-generated oil painting images.}
        \label{fig:op_example}
\end{figure}

\begin{figure}[!htbp]
    \renewcommand\thefigure{D}
        \centering
         \begin{minipage}[]{0.32\columnwidth}
            \centering
    	{\includegraphics[width = \columnwidth]{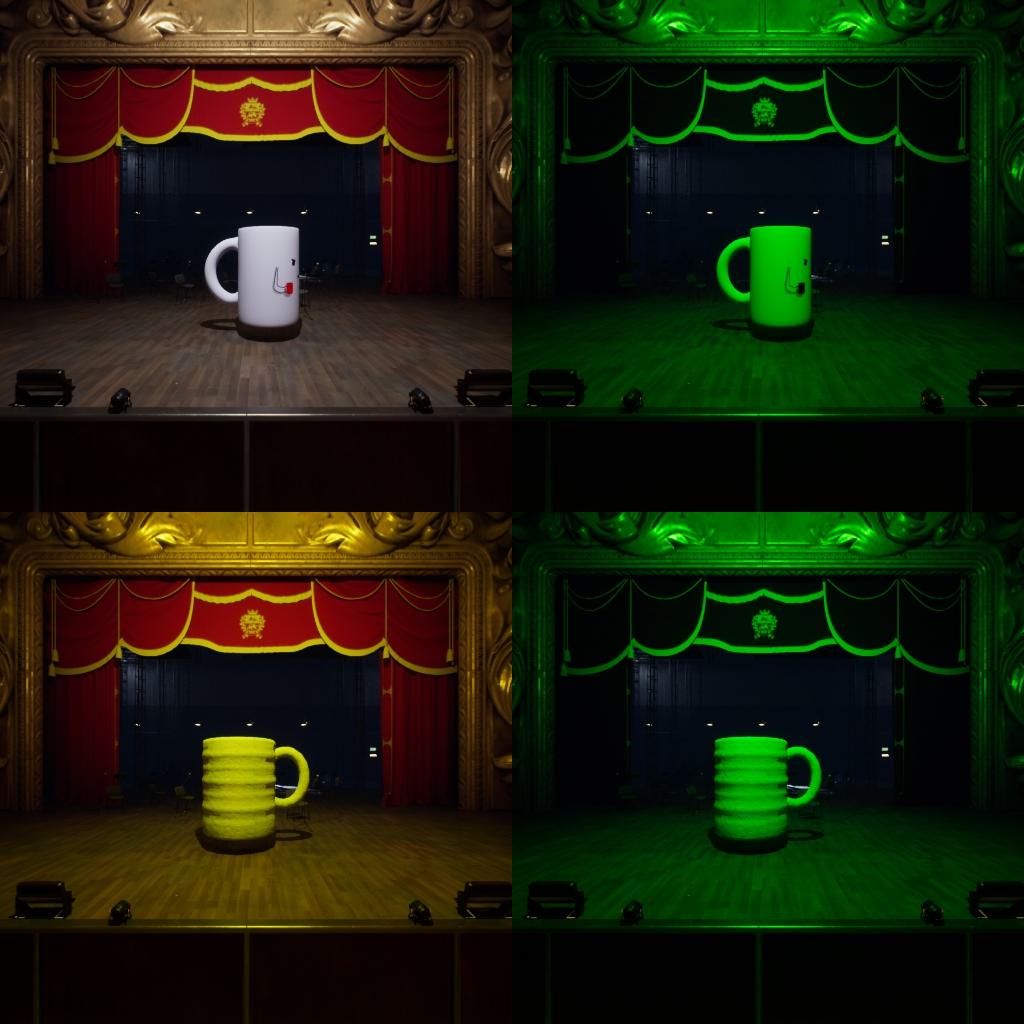}}
            \centerline{\texttt{coffee mug}}
    	\label{fig:Cofee_Mug}
	\end{minipage}
        \hfill
             \begin{minipage}[]{0.32\columnwidth}
            \centering
    	{\includegraphics[width = \columnwidth]{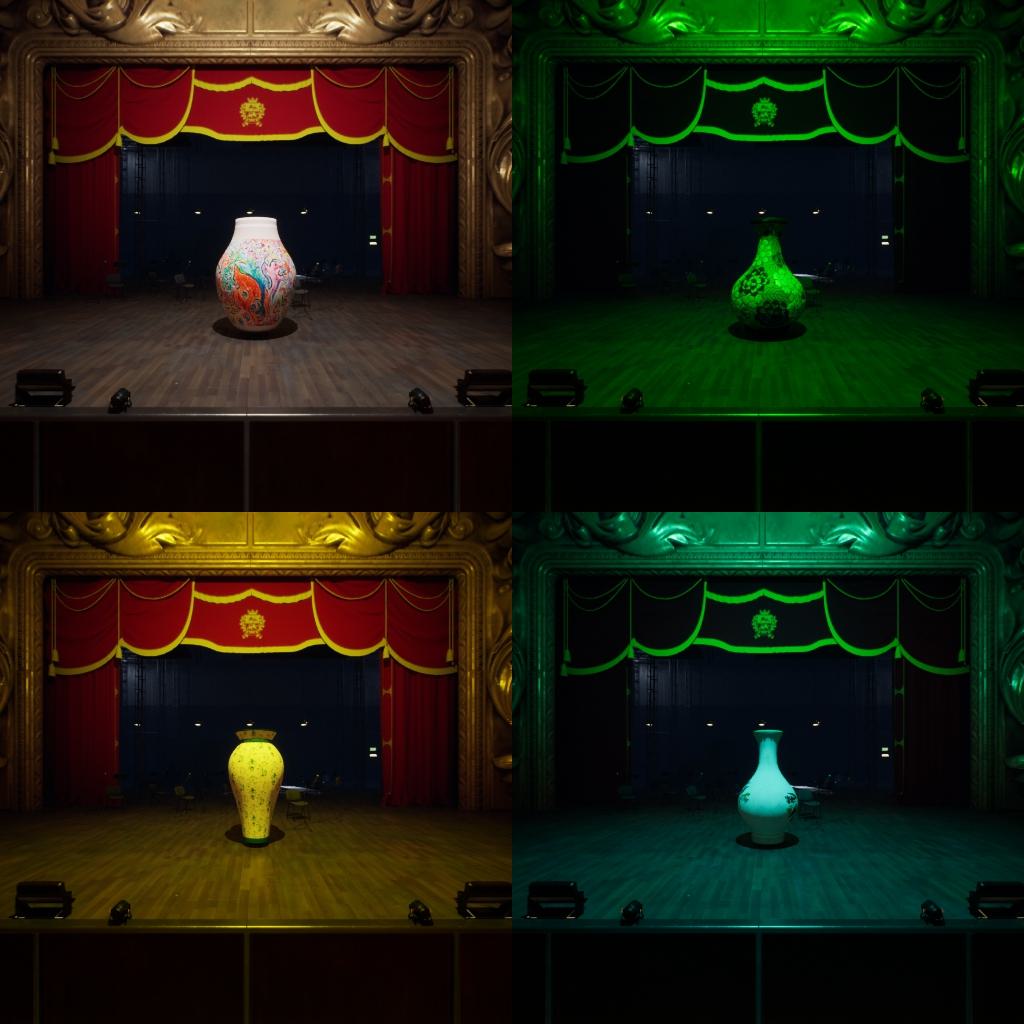}}
            \centerline{\texttt{vase}}
            \label{fig:Vase}
	\end{minipage}
        \hfill
         \begin{minipage}[]{0.32\columnwidth}
            \centering
    	{\includegraphics[width = \columnwidth]{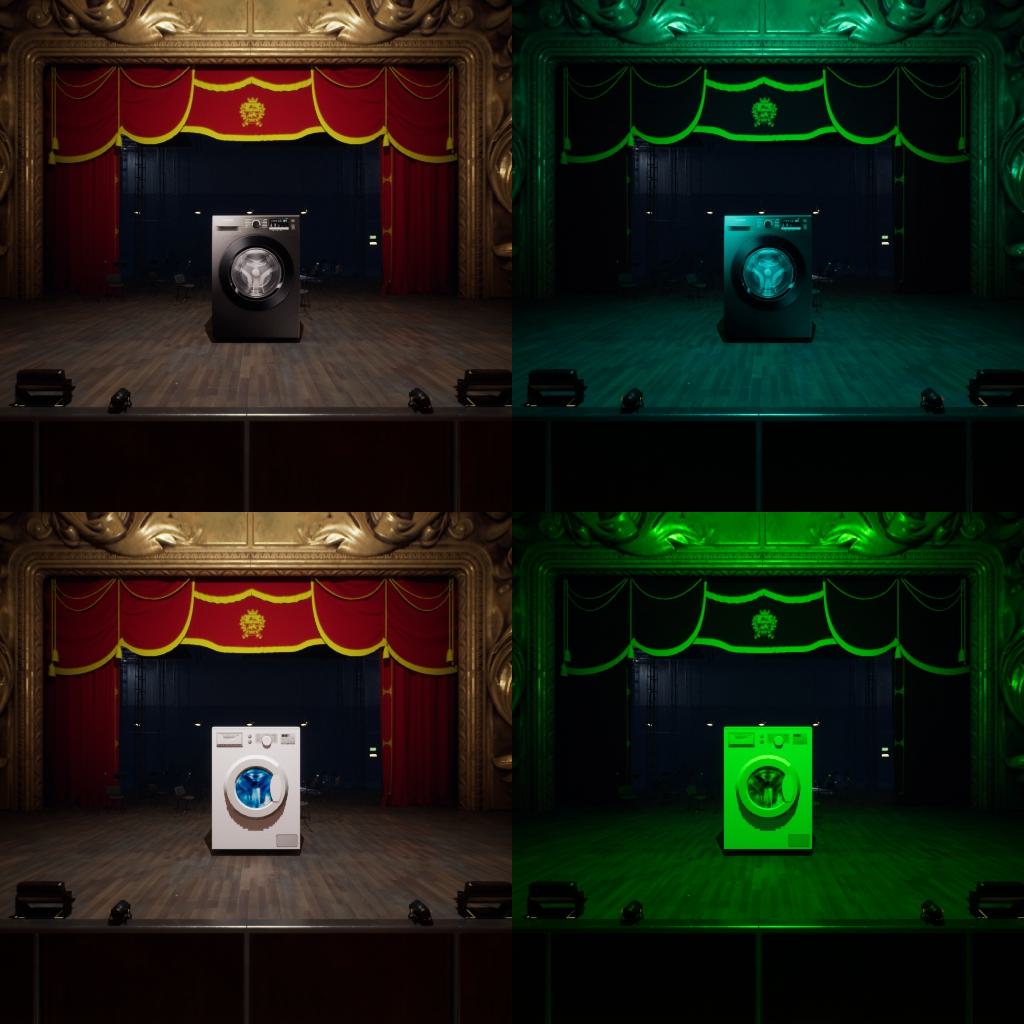}}
            \centerline{\texttt{washer}}
    	\label{fig:Washer}
	\end{minipage}        \caption{Visual examples of the AI-generated stage light images.}
        \label{fig:sl_example}
\end{figure}

Visual examples of the oil painting and stage light images are shown in Fig.~\ref{fig:op_example} and Fig.~\ref{fig:sl_example}, respectively.

\begin{table}[t]
        \renewcommand\thetable{B}
	\caption{Statistics of the AI-generated image subset.}
	\label{app_table:ai_dataset_statistics}
	\centering
        {\begin{tabular}{ccc} 
		\toprule
		  Group &  Class Number  & Sample Number \\
		\hline
		oil painting &  22 & 860 \\
		stage light & 32 &1,092 \\
		\bottomrule
	\end{tabular}}
\end{table}

\subsection{Potential Dataset Filtering} 
Recall that the editing benchmark is designed to challenge the ViT/B-16 model. Thus, it is likely that some images might not induce predictive errors in other base models, which vary in terms of training data, model architecture, loss function, and optimization pipeline.  For the ViT/S-16 model, the benchmark is subject to an additional filtering process based on its predictions. Consequently,  $65\%$ of the natural images and $100\%$ of the AI-generated images are retained.

\section{More Experimental Details}
In this section, we give more implementation details of the proposed and competing model editing methods. Algorithm~\ref{algo:meta-training} presents the pseudo-code of our method.


\begin{algorithm}[t]
	\caption{Hypernetwork Meta-training via \textcolor{magenta}{Standard Implementation} / \textcolor{teal}{Decoupling Trick}}
	\label{algo:meta-training}
	\begin{algorithmic}[1]
		\REQUIRE  Hypernetwork, $g(\cdot;\varphi)$; ViT feature extractor,  $e\left(\cdot;\phi^{(0)}\right)$; CutMix dataset, $\mathcal{B}$; training iteration, $\mathrm{MaxIter}$;  inner-loop learning rate, $\alpha$; inner-loop step, $T$; outer-loop hypernetwork learning rate, $\beta$;  outer-loop $\tilde{{m}}$ learning rate, $\gamma$;  outer-loop step, $\mathrm{MaxOuterIter}$; trade-off parameter, $\lambda$
        \STATE Randomly initialize $\varphi$
		\FOR{$\mathrm{MainIter}$ = $1$ to $\mathrm{MaxIter}$}  
			\STATE Create a CutMix dataset $\mathcal{B}$ from ImageNet-1k
                \FOR{$(x',x)\in \mathcal{B}$} 
                    \STATE Calculate $\hat{m} = g\left(e\left(x;\phi^{(0)}\right);\varphi\right)$ \hfill \textcolor{gray}{\small{// Approximated by Eq.~\eqref{eq:qa}}}
                    \STATE Set $\Delta \phi^{(0)} = 0$
                    \textcolor{magenta}{\FOR{$t = 1$ to $T$}  
                        \STATE $\Delta{ \phi}^{(t)}  = \Delta{ \phi}^{(t-1)} - \alpha \nabla_{ \phi} \ell\left( x, {p\left(y| {x'};\phi^{(0)}\right)};  \phi^{(t-1)}\right)$
                         \STATE  ${\phi}^{(t)} = { \phi}^{(0)} + \hat{m} \odot \Delta{\phi}^{(t)}$
                    \ENDFOR
                    \STATE $ {\varphi}  \leftarrow {\varphi} - \beta \nabla_{ \varphi} \left[\ell\left({x},  {p\left(y| {x'};\phi^{(0)}\right)};\phi^{(T)}\right)+ \lambda\Vert  \hat{m}\Vert_1\right]$}
                    \textcolor{teal}{\STATE Randomly initialize  $\tilde{m}$}
                    \textcolor{teal}{\FOR{$\mathrm{OuterIter =1}$  to $\mathrm{MaxOuterIter}$}  
                        \FOR{ $t=1$ to $T$}  
                            \STATE $\Delta{ \phi}^{(t)}  = \Delta{ \phi}^{(t-1)} - \alpha \nabla_{ \phi} \ell\left( x, {p\left(y| {x'};\phi^{(0)}\right)}  \phi^{(t-1)}\right)$
                             \STATE  ${\phi}^{(t)} = { \phi}^{(0)} + \tilde{{m}} \odot \Delta{\phi}^{(t)}$
                        \ENDFOR
                    \STATE $ \tilde{{m}} \leftarrow \tilde{{m}} - \gamma \nabla_{\tilde{{m}}} \left[\ell\left({x},  {p\left(y| {x'};\phi^{(0)}\right)};\phi^{(T)}\right)+ \lambda\Vert \tilde{{m}}\Vert_1\right]$
                    \ENDFOR}
                \ENDFOR
                \textcolor{teal}{\STATE $ {\varphi}  \leftarrow {\varphi} - \beta \nabla_{ \varphi}\mathrm{KL}\left(\hat{m}, \tilde{m}\right)$}
		\ENDFOR
	\end{algorithmic}
\end{algorithm}

\subsection{More Details of Our Method} \label{app:method_details}
\paragraph{Decoupling Trick} In meta-learning, optimization of the hypernetwork entails 
differentiating the outer-loop loss with respect to the output of the inner loop  ${\phi}^{(T)}$, and propagating the gradient through the inner-loop optimization to the output of the hypernetwork $\hat{m}$ (approximated by Eq.~\eqref{eq:qa}), and finally to the parameters of the hypernetwork, $\varphi$. This extended chain of computation not only demands substantial computational resources but also hampers efficient optimization. To mitigate these, we decouple the pathway of hypernetwork optimization from the meta-learning gradient. Specifically, we introduce an auxiliary variable $\tilde{{m}}$, matching the dimensionality of $\hat{m}$, to substitute for the hypernetwork's output during bi-level optimization. As a result, ${\phi}^{(T)}$ is now dependent on  $\tilde{{m}}$, rather than $\hat{m}$.
We first optimize the auxiliary variable:
\begin{align}
     \tilde{{m}}^\star = \arg\min_{ \tilde{{m}}} \ell\left(x,y;\phi^{(T)}\right)+ \lambda \Vert \tilde{m}\Vert_1 .
\end{align}
Subsequently, $\tilde{{m}}^\star$ directs the parameter optimization of the hypernetwork using the element-wise KL divergence averaged across all positions: 
\begin{align}
   {\varphi}^\star = \arg\min_{{\varphi}}\frac{1}{\mathrm{dim}(\tilde{m}^\star)} \sum_i\mathrm{KL}\left(g_i\left(e\left(x;\phi^{(0)}\right);\varphi\right), \tilde{m}_i^\star\right),
\end{align}
where $i$ is the positional index and $\mathrm{dim}(\tilde{m}^\star)=N_m\times 6$ in our implementation.

\paragraph{Pseudo-sample Generation} When applying CutMix, we vary the sizes of the pasted patches from $48 \times 48$ to $128 \times 128$, ensuring the preservation of the primary structural and textural details in the original images, which are $224\times 224$ in size.

\paragraph{Hypernetwork Architecture} We design the hypernetwork to mirror the architecture of its corresponding base model (\ie, ViT/B-16 or ViT/S-16), with the same input and intermediate dimensions. Nevertheless, we reduce the number of attention blocks to five.

\paragraph{Hyperparameter Configuration} We set the learning rate in the inner loop as $0.001$, and perform gradient descent for five steps (\ie, $T=5$). In the outer loop, we apply the Adam optimizer with a learning rate of $0.1$ to optimize $\tilde{{m}}$ from random initialization for a total of ten steps.
For the hypernetwork optimization, RMSProp\footnote{\url{https://www.cs.toronto.edu/~tijmen/csc321/slides/lecture_slides_lec6.pdf}} is utilized with a learning rate of $10^{-4}$, a minibatch size of eight, and a maximum iteration number of $7,000$. Training a hypernetwork for the base ViT/B-16 takes approximately 9 hours on a single RTX A6000 GPU (48G).

\subsection{Implementation Details of Competing Methods}\label{app:exp_details}
For methods that involve updating the base model parameters through backpropagation---including FT, FT-$\ell_2$, KN~\cite{dai2022knowledge}, SPT~\cite{he2023sensitivity}, and our method---we follow~\cite{de2021editing} and adopt RMSProp as the optimizer, where the learning rate is set to $2 \times 10^{-5}$ for ViT/B-16 and $10^{-4}$ for ViT/S-16, respectively. 

T-Patcher~\cite{huang2023transformer} adds one neuron in the last FFN, together with a trainable multiplier initialized as $10$. The new parameters are optimized using Adam with a learning rate of $5\times10^{-3}$.

ROME~\cite{meng2022locating} employs Adam with a learning rate of $0.01$ to obtain the target hidden representations of the last FFN, and then solves a constrained least squares problem to update the second FC layer.

We follow the default setting in LoRA~\cite{hu2022lora}, adding learnable matrices with a rank of eight. These low-rank matrices are optimized by  Adam with a learning rate of $10^{-4}$. 

For KE~\cite{de2021editing} and MEND~\cite{de2021editing}, we adhere to their training protocols to edit the six FC layers within the last three FFNs. The hypernetworks are meta-trained on editing samples sourced from ImageNet-1k  to alter the base model's predictions to match the top-$k$ randomly selected classes. The optimizer is  Adam~\cite{kingma2015adam} with a learning rate of $10^{-5}$. 

\begin{figure} 
\renewcommand\thefigure{E}
    \begin{subfigure}[]{0.24\textwidth}
        \centering
        \includegraphics[width = \columnwidth]{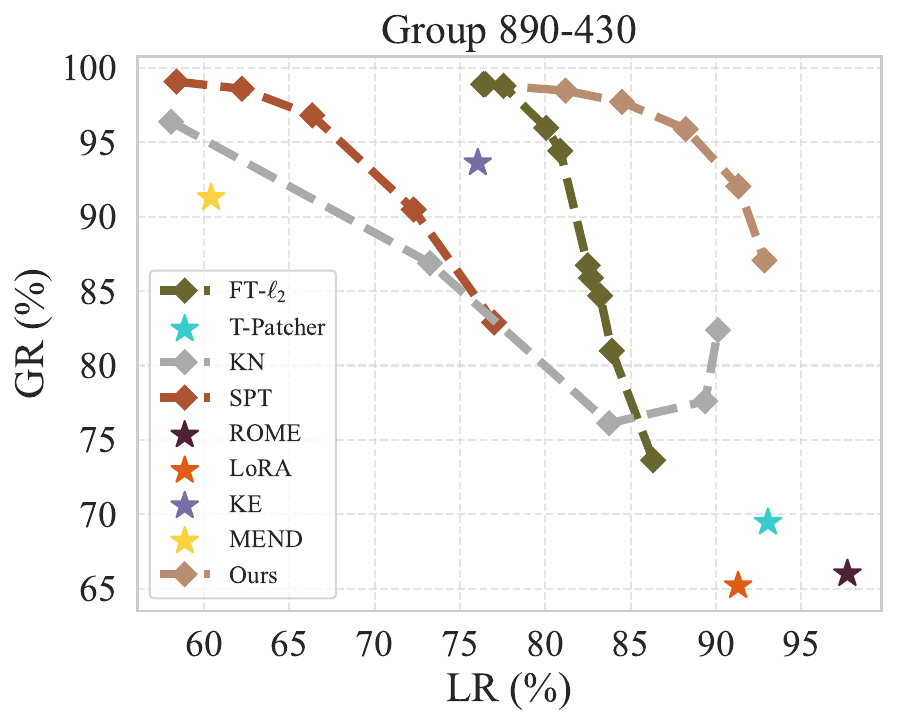}
    \end{subfigure}
    \hfill    
    \begin{subfigure}[]{0.24\textwidth}
        \centering
        \includegraphics[width = \columnwidth]{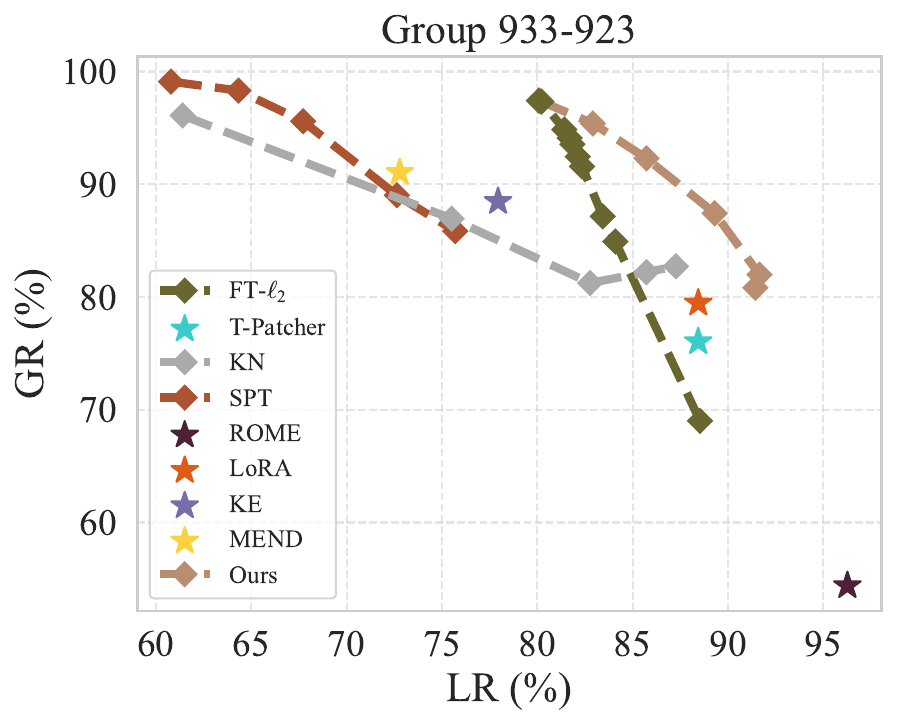}
    \end{subfigure}
    \hfill
    \begin{subfigure}[]{0.24\textwidth}
        \centering
        \includegraphics[width = \columnwidth]{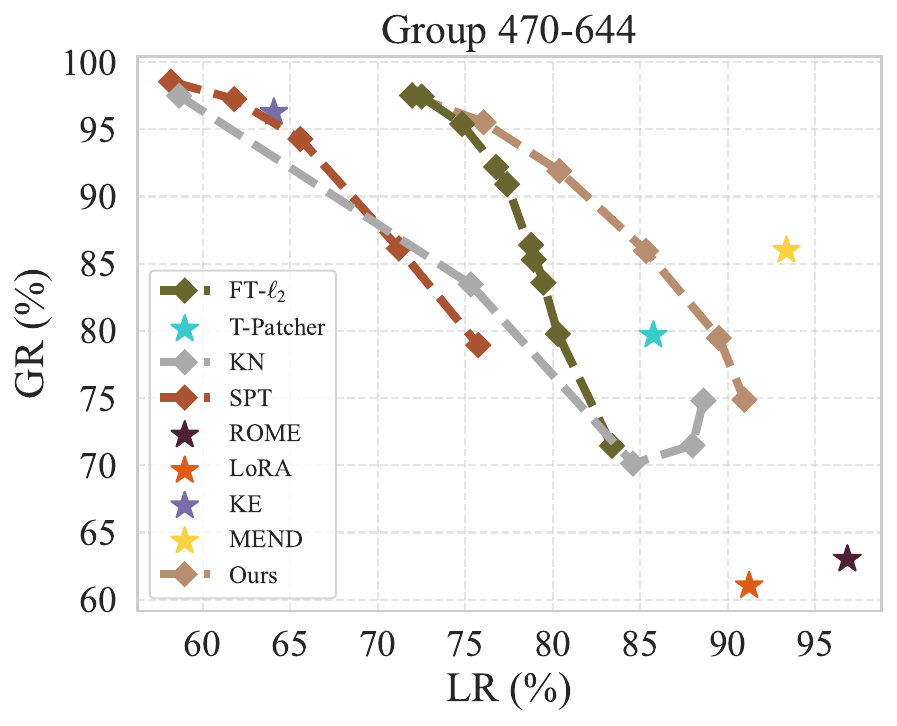}
    \end{subfigure}
    \hfill
    \begin{subfigure}[]{0.24\textwidth}
    \centering
    \includegraphics[width = \columnwidth]{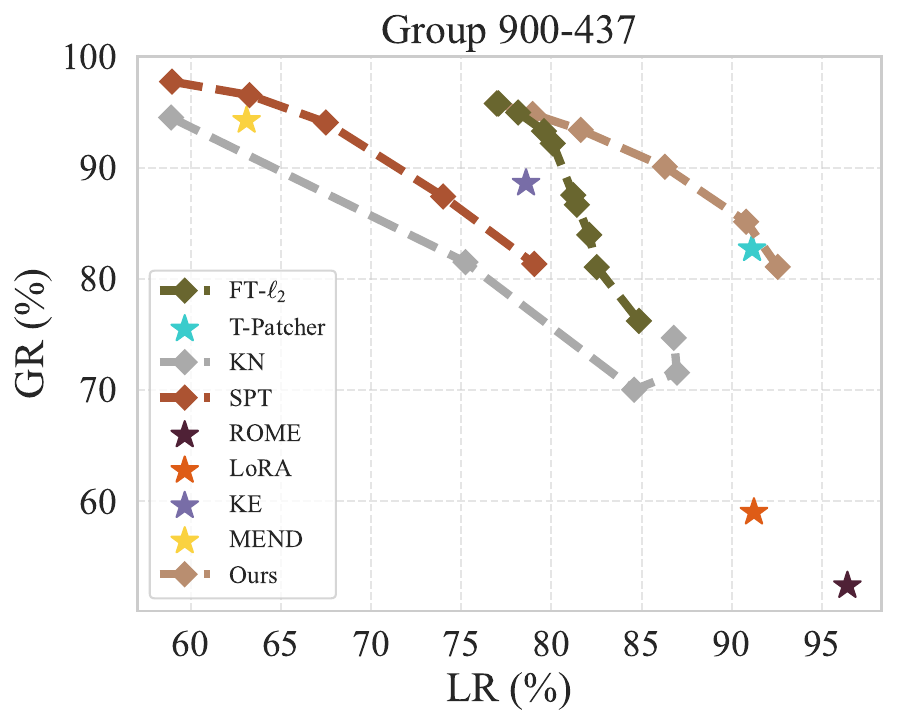}
    \end{subfigure}
    \begin{subfigure}[]{0.24\textwidth}
        \centering
        \includegraphics[width = \columnwidth]{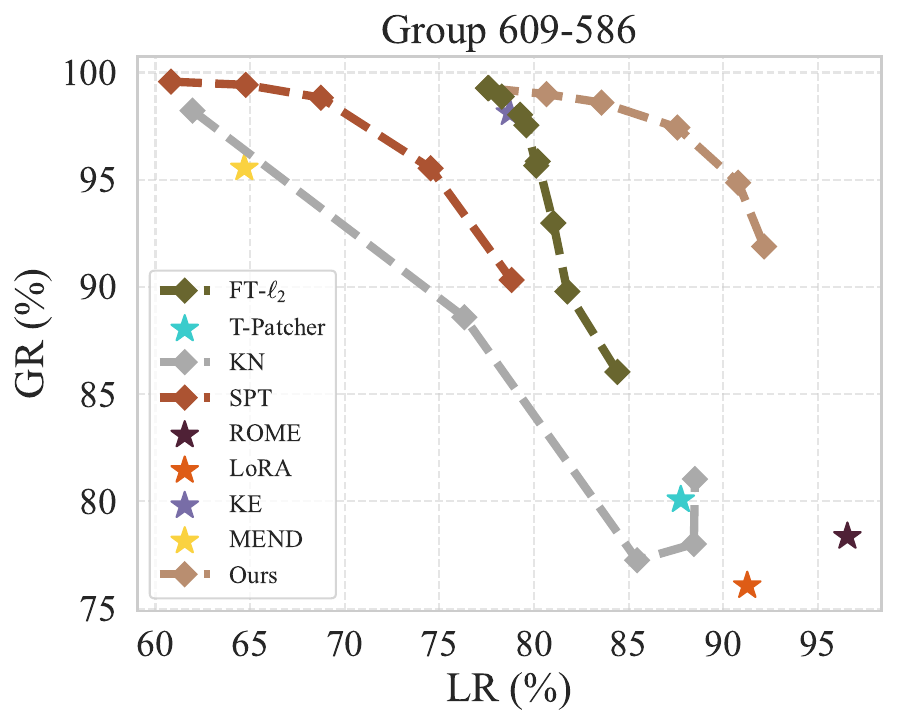}
    \end{subfigure}
    \hfill    
    \begin{subfigure}[]{0.24\textwidth}
        \centering
        \includegraphics[width = \columnwidth]{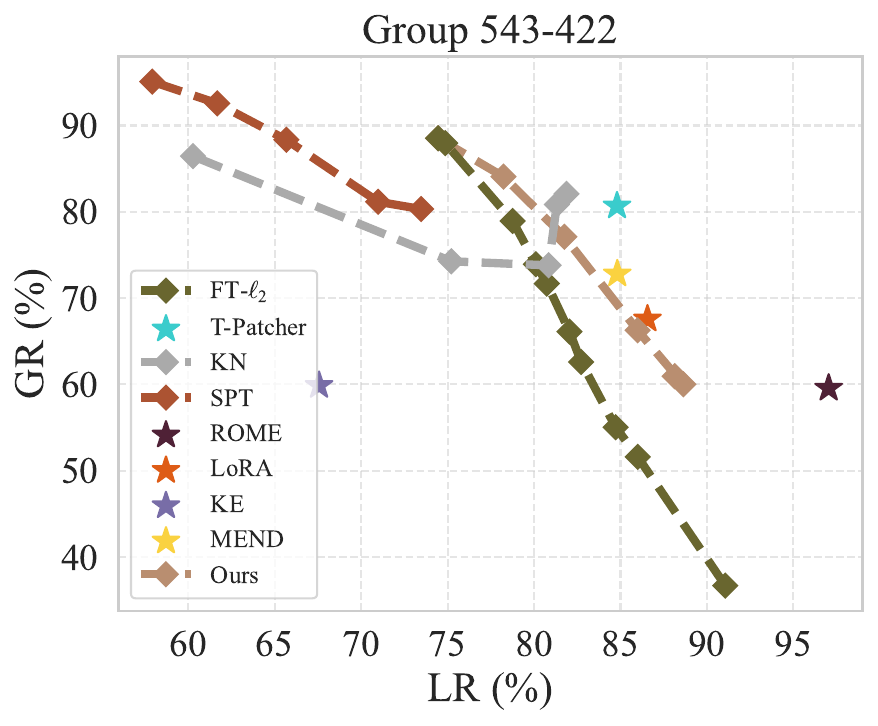}
    \end{subfigure}
    \hfill
    \begin{subfigure}[]{0.24\textwidth}
        \centering
        \includegraphics[width = \columnwidth]{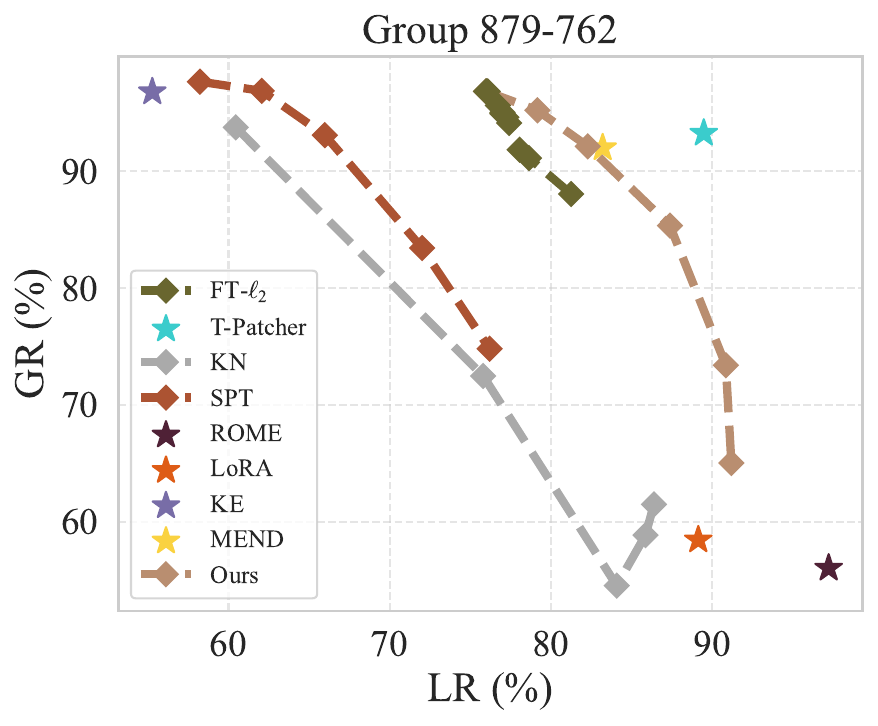}
    \end{subfigure}
    \hfill
    \begin{subfigure}[]{0.24\textwidth}
        \centering
        \includegraphics[width = \columnwidth]{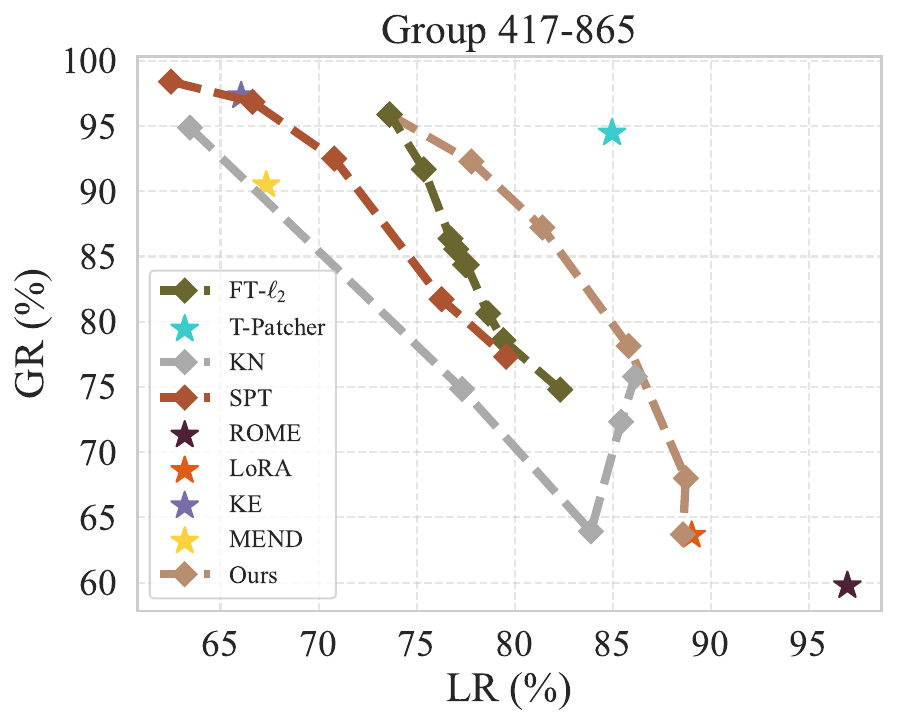}
    \end{subfigure}
    \begin{subfigure}[]{0.24\textwidth}
        \centering
        \includegraphics[width = \columnwidth]{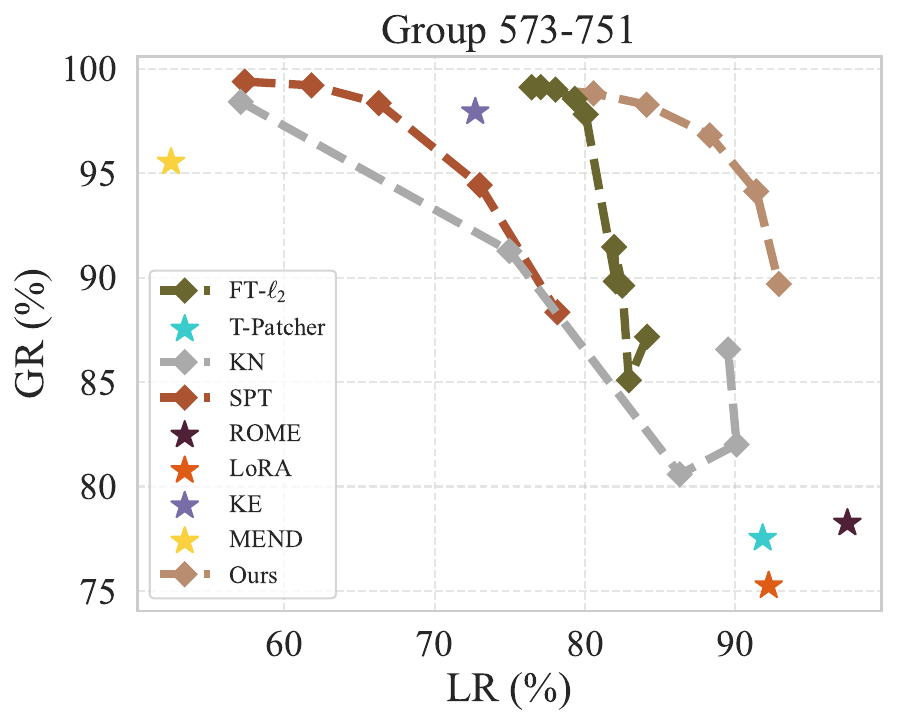}
    \end{subfigure}
    \hfill    
    \begin{subfigure}[]{0.24\textwidth}
        \centering
        \includegraphics[width = \columnwidth]{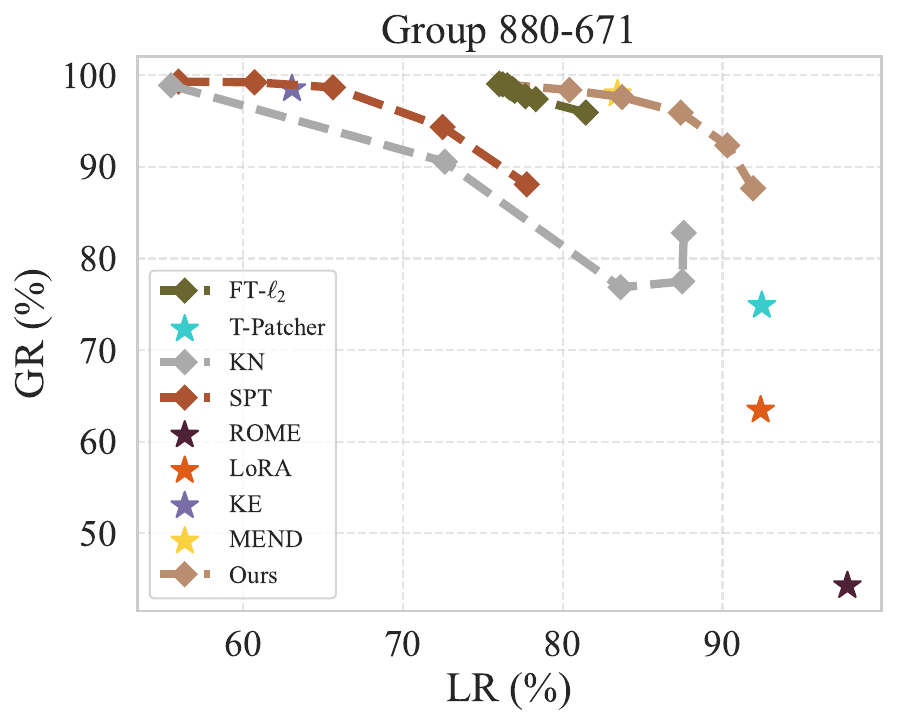}
    \end{subfigure}
    \hfill
    \begin{subfigure}[]{0.24\textwidth}
        \centering
        \includegraphics[width = \columnwidth]{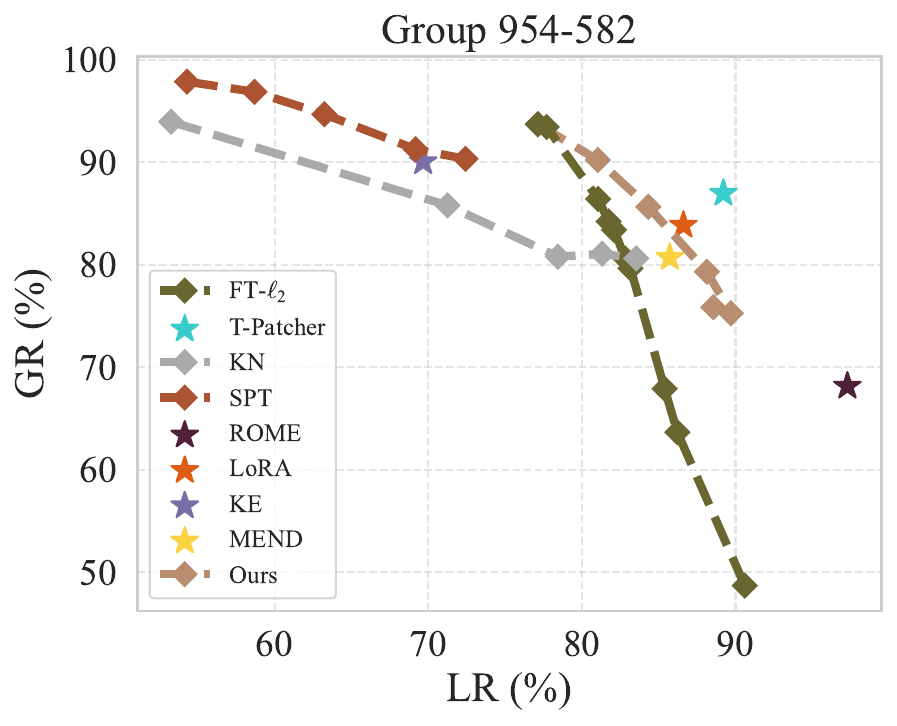}
    \end{subfigure}
    \hfill
    \begin{subfigure}[]{0.24\textwidth}
        \centering
        \includegraphics[width = \columnwidth]{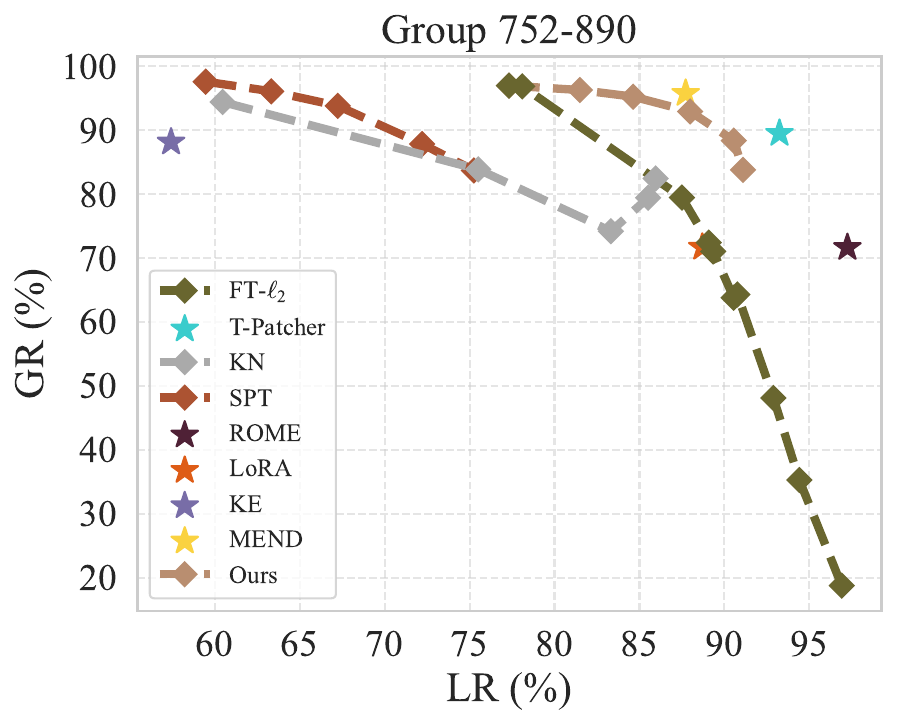}
    \end{subfigure}
            \begin{subfigure}[]{0.24\textwidth}
        \centering
        \includegraphics[width = \columnwidth]{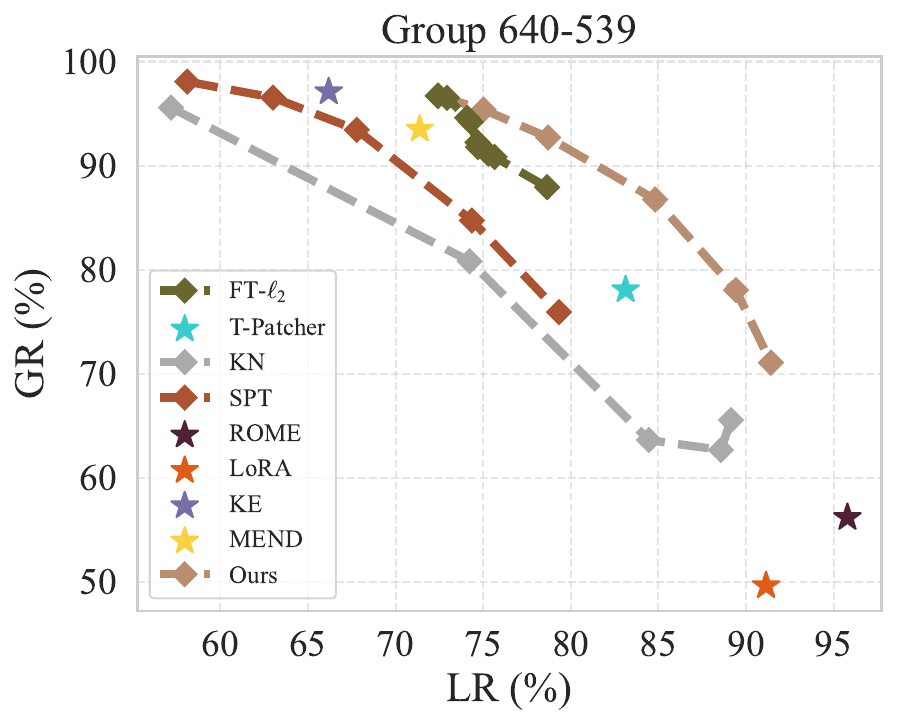}
    \end{subfigure}
    \hfill    
    \begin{subfigure}[]{0.24\textwidth}
        \centering
        \includegraphics[width = \columnwidth]{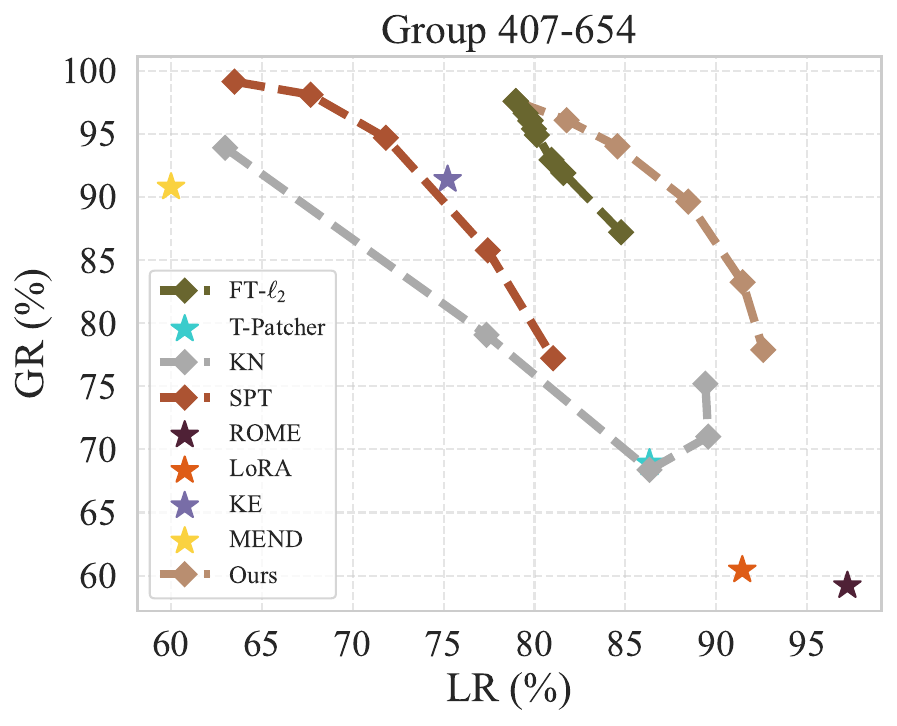}
    \end{subfigure}
    \hfill
    \begin{subfigure}[]{0.24\textwidth}
        \centering
        \includegraphics[width = \columnwidth]{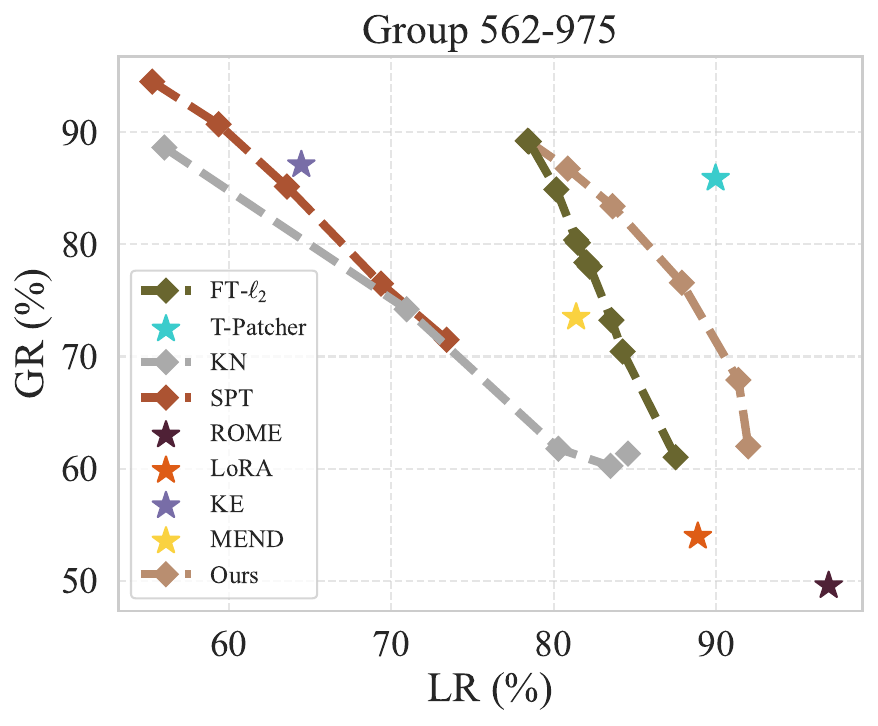}
    \end{subfigure}
    \hfill
    \begin{subfigure}[]{0.24\textwidth}
        \centering
        \includegraphics[width = \columnwidth]{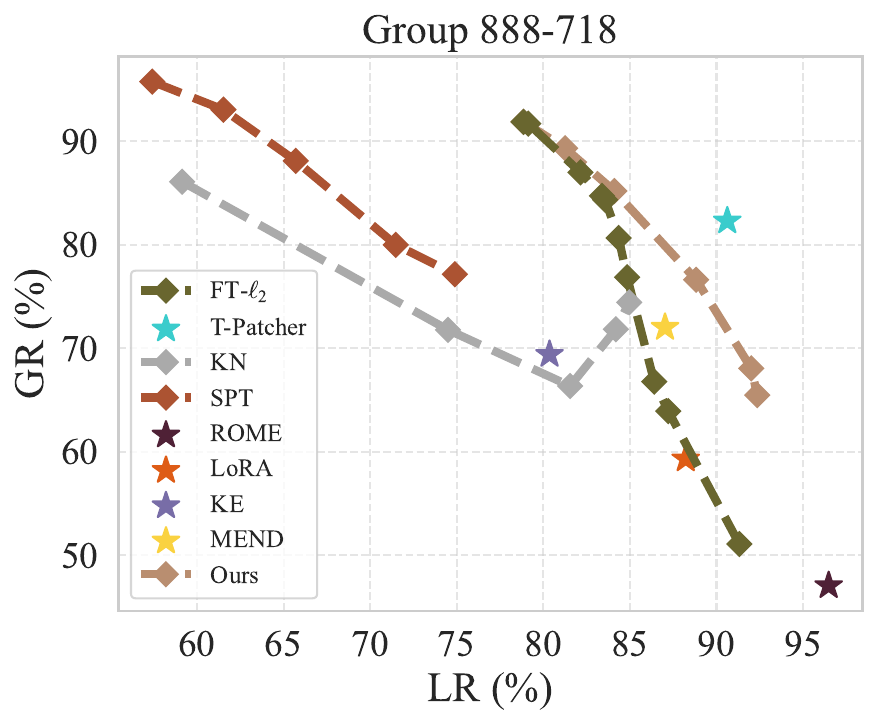}
    \end{subfigure}
\caption{Editing results for ViT/B-16 on the sixteen groups in the natural image subset.}
\label{app_fig:exp_curves_u_b}
\end{figure}

\begin{figure}[t]
\renewcommand\thefigure{F}
        \centering
        \includegraphics[width = \columnwidth]{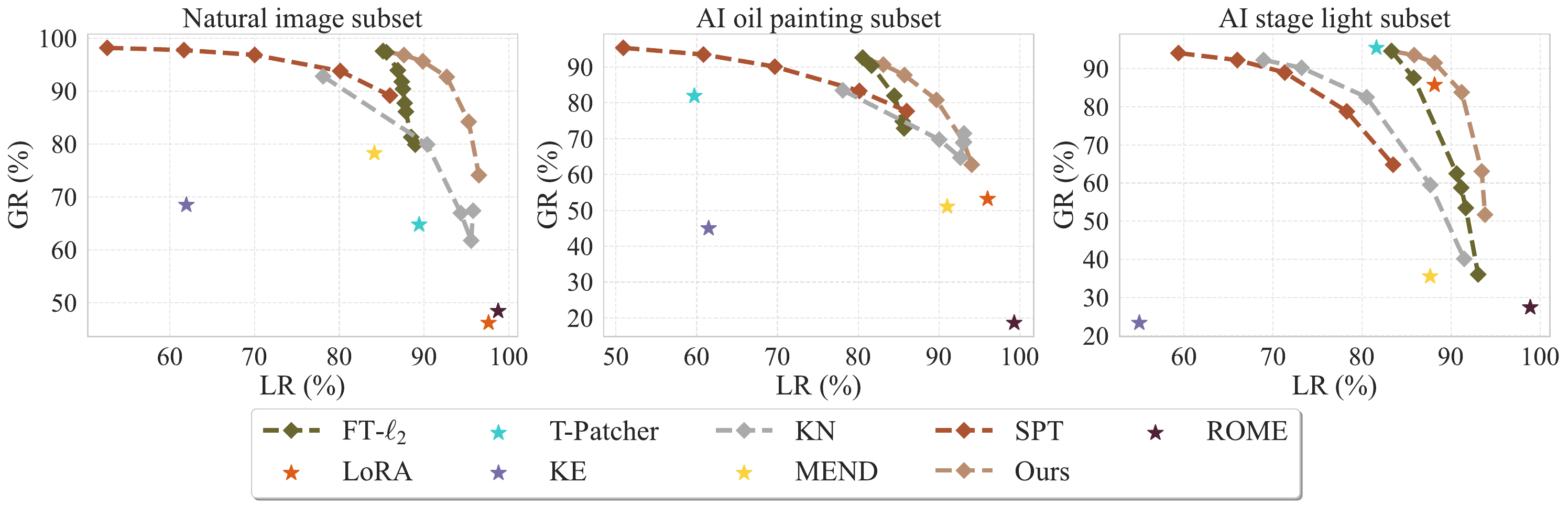}
        \caption{Editing results for ViT/S-16 on the proposed benchmark.}
        \label{app_fig:exp_curves_vits}
\end{figure}

\begin{figure} 
\renewcommand\thefigure{G}
    \begin{subfigure}[]{0.24\textwidth}
        \centering
        \includegraphics[width = \columnwidth]{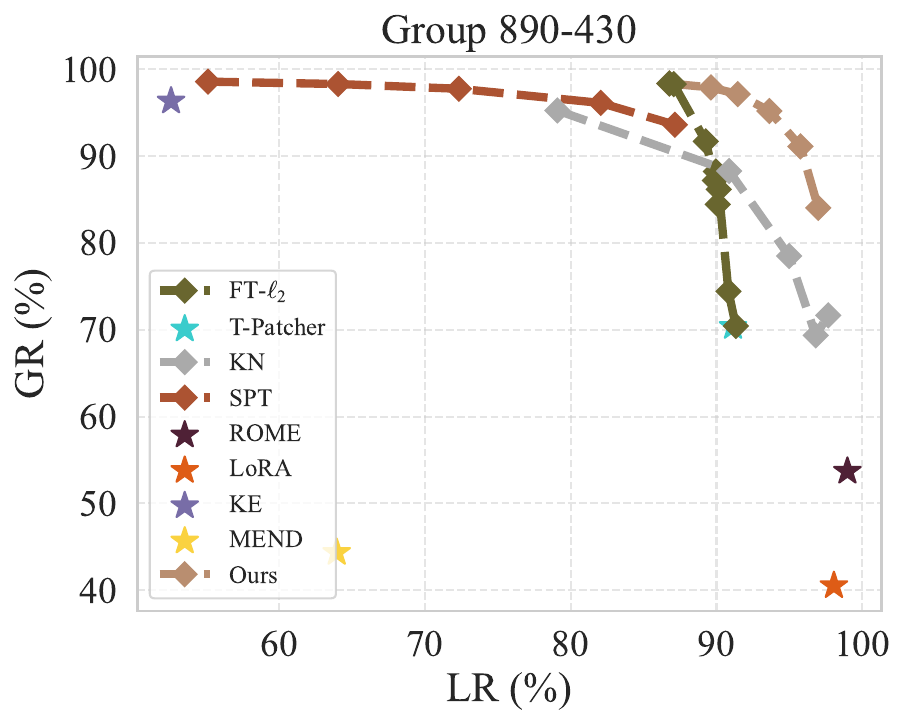}
    \end{subfigure}
    \hfill    
    \begin{subfigure}[]{0.24\textwidth}
        \centering
        \includegraphics[width = \columnwidth]{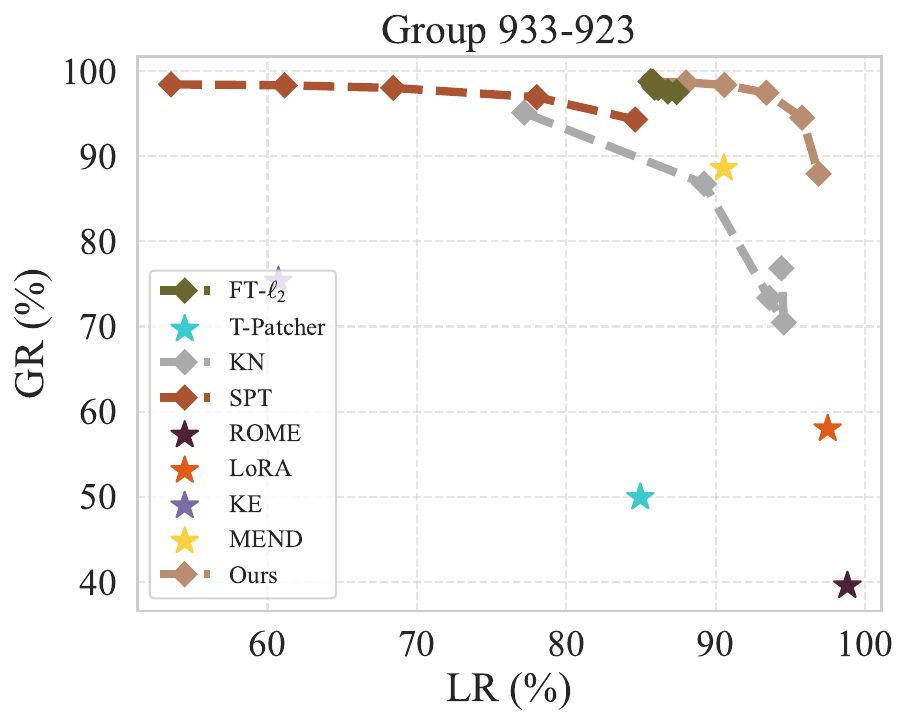}
    \end{subfigure}
    \hfill
    \begin{subfigure}[]{0.24\textwidth}
        \centering
        \includegraphics[width = \columnwidth]{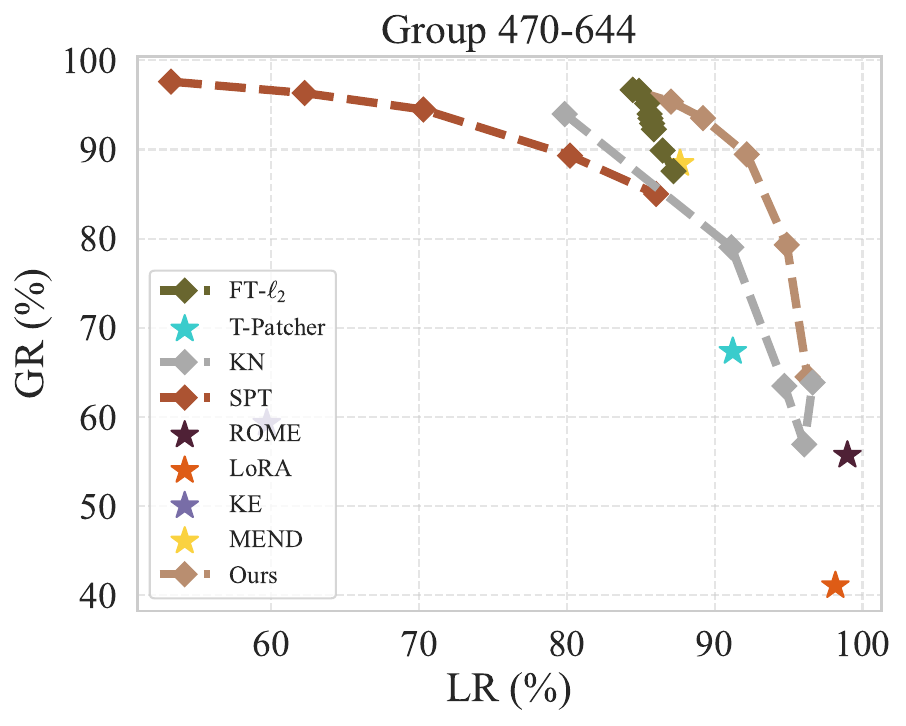}
    \end{subfigure}
    \hfill
    \begin{subfigure}[]{0.24\textwidth}
    \centering
    
    \includegraphics[width = \columnwidth]{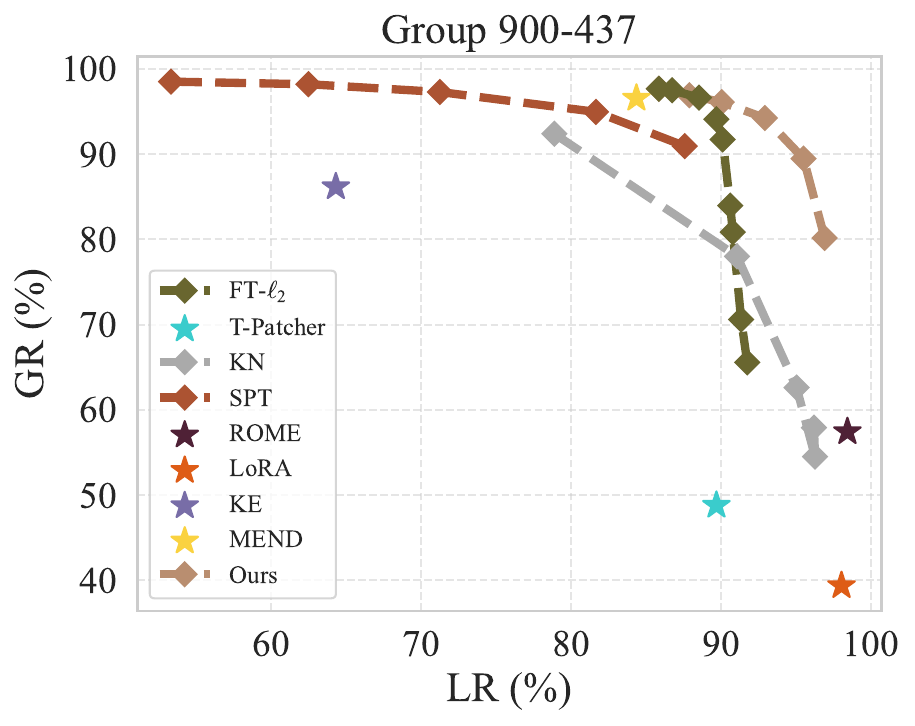}
    \end{subfigure}
    \begin{subfigure}[]{0.24\textwidth}
        \centering
        \includegraphics[width = \columnwidth]{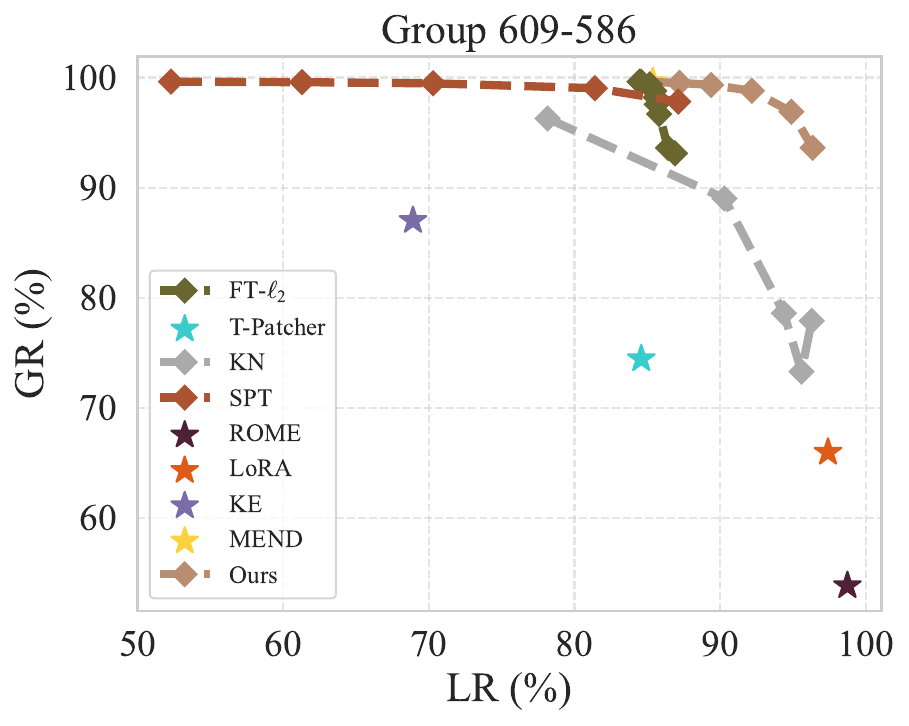}
    \end{subfigure}
    \hfill    
    \begin{subfigure}[]{0.24\textwidth}
        \centering
        \includegraphics[width = \columnwidth]{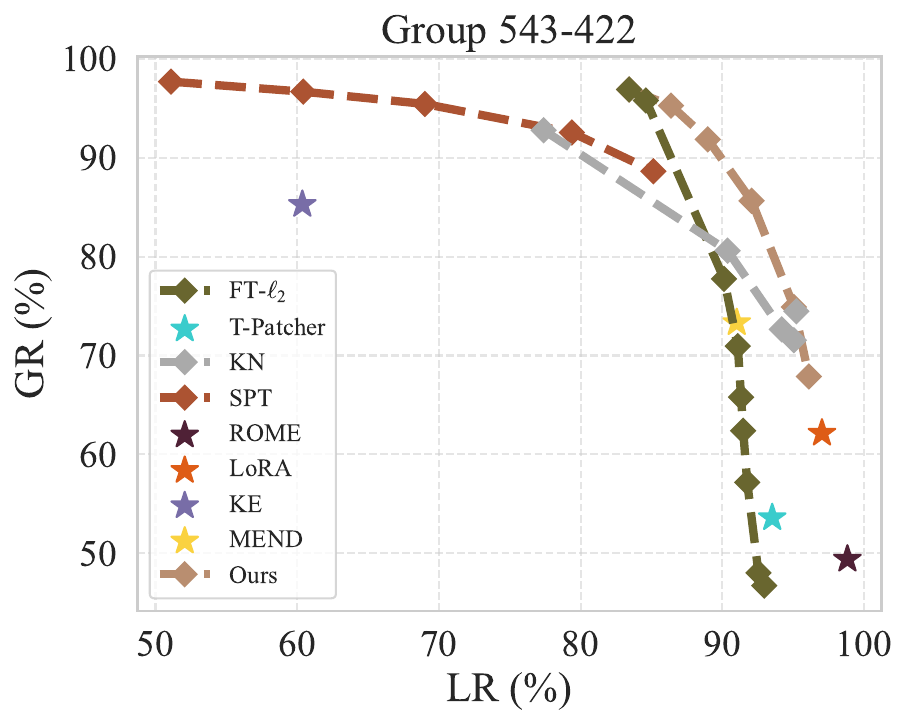}
    \end{subfigure}
    \hfill
    \begin{subfigure}[]{0.24\textwidth}
        \centering
        \includegraphics[width = \columnwidth]{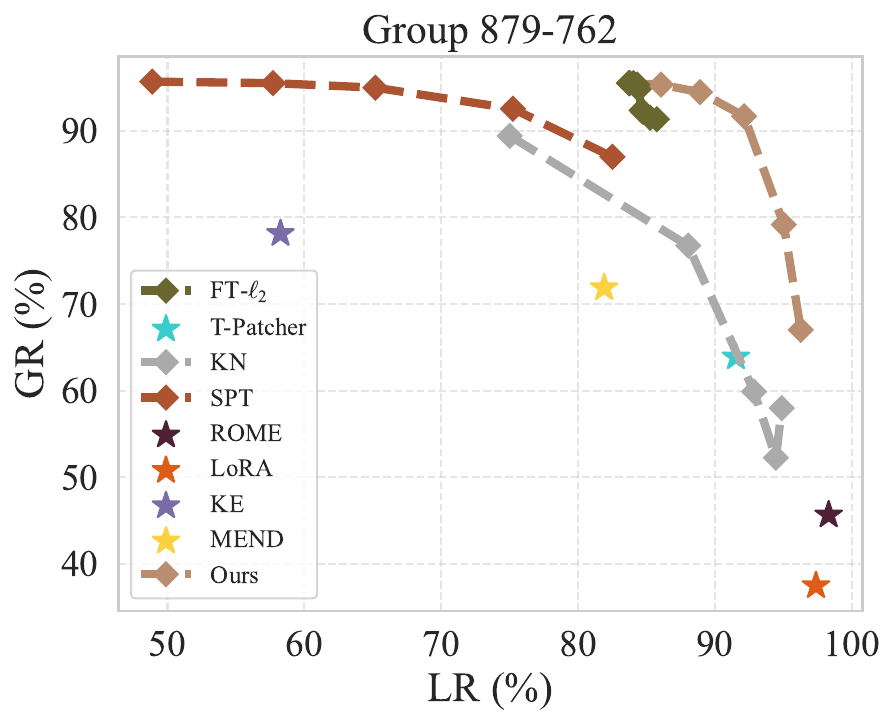}
    \end{subfigure}
    \hfill
    \begin{subfigure}[]{0.24\textwidth}
        \centering
        \includegraphics[width = \columnwidth]{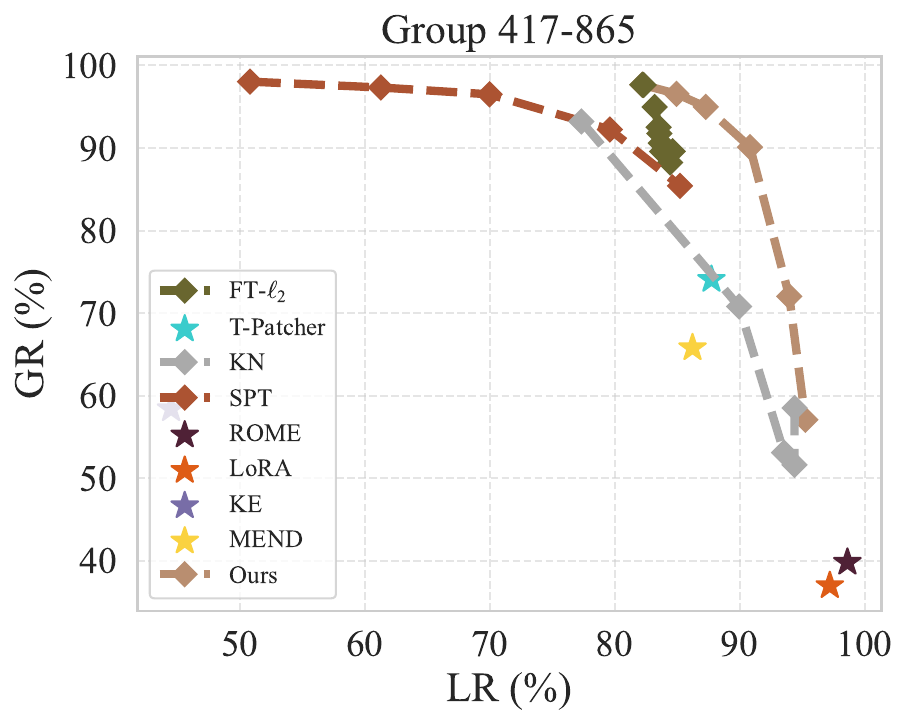}
    \end{subfigure}
    \begin{subfigure}[]{0.24\textwidth}
        \centering
        \includegraphics[width = \columnwidth]{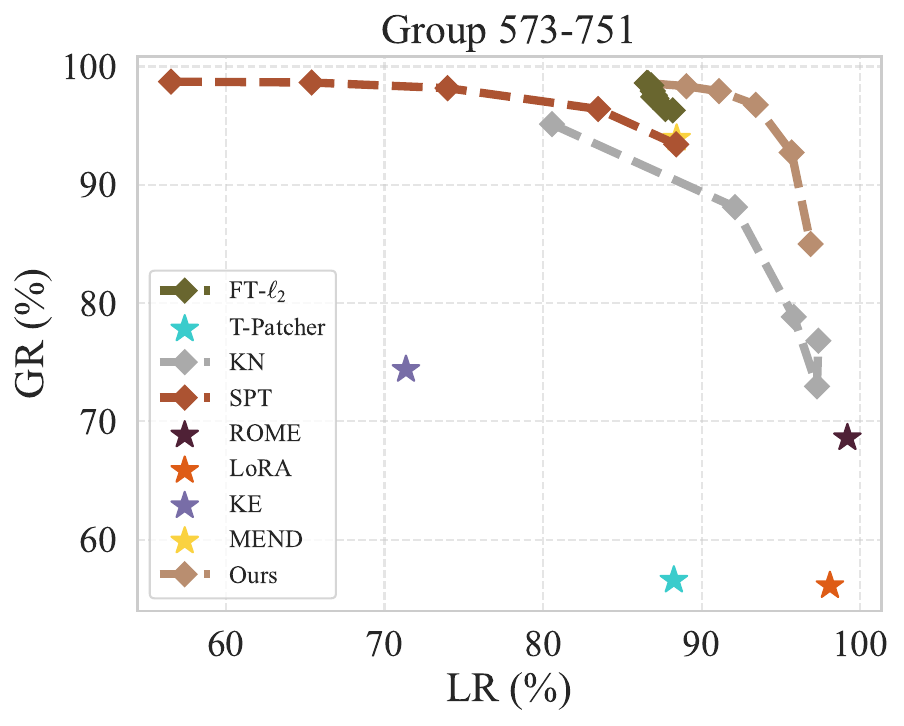}
    \end{subfigure}
    \hfill    
    \begin{subfigure}[]{0.24\textwidth}
        \centering
        \includegraphics[width = \columnwidth]{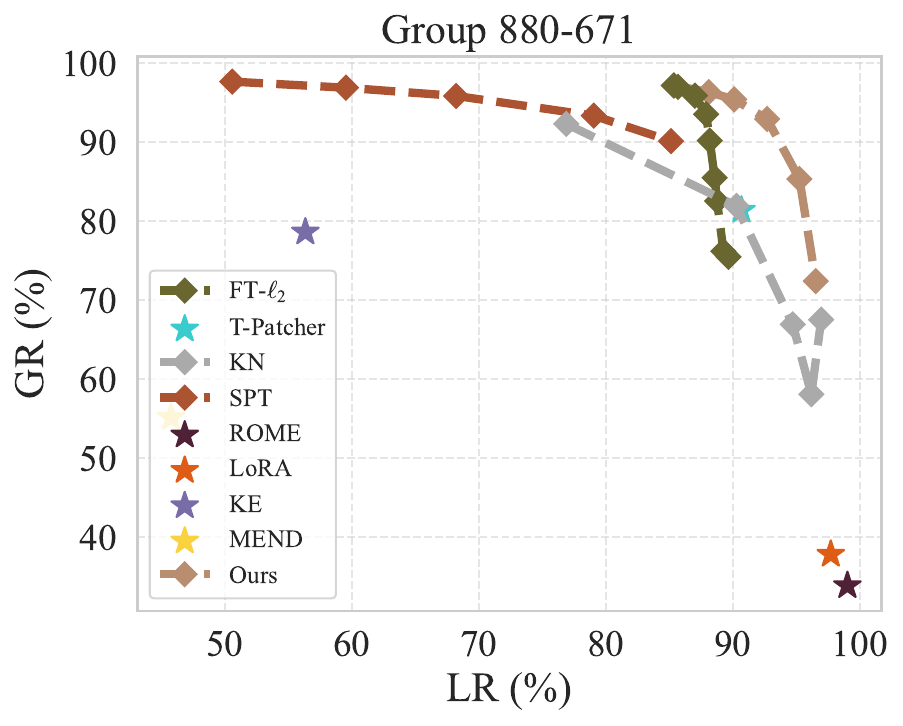}
    \end{subfigure}
    \hfill
    \begin{subfigure}[]{0.24\textwidth}
        \centering
        \includegraphics[width = \columnwidth]{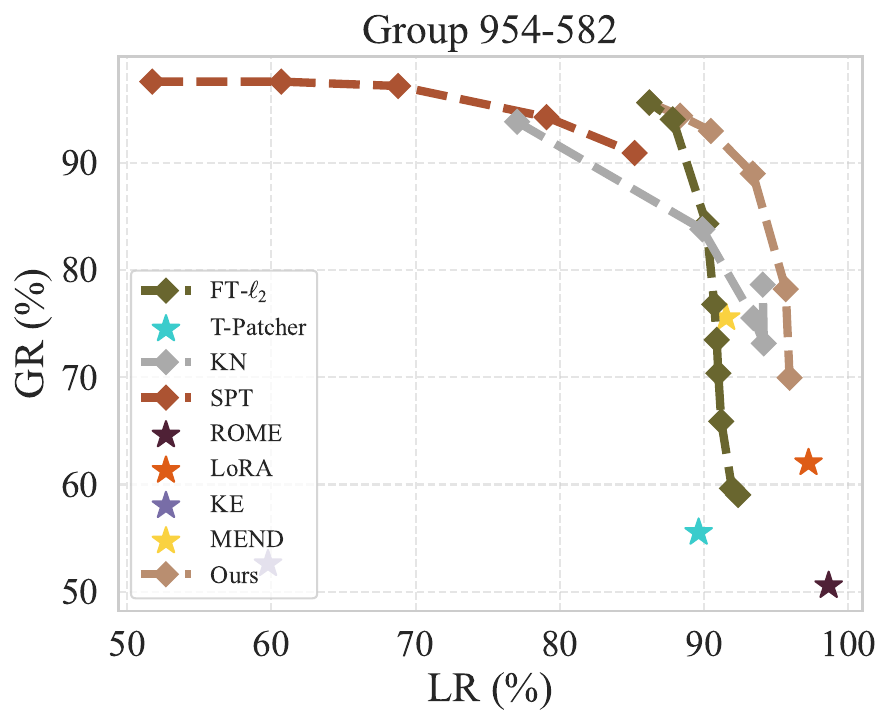}
    \end{subfigure}
    \hfill
    \begin{subfigure}[]{0.24\textwidth}
        \centering
        \includegraphics[width = \columnwidth]{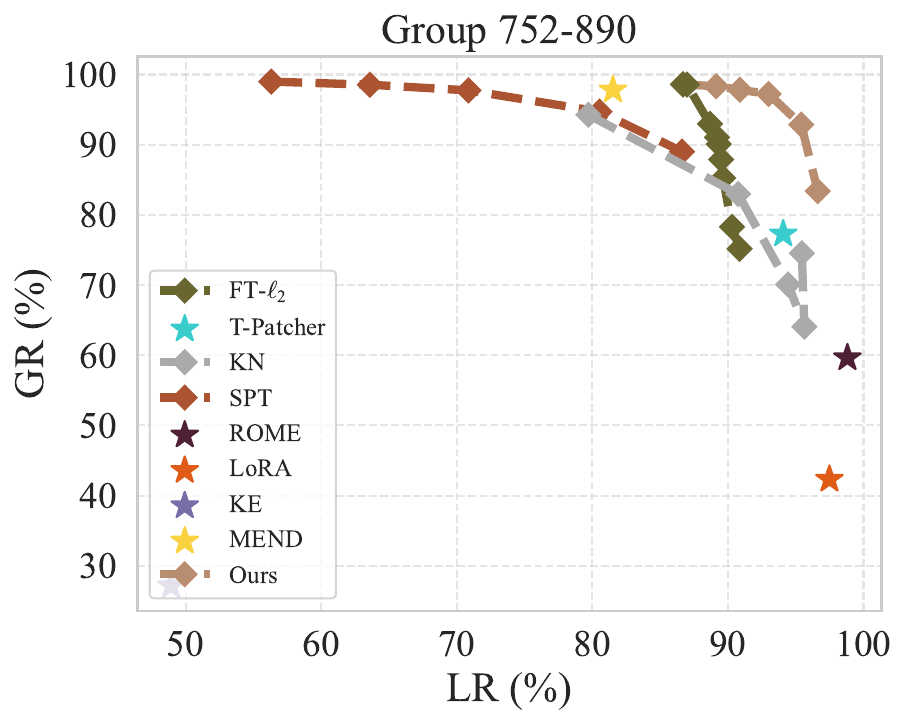}
    \end{subfigure}
            \begin{subfigure}[]{0.24\textwidth}
        \centering
        \includegraphics[width = \columnwidth]{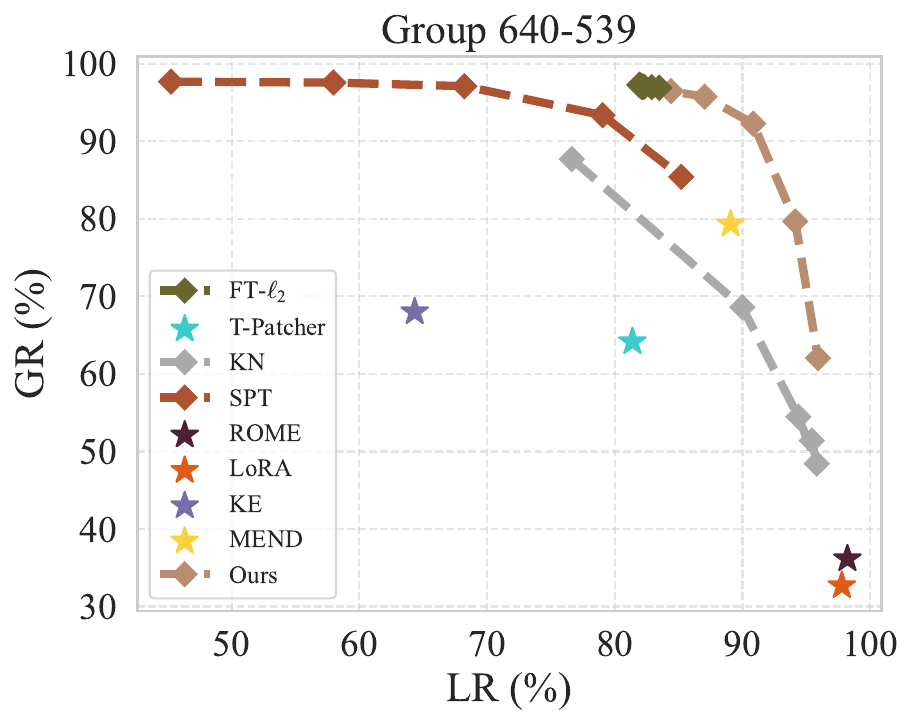}
    \end{subfigure}
    \hfill    
    \begin{subfigure}[]{0.24\textwidth}
        \centering
        \includegraphics[width = \columnwidth]{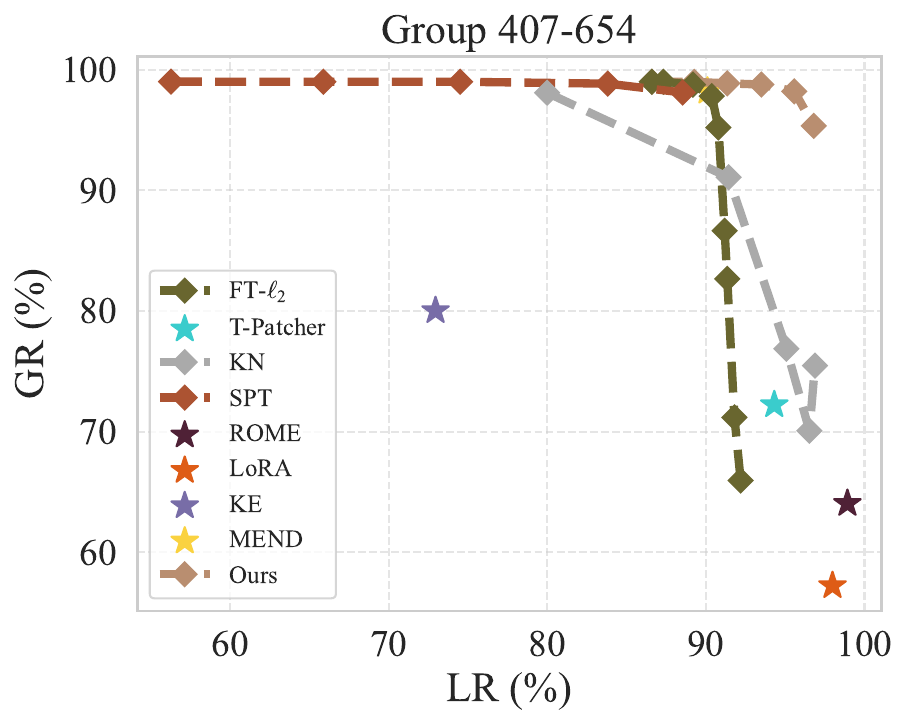}
    \end{subfigure}
    \hfill
    \begin{subfigure}[]{0.24\textwidth}
        \centering
        \includegraphics[width = \columnwidth]{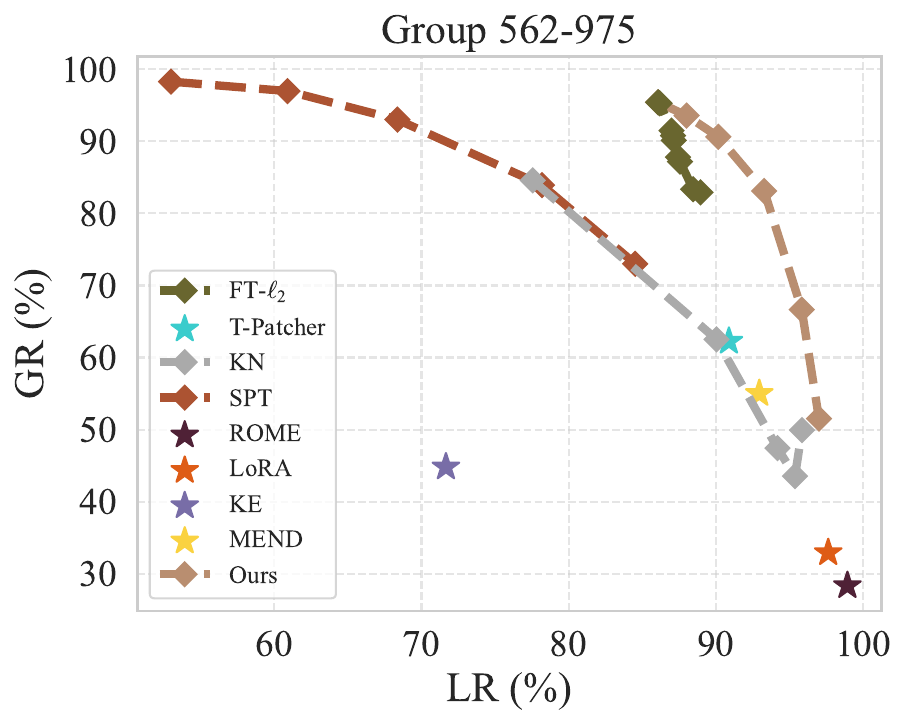}
    \end{subfigure}
    \hfill
    \begin{subfigure}[]{0.24\textwidth}
        \centering
        \includegraphics[width = \columnwidth]{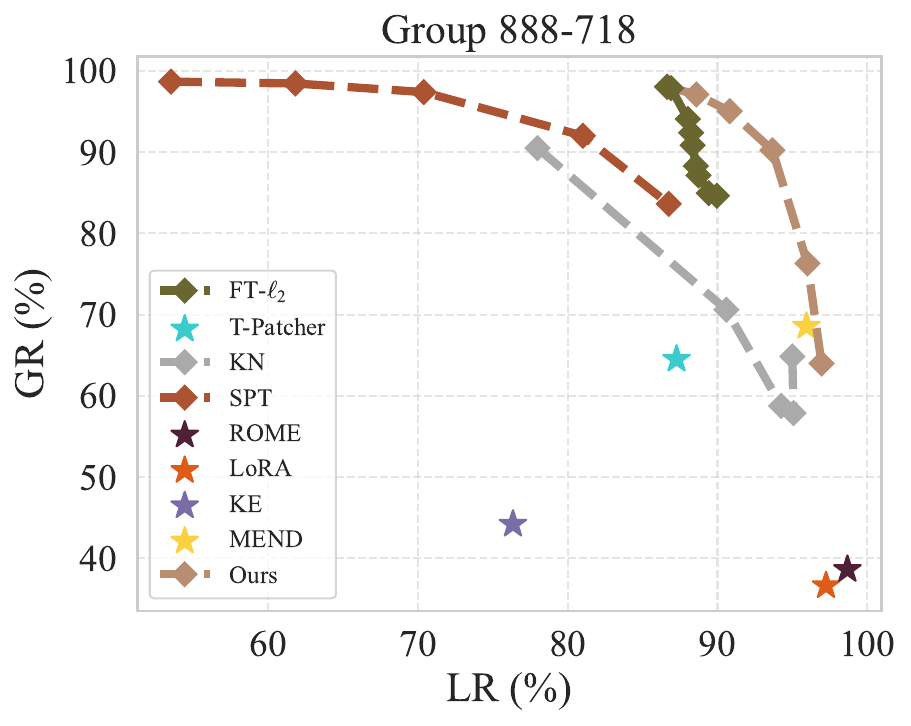}
    \end{subfigure}
\caption{Editing results for ViT/S-16 on the sixteen groups in the natural image subset.}
\label{app_fig:exp_curves_u_s}
\end{figure}

\begin{figure}[ht]
    \centering
    \begin{minipage}{0.49\textwidth}
    \renewcommand\thefigure{H}
        \centering
        \includegraphics[width=\textwidth]{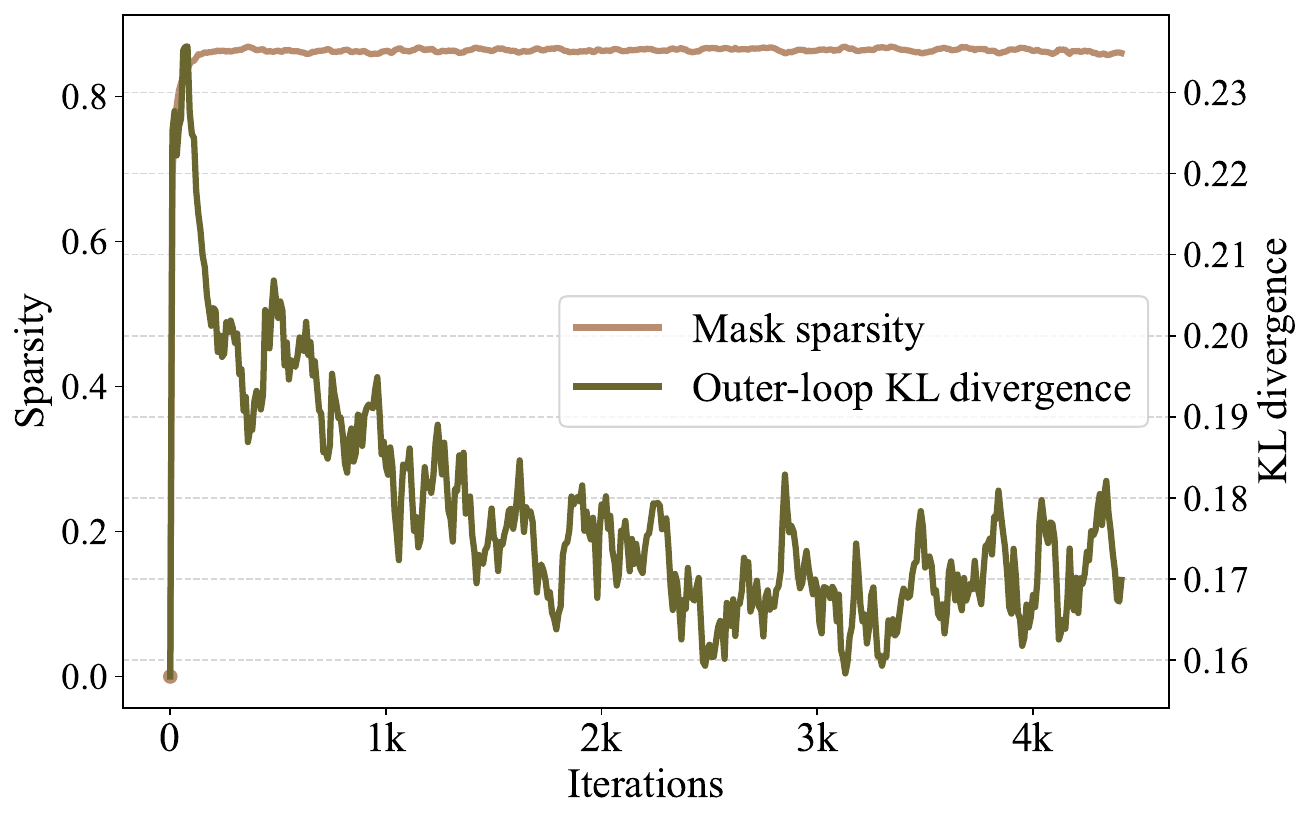}
        \caption{Training curves of the hypernetwork.}
        \label{app_fig:training_loss}
    \end{minipage}\hfill
    \begin{minipage}{0.49\textwidth}
    \renewcommand\thefigure{I}
        \centering
        \includegraphics[width=0.9\textwidth]{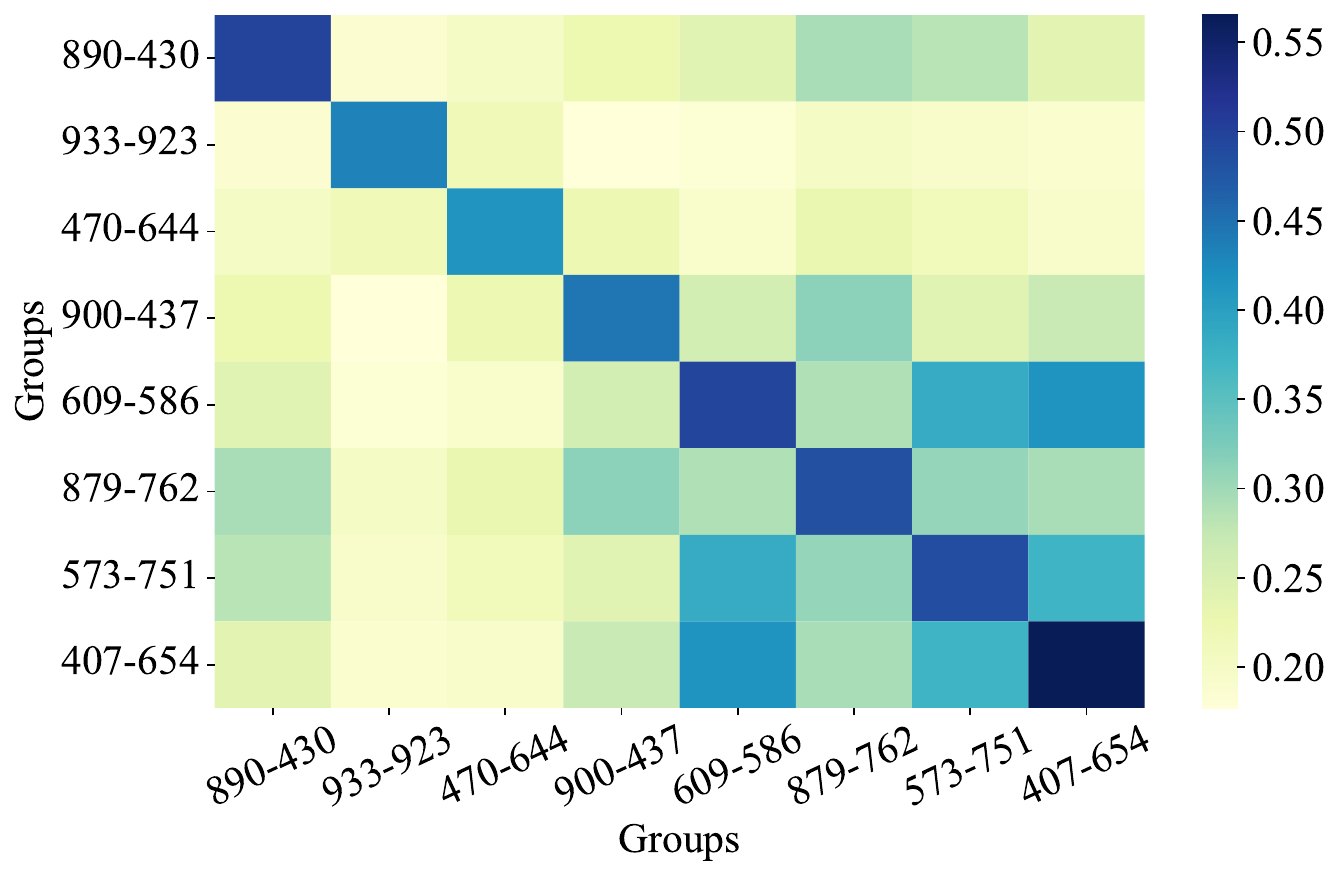}
        \caption{Binary mask IoU results for samples among eight groups of the natural image subset.}
        \label{app_fig:masks_overlaps_B}
    \end{minipage}
\end{figure}

\begin{figure}[t]
\renewcommand\thefigure{J}
        \centering
        \includegraphics[width = \columnwidth]{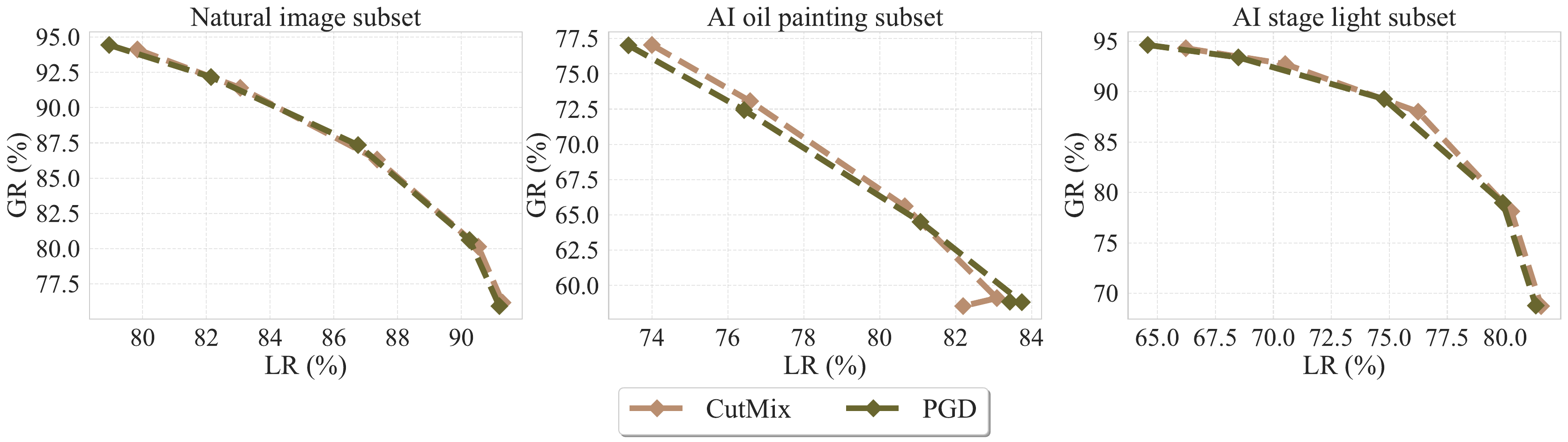}
        \caption{Editing results for ViT/B-16 on the proposed benchmark, using the hypernetworks meta-trained by two different pseudo-sample generation approaches.}
        \label{app_fig:exp_curves_augmentation}
\end{figure}

\begin{figure}[t]
\renewcommand\thefigure{K}
    \centering
    \begin{subfigure}[]{0.49\textwidth}
        \includegraphics[width = \columnwidth]{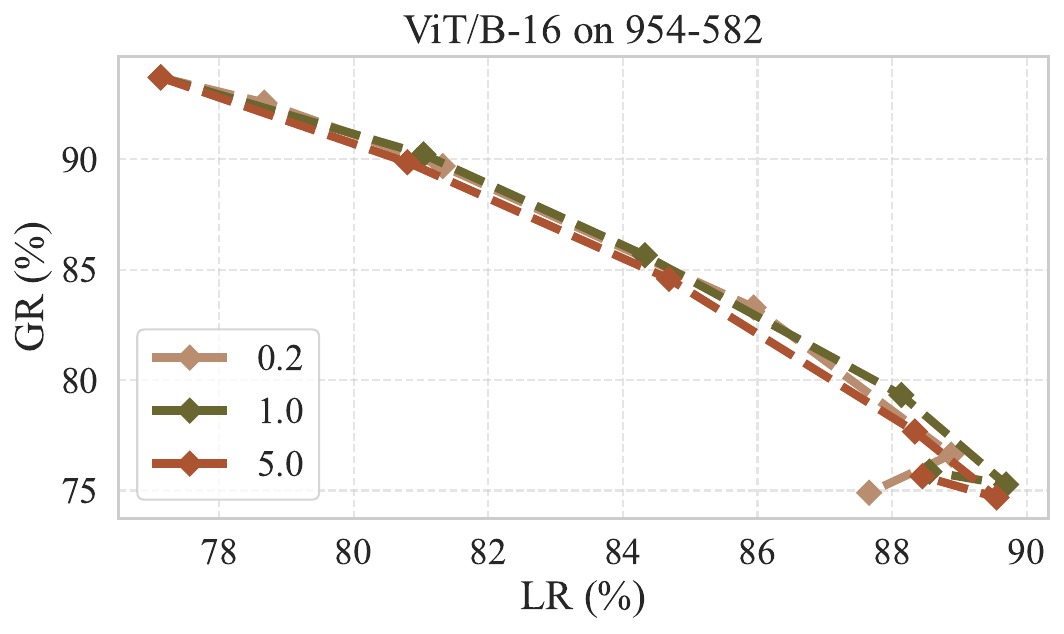}
         \label{fig:954_582_B_sen}
    \end{subfigure}
    \hfill
    \centering
    \begin{subfigure}[]{0.49\textwidth}
        \includegraphics[width = \columnwidth]{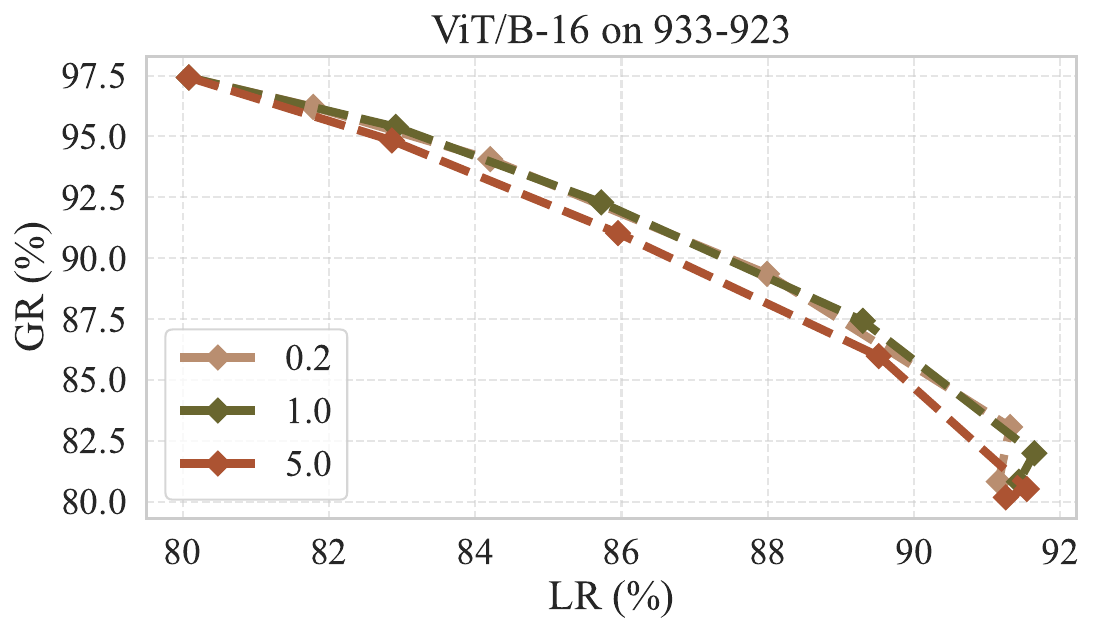}
        \label{fig:933_923_B_sen}
    \end{subfigure}
    \caption{Ablation results of the hyperparameter $\lambda$ in the outer-loop optimization of Problem~\eqref{eq:outer-loops}.}
    \label{app_fig:ab_sparsity}
\end{figure}

\begin{figure} [!t]
\renewcommand\thefigure{L}
    \centering
    \begin{subfigure}[]{0.49\textwidth}
        \centering
        \includegraphics[width = \columnwidth]{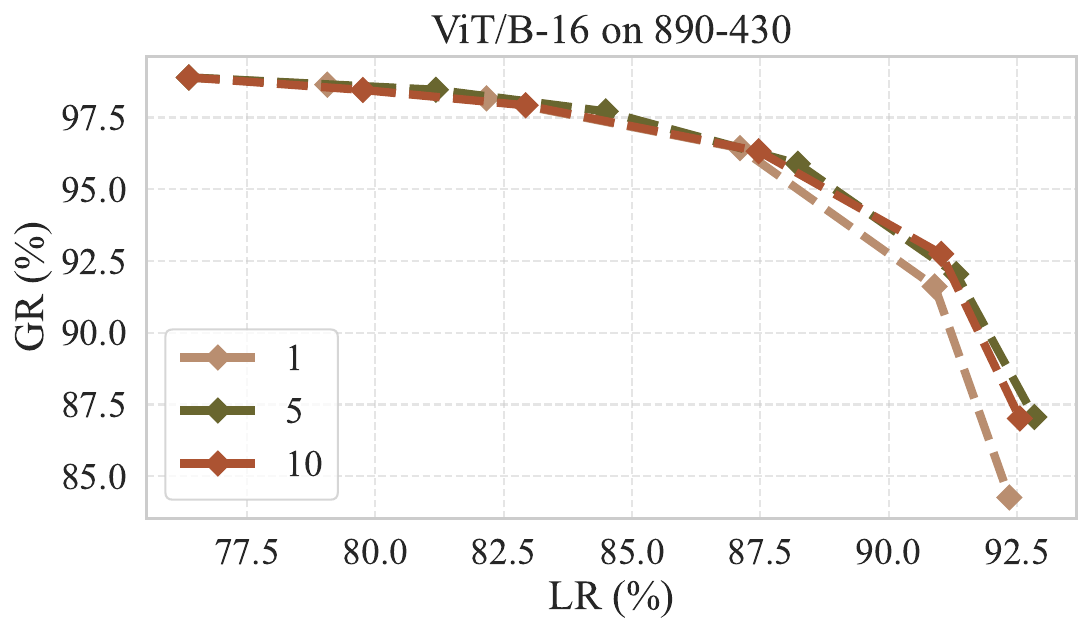}
    \end{subfigure}
    \hfill
    \begin{subfigure}[]{0.49\textwidth}
    \centering
    \includegraphics[width = \columnwidth]{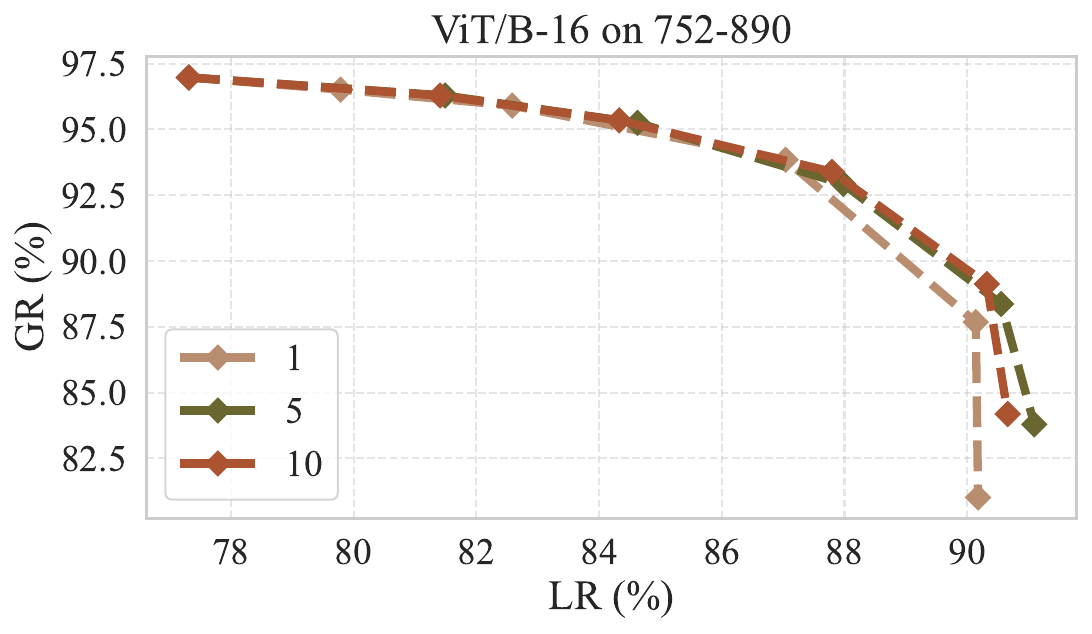}
    \end{subfigure}
\caption{Ablation results of the gradient step $T$ in the inner loop.}
\label{app_fig:ab_mes}
\end{figure}

\begin{figure} [!t]
\renewcommand\thefigure{M}
    \centering
    \begin{subfigure}[]{0.49\textwidth}
        \centering
        \includegraphics[width =\columnwidth]{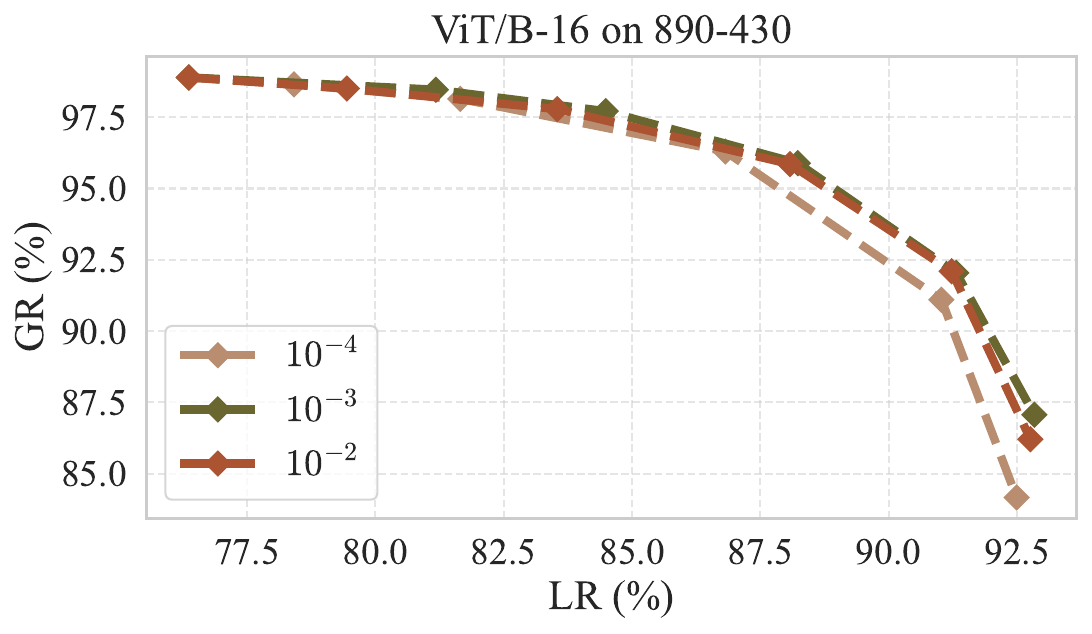}
    \end{subfigure}
    \hfill
    \begin{subfigure}[]{0.49\textwidth}
    \centering
    \includegraphics[width = \columnwidth]{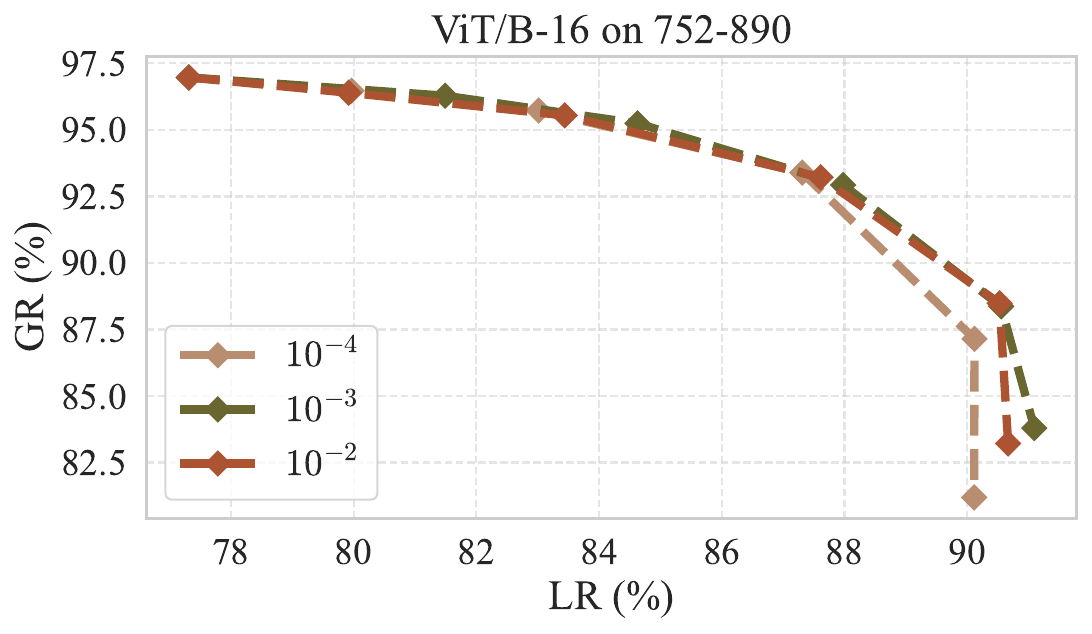}
    \end{subfigure}
\caption{Ablation results of the learning rate in the inner loop.}
\label{app_fig:ab_lr}
\end{figure}

\begin{figure} [!t]
\renewcommand\thefigure{N}
    \centering
    \begin{subfigure}[]{0.49\textwidth}
        \centering
        \includegraphics[width = \columnwidth]{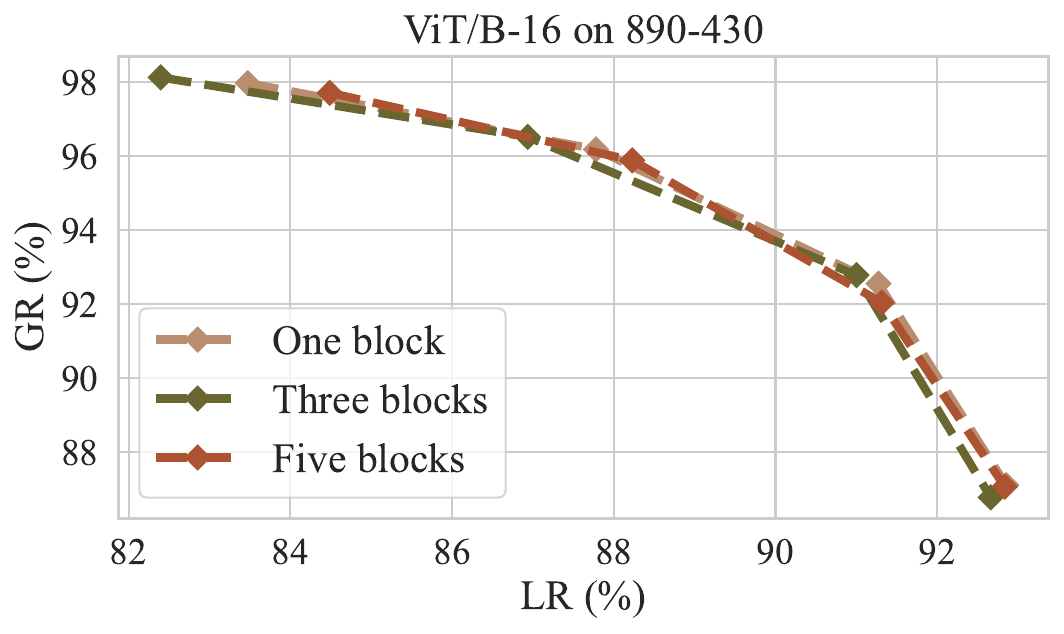}
    \end{subfigure}
    \hfill
    \begin{subfigure}[]{0.49\textwidth}
    \centering
    \includegraphics[width = \columnwidth]{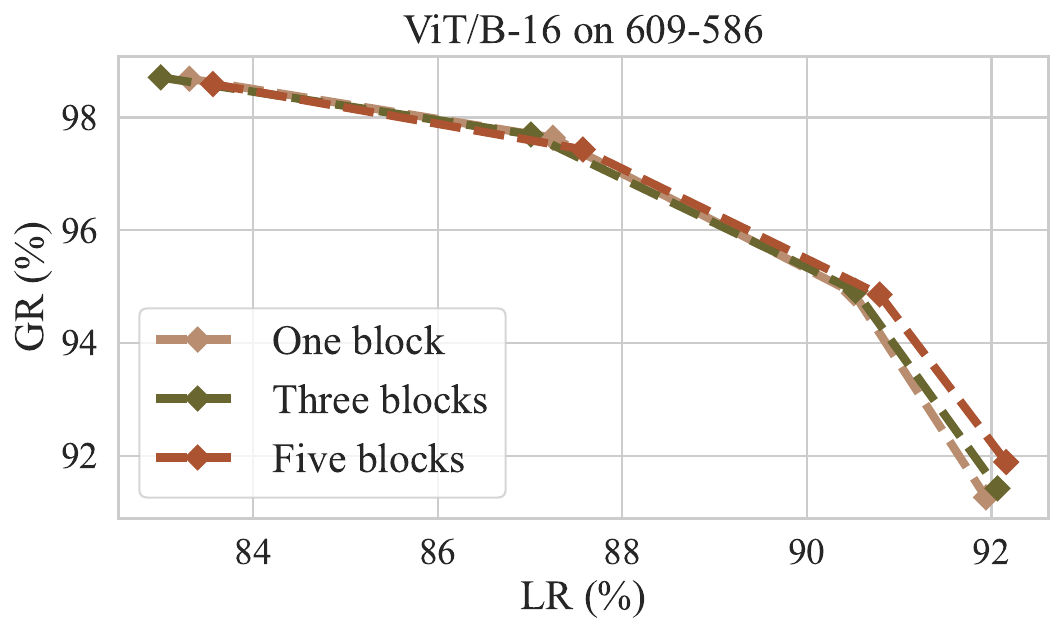}
    \end{subfigure}
\caption{Ablation results of the number of attention blocks in the hypernetwork.}
\label{app_fig:n_blocks}
\end{figure}

\section{More Experimental Results}

\subsection{More Editing Results for ViT/B-16}
In the main paper, we report the averaged editing results for ViT/B-16 across the sixteen groups in the natural image subset. Here, we further report the editing results on each group in Fig.~\ref{app_fig:exp_curves_u_b}.

\subsection{Editing Results for ViT/S-16}
Fig.~\ref{app_fig:exp_curves_vits} presents the editing outcomes for ViT/S-16, where our method continues to exhibit the optimal generation-locality trade-off, demonstrating its adaptability across various model architectures. Meanwhile, Fig.~\ref{app_fig:exp_curves_u_s} presents the editing results on each group in the natural image subset.

\subsection{More Analysis}

We present the training curves of the hypernetwork in Fig.~\ref{app_fig:training_loss}. We find that the mask sparsity increases rapidly at the beginning of training from $0.0$ to $0.86$, which poses challenges for successful edits. As training progresses, the mask sparsity stabilizes while the KL divergence decreases. 
This suggests that the hypernetwork has effectively located key parameters relevant to successful edits.

\subsection{Ablation Studies}

\paragraph{Mask Specificity} We further compute the averaged IoU results of the binary masks at the $0.95$ sparsity level for editing samples among eight groups in the natural image subset. The results in Fig.~\ref{app_fig:masks_overlaps_B} show that the identified parameters exhibit substantial overlaps for samples within the same group and much lower overlaps for samples from different groups. 

\paragraph{Alternative Strategy for Pseudo-sample Generation} 
We examine another more computationally expensive pseudo-sample generation strategy, \ie, PGD~\cite{madry2018towards}, which has been validated to capture diverse distribution variations~\cite{guo2022adversarially, salman2020adversarially}.  Given a natural image $x'$ with the label $y'$ in the pre-training set, we apply PGD~\cite{madry2018towards} on $x'$ to obtain the pseudo-sample $x$ with the prediction different from $y'$. We set the number of attack steps to $10$ with a step size of $2/255$, under the feasible set of $\ell_{\infty}(x,x') \le 8/255$. During training, we employ the cross-entropy loss  $\ell\left(x, y'; {\phi}^{(t)}\right)$ to correct the prediction of $x$. 

Fig.~\ref{app_fig:exp_curves_augmentation} shows the editing results of two hypernetworks meta-trained using the two different pseudo-sample generation approaches. Remarkably, the simple CutMix rivals PGD in simulating distribution shifts, even in the two AI-generated image subsets.

\paragraph{Sparsity Regularization in the Outer Loop}
In the outer loop, we introduce a trade-off hyperparameter, $\lambda$, to balance the reliability objective with the sparsity regularizer. Here, we explore the impact of $\lambda$ and observe that the sensitivity of hypernetwork to this trade-off parameter is minimal, as shown in Fig.~\ref{app_fig:ab_sparsity}.

\paragraph{Gradient Step in the Inner Loop} 
For the gradient step, $T$, in the inner loop, we test values of $\{1, 5, 10\}$. The performance of ViT/B-16 for each setting is illustrated in Fig.~\ref{app_fig:ab_mes}, where we find that one gradient step yields slightly inferior results compared to more steps. Five and ten steps perform similarly, yet ten steps have greater training costs. Thus, we opt for five gradient steps as the default.

\paragraph{Learning Rate in the Inner Loop} 
We explore the impact of the learning rate in the inner loop with values from $\{{10}^{-4}, {10}^{-3}, {10}^{-2}\}$. The editing results shown in Fig.~\ref{app_fig:ab_lr} indicate that a lower learning rate (\ie, ${10}^{-4}$) exhibits slightly inferior performance than a larger learning rate. This may arise because a lower learning rate results in minimal updates to the base model within five gradient steps, thereby ineffective in guiding the hypernetwork training.

\paragraph{Number of Attention Blocks} 
We additionally conduct ablative experiments to evaluate the impact of the number of attention blocks in the hypernetwork. We test values of $\{1, 3, 5\}$, and the editing performance for ViT/B-16 is illustrated in Fig.~\ref{app_fig:n_blocks}, where we find that a small hypernetwork can achieve comparable performance to larger hypernetworks. Decreasing the number of attention blocks in the hypernetwork from five to three, and to one, does not incur a noticeable performance drop.

\section{Limitations}\label{limilation}
See the Conclusion and Discussion section in the main text.

\section{Broader Impact}
Model editing has a broad impact by accelerating innovation in AI development through rapid iterations and refinements without extensive retraining, thus conserving resources and reducing environmental impact. The proposed method enables error correction of CV models, thereby enhancing adaptability and accessibility. We believe our method has great potential in addressing ethical concerns by mitigating biases and improving fairness in CV applications, while also increasing the robustness of CV systems against security threats like adversarial attacks.

\end{document}